\documentclass[10pt]{article}
\usepackage{graphicx} % Required for inserting images
\usepackage[margin=1in]{geometry} % Adjust the margins to 1 inch
\usepackage{url}
\usepackage{pgfplots}
\usepackage{csvsimple}
\usepackage[labelfont=bf,textfont=it]{caption}
\usepackage{subcaption}
\usepackage{tcolorbox}
\usepackage{helvet}
\usepackage{enumitem}
\usepackage{times}
\usepackage{overpic}

\pgfplotsset{compat=1.11,
    /pgfplots/ybar legend/.style={
        /pgfplots/legend image code/.code={%
            \draw[##1,/tikz/.cd,bar width=6pt,yshift=-0.2em,bar shift=0pt]
            plot coordinates {(0cm,0.8em)};},
    },
}

\title{3D Neural Operator-Based Flow Surrogates around 3D geometries: Signed Distance Functions and Derivative Constraints}
\author{Ali Rabeh, Adarsh Krishnamurthy, Baskar Ganapathysubramanian\footnotemark[1]}
\date{}

\usepackage[numbers,sort&compress]{natbib}
\usepackage{multirow} % For multirow cells
\usepackage{array}    % For more flexible table formatting
\usepackage{booktabs} % For enhanced table rules (optional, but recommended)
\usepackage{amsmath,amssymb}

\usepackage{tikz}
\usetikzlibrary{shapes, arrows.meta, positioning}
\usetikzlibrary{calc, positioning}
\usepackage{array}
\usepackage[bookmarks=false]{hyperref}
    \hypersetup{colorlinks,
      linkcolor=blue,
      citecolor=blue,
      urlcolor=blue}

\newcommand{\colref}[2]{\hyperref[#2]{#1~\ref*{#2}}}

\newcommand{\figref}[1]{\colref{Figure}{#1}}

\newcommand{\secref}[1]{\colref{Section}{#1}}
\newcommand{\tabref}[1]{\colref{Table}{#1}}
\newcommand{\appendixref}[1]{\colref{Appendix}{#1}}
\newcommand{\coloredref}[2]{\hyperref[#2]{#1~\ref*{#2}}}
\newcommand{\coloredsubref}[3]{\hyperref[#2]{#1~\ref*{#2}{#3}}}
\definecolor{uColor}{rgb}{0,0,0} % L1 (Velocity MSE) - Black
\definecolor{gradColor}{rgb}{1.0,0.0,0.0} % L3 (Gradient Loss) - Red
\definecolor{boundaryColor}{rgb}{0.6,0.2,0.8}

\usepackage{cleveref} % For enhanced cross-referencing

\begin{document}

\maketitle
\begin{abstract}
\noindent
Accurate modeling of fluid dynamics around complex geometries is critical for applications such as aerodynamic optimization and biomedical device design. While advancements in numerical methods and high-performance computing have improved simulation capabilities, the computational cost of high-fidelity 3D flow simulations remains a significant challenge. Scientific machine learning (SciML) offers an efficient alternative, enabling rapid and reliable flow predictions. 
In this study, we evaluate Deep Operator Networks (DeepONet) and Geometric-DeepONet, a variant that incorporates geometry information via signed distance functions (SDFs), on steady-state 3D flow over complex objects. Our dataset consists of 1,000 high-fidelity simulations spanning Reynolds numbers from 10 to 1,000, enabling comprehensive training and evaluation across a range of flow regimes. To assess model generalization, we apply two train–test splitting strategies: (1) a random split, evaluating (interpolation) performance within the data distribution, and (2) an extrapolatory split, testing performance on unseen flow conditions. 
Additionally, we explore a derivative-informed training strategy that augments standard loss functions with velocity gradient penalties and incompressibility constraints, improving physics consistency in 3D flow prediction. Our results show that Geometric-DeepONet improves boundary-layer accuracy by up to 32\% compared to standard DeepONet. Moreover, incorporating derivative constraints enhances gradient accuracy by 25\% in interpolation tasks and up to 45\% in extrapolatory test scenarios, suggesting significant improvement in generalization capabilities to unseen 3D Reynolds numbers. 
By refining loss function design and integrating rigorous evaluation metrics, this study advances SciML methodologies for scalable and generalizable high-fidelity 3D fluid flow modeling around complex geometries.
\end{abstract}

\section{Introduction}

Accurate modeling and control of fluid flow around complex three-dimensional (3D) structures are essential in a wide range of engineering and scientific applications, including aerodynamic optimization, biomedical device design, and industrial fluid processing. The ability to predict and manipulate flow behavior influences critical outcomes such as drag reduction in aircraft wings, cooling efficiency in heat exchangers, and hemodynamic performance in patient-specific vascular implants. More broadly, phenomena such as turbulent mixing, heat transfer, and chemical dispersion depend on a fundamental understanding of fluid behavior in complex 3D environments~\citep{smyth2009three, sreenivasan2019turbulent}.

Computational fluid dynamics (CFD) solvers provide high-fidelity solutions to these problems, offering physically accurate predictions through numerical discretization of the governing equations. However, these solvers are computationally expensive, particularly for large-scale 3D simulations where the need to resolve intricate boundary-layer effects, turbulence, and geometric complexity leads to prohibitive simulation costs~\citep{Fimbres2010, Chawner2016}. The high computational overhead limits their applicability in real-time design exploration, optimization, and decision-making, creating a demand for more efficient alternatives that can provide rapid and reliable flow predictions.

To address these challenges, surrogate modeling has emerged as a promising approach, combining reduced-order models (ROMs) with data-driven machine learning (ML) techniques to approximate complex flow fields while dramatically reducing computational expense~\citep{Ong2005, Hou2022}. By leveraging historical high-fidelity CFD data, these models learn low-dimensional representations of the flow, enabling orders-of-magnitude acceleration in simulation time without sacrificing essential flow physics. This capability is particularly beneficial in applications requiring rapid prototyping, uncertainty quantification, and real-time feedback, such as automated aerodynamic shape optimization, industrial process control, and medical device design~\citep{Zhao2024, plathottam2023review, keith2025scientific}.

Recent advances in scientific machine learning (SciML) have further enhanced the potential of data-driven surrogate modeling. SciML methods use neural networks to learn nonlinear operator mappings from geometric and flow conditions to full flow-field solutions, bypassing the need to explicitly solve governing equations at each evaluation~\citep{Choudhary2022, li2021}. Several neural operator architectures have demonstrated strong potential for fluid flow modeling, including Fourier Neural Operators (FNOs)~\citep{li2021}, which employ spectral transforms to capture long-range spatial correlations efficiently, Convolutional Neural Operators (CNOs)~\citep{raonic2023}, which utilize hierarchical convolutional architectures for local feature extraction, and Deep Operator Networks (DeepONet)~\citep{lu2020}, which leverage the universal approximation theorem for operators to learn complex solution mappings. More recently, transformer-based models such as Scalable Operator Transformers (scOT) and the Poseidon framework have introduced self-attention mechanisms that improve generalization in physics-informed learning~\citep{herde2024poseidon, liu2022convnet2020s}.

While these methods have demonstrated impressive performance on two-dimensional (2D) flow problems, their scalability and effectiveness in 3D flow modeling remain underexplored~\citep{rabeh2024geometry, Xiao2020survey, NeuFENetKhara2024, Khara2024}. Moving from 2D to 3D simulations introduces substantial challenges, including increased computational memory demands, more complex boundary interactions, and greater difficulty in learning high-dimensional flow representations. Traditional convolution-based architectures struggle with these demands due to the curse of dimensionality, prompting the need for more efficient geometric representations and training strategies~\citep{perera2023enhancing, ullah2024conventional}. Recent studies have explored innovative approaches such as Factorized Implicit Global Convolution (FIGConv)~\citep{choy2025factorized}, which reduces computational overhead by factorizing 3D convolutions into lower-dimensional implicit grids, and MeshNet~\citep{feng2019meshnet}, which applies mesh-based learning techniques to operate directly on unstructured computational grids. Another promising approach is the use of signed distance functions (SDFs)~\citep{he2024}, which offer a continuous, differentiable representation of complex geometries, making them particularly effective for encoding boundary interactions and improving generalization across different flow configurations.

Despite these architectural innovations, a fundamental challenge in SciML-based flow modeling is ensuring physical consistency in learned surrogates. Standard deep learning models often fail to accurately capture velocity gradients and sharp flow features, particularly near complex boundaries, leading to numerical artifacts and reduced predictive reliability. To address this, recent studies have incorporated derivative-informed loss functions, which enforce constraints on velocity derivatives to improve smoothness and accuracy~\citep{qiu2024derivative}. Physics-informed frameworks that integrate conservation laws, such as mass continuity and Navier-Stokes residuals, into the learning process have also shown promise in improving model robustness~\citep{willard2020integrating, nguyen2023parc, weikun2023physics}.

Another critical limitation in existing SciML approaches is their ability to generalize beyond training conditions. While traditional CFD solvers can be applied to any flow scenario given sufficient computational resources, data-driven models often struggle with extrapolating to unseen Reynolds numbers, boundary conditions, and geometric variations~\citep{Liu2024, Wang2020, Ciampiconi2023, Nie2018}. Achieving robust generalization requires not only carefully selected network architectures but also diverse and well-curated training datasets that capture the full range of expected flow phenomena.

Most publicly available CFD datasets are heavily skewed toward 2D geometries, leaving 3D benchmarks underexplored. However, several recent initiatives have aimed to address this gap by creating large-scale 3D flow datasets for SciML benchmarking. Notable datasets include FlowBench~\citep{Tali2024}, which provides high-fidelity 3D flow simulations over diverse geometries and Reynolds numbers, Drivaernet++~\citep{elrefaie2024drivaernet++}, which focuses on aerodynamic applications, and WindsorML~\citep{ashton2024windsorml}, which contains wind tunnel data for various bluff-body configurations. These datasets enable systematic evaluation of SciML surrogates under realistic 3D flow conditions and provide a foundation for assessing the impact of different geometric representations on model accuracy. Empirical studies have shown that the choice of geometric encoding, such as voxel grids, mesh representations, or signed distance fields (SDF), significantly affects the performance of neural operators, reinforcing the importance of dataset design in advancing 3D SciML applications~\citep{he2024, gao2023scientific, rabeh2024geometry, trask2019gmls}.

Given these challenges and recent developments, this study aims to bridge key gaps in SciML for 3D flow modeling by evaluating neural operator-based surrogate models on a canonical 3D lid-driven cavity (LDC) problem, a well-established benchmark for studying recirculating flows in enclosed domains. Specifically, we investigate the performance of DeepONet and Geometric-DeepONet, a variant that incorporates SDF-based geometric encoding, to assess the impact of geometry-aware learning on predictive accuracy. In addition to standard training approaches, we evaluate the effect of derivative-informed loss functions that introduce velocity gradient penalties and incompressibility constraints, enabling the model to better capture boundary-layer effects and enhance physics consistency.

Our specific contributions revolve around two key questions: 
\begin{itemize}[itemsep=0pt,topsep=0pt] 
\item How does incorporating velocity gradients into the loss function affect training accuracy, and which design is most robust? 
\item To what extent can SciML models generalize under distribution shifts, for instance, when tested at Reynolds numbers outside the training range? 
\end{itemize}
To systematically evaluate model performance, we utilize four key metrics that quantify accuracy, boundary-layer fidelity, velocity gradients, and mass conservation. Since multiple performance indicators can complicate direct comparisons, we further introduce a unified score to streamline model ranking. By addressing these critical aspects (dataset diversity, model architecture, and loss function design) this work contributes to advancing scalable and generalizable SciML surrogates for high-fidelity 3D fluid flow modeling. %The findings provide valuable insights into best practices for training physics-aware neural operators, informing future developments in SciML-driven fluid dynamics, multi-physics modeling, and real-time flow control.

The remainder of the paper is organized as follows. \secref{sec:data} details the dataset, including visualization of the flow streamlines and the Signed Distance Field (SDF) of the geometry. \secref{sec:models} describes the SciML architectures and evaluation metrics. \secref{sec:loss-functions} describes the different loss functions designed to train the SciML models. In \secref{sec:results}, we present and analyze empirical findings that evaluate model accuracy and generalization across various geometries. Finally, \secref{sec:conclusion} summarizes key insights, discusses open challenges, and outlines potential pathways for future research.

\section{Overview of 3D CFD Dataset} \label{sec:data}
\subsection{Training Data} \label{subsec:training-data}

We use the three-dimensional (3D) lid-driven cavity (LDC) dataset from the publicly available FlowBench dataset~\citep{Tali2024}, hosted on Hugging Face at \url{https://huggingface.co/datasets/BGLab/FlowBench/tree/main/LDC_NS_3D}. FlowBench provides high-fidelity flow data in an AI-ready format, and is ideal for evaluating and improving scientific machine-learning models. We extended the original dataset to include 1{,}000 high-fidelity 3D CFD simulations, extending the original 500 samples by adding an additional 500 simulations around cubes and cylinders. This expansion enhances the dataset's diversity, contributing a broader range of geometric conditions for improved model generalization. These simulations are performed for 100 analytical geometries---including ellipsoids, toroids, boxes, and cylinders in various orientations--- with a quasi-random sampling used to select 10 Reynolds numbers per geometry such that 5 values lie in the range 10--100 and 5 values lie in the range 100--1{,}000. We limit the maximum Reynolds number of the simulations at 1{,}000 to ensure the flow remains steady-state, thus providing a single field prediction baseline for SciML testing. 

\begin{figure}[ht]
    \centering   
    \begin{subfigure}[b]{0.45\linewidth}
        \centering
        \includegraphics[width=\linewidth]{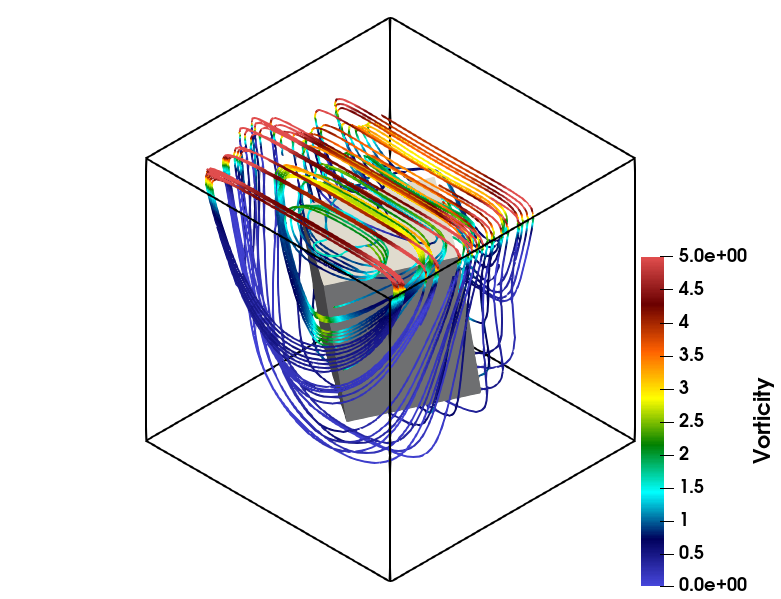}
        \caption{Flow streamlines around a cube at $Re=13$}
    \end{subfigure}
    \hspace{0.02\linewidth}
    \begin{subfigure}[b]{0.45\linewidth}
        \centering
        \includegraphics[width=\linewidth]{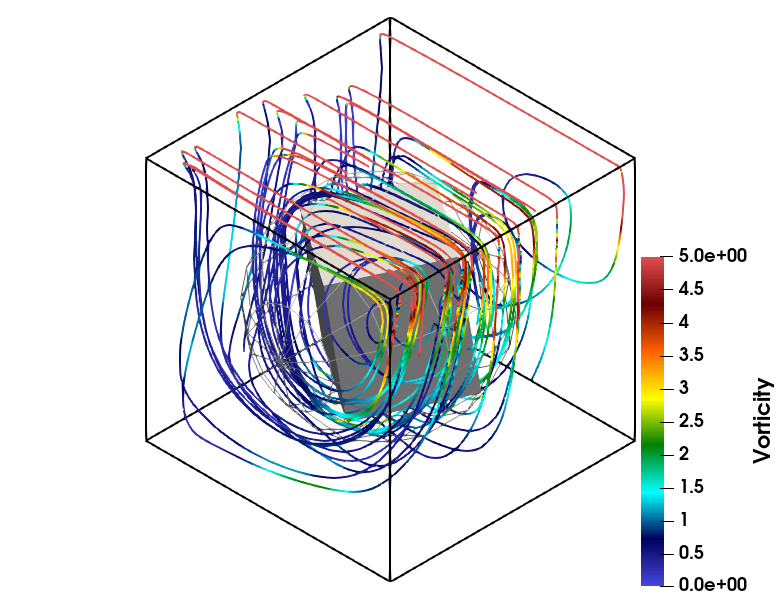}
        \caption{Flow streamlines around a cube at $Re=933$}
    \end{subfigure}
    \hspace{0.02\linewidth}
    \begin{subfigure}[b]{0.45\linewidth}
        \centering
        \includegraphics[width=\linewidth]{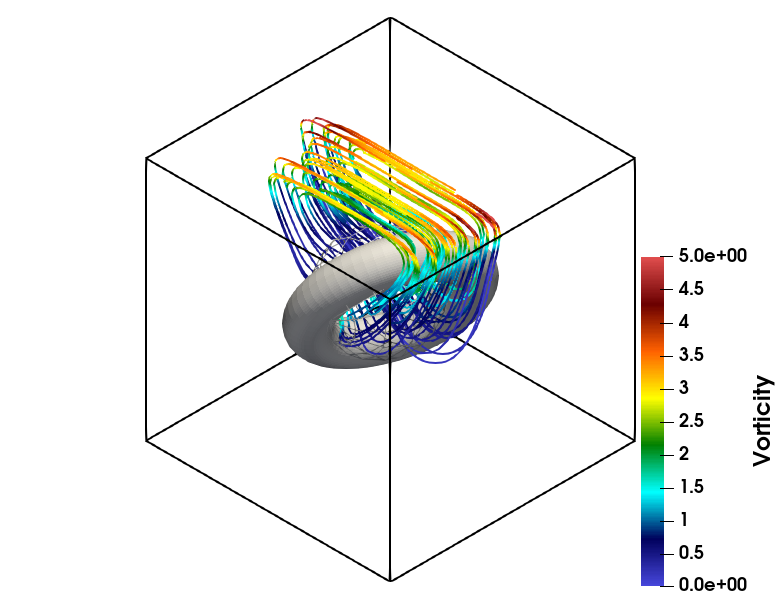}
        \caption{Flow streamlines around a torus at $Re=16$}
    \end{subfigure}
    \hspace{0.02\linewidth}
    % Row 2
    \begin{subfigure}[b]{0.45\linewidth}
        \centering
        \includegraphics[width=\linewidth]{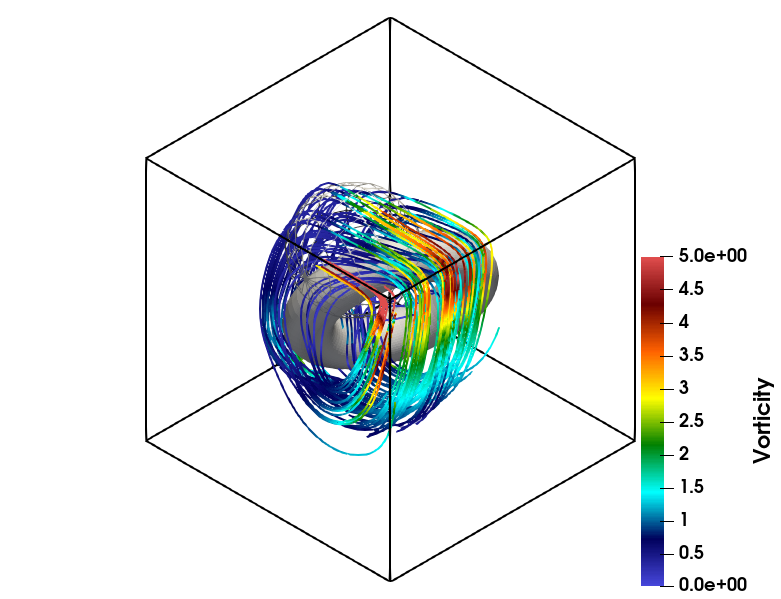}
        \caption{Flow streamlines around a torus at $Re=909$}
    \end{subfigure}
    \hspace{0.02\linewidth}
    \caption{Visualization of flow streamlines for representative geometries from the CFD dataset at various Reynolds numbers. Panels (a) and (b) show the flow around a cube at $Re=13$ and $Re=933$, respectively; panels (c) and (d) show the flow around a torus at $Re=16$ and $Re=909$, respectively.}
    \label{fig:streamlines}
\end{figure}

Data generation relies on a validated Navier--Stokes framework using the shifted boundary method~\citep{main2018shifted, yang2024optimal} to impose boundary conditions on surrogate boundaries. The computational framework used to generate this dataset has been extensively validated against established 3D flow benchmarks. Specifically, FlowBench's 3D LDC simulations were cross-validated with reference results from existing literature, including comparisons of velocity profiles and recirculation structures. Additionally, force coefficients—such as the coefficient of drag \(C_D \) and coefficient of lift \(C_L \)—for various embedded geometries were benchmarked against high-fidelity numerical studies, ensuring an accurate representation of pressure-driven and shear-induced forces. Detailed validation cases against 3D results from the literature, including comparisons of velocity profiles, force coefficients, and vortex structures, can be found in~\citet{yang2024simulating} and this \href{https://baskargroup.bitbucket.io/#/cfdsimulation}{website}. To illustrate the flow characteristics, \figref{fig:streamlines} presents streamline visualizations around two representative objects in the 3D LDC domain. As the Reynolds number increases, additional circulation regions emerge around the objects, reflecting the increasing complexity of the flow structures.

The CFD dataset simulates internal flow around a 3D object placed inside a 3D lid-driven cavity (LDC), which consists of a cubic cavity with five stationary walls and one moving lid. This builds on the well-known CFD benchmark of 3D lid-driven flow, which is known to exhibit complex internal flow patterns, including vortices and recirculation zones~\cite {Johnston2005, Launder2010}. In our simulations, a uniform velocity \(u=1\) is applied at the top boundary (with \(v=0\) and \(w=0\)), while a no-slip condition (\(u=v=w=0\)) is enforced on the remaining five walls. Further details on boundary conditions, geometric variations, and the (octree) mesh refinement strategy are provided in \appendixref{sec:geom-domain}. \figref{fig:3D-LDC-BC} illustrates the used boundary conditions. \figref{fig:geoms} presents representative shapes used in the dataset, while \figref{fig:Mesh_LDC} illustrates mesh slices that highlight local refinement near immersed objects.

%Although the FlowBench dataset includes both velocity and pressure fields, this study focuses exclusively on the velocity components (\(u,v,w\)), omitting pressure from the analysis. This choice is driven by the critical role of velocity distributions in characterizing flow behavior around complex geometries, particularly in regions with strong shear and recirculation. By prioritizing velocity fields, we aim to assess model performance in capturing flow structures, gradients, and transport phenomena essential for aerodynamic and hydrodynamic applications. Future extensions may incorporate the pressure field.

The datasets are provided as numpy compressed (\texttt{.npz}) files. We provide two \texttt{.npz} files—one for the inputs, suffixed with the marker \texttt{"\_X.npz"}, and one for the outputs, suffixed with the marker \texttt{"\_Y.npz"}. Each of these \texttt{.npz} files contains a 5D \texttt{numpy} tensor structured as follows:

{\small
\[
    [\mathbf{samples}][\mathbf{number\_of\_channels}][\mathbf{resolution\_x}][\mathbf{resolution\_y}][\mathbf{resolution\_z}]
\]
}

The dataset consists of 1{,}000 samples, with a spatial resolution of \(128 \times 128 \times 128\). The input tensor has 2 channels: one representing the Signed Distance Field (SDF) and the other a constant field for the Reynolds number. Similarly, the output tensor has 4 channels, corresponding to the velocity components (\(u,v,w\)) and pressure (\(p\)). This format ensures easy retrieval of simulation data, enabling seamless integration into machine learning workflows for training and evaluation. We choose to focus only on the velocity components in this work.

\subsection{Geometric Representation} \label{subsec:geometric-representation}

\begin{figure}[ht]
    \centering   
    % Row 1
    \begin{subfigure}[b]{0.45\linewidth}
        \centering
        \includegraphics[width=\linewidth]{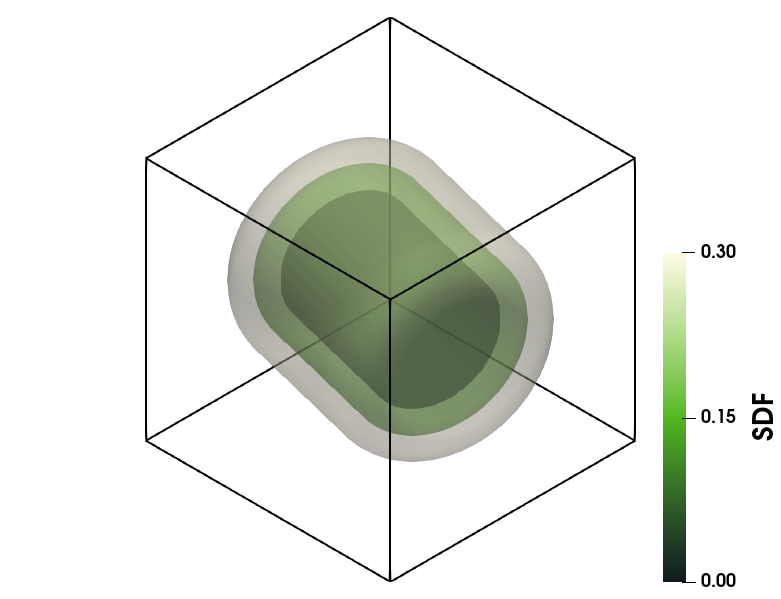}
        \caption{Cylinder}
    \end{subfigure}
    \hspace{0.02\linewidth}
    \begin{subfigure}[b]{0.45\linewidth}
        \centering
        \includegraphics[width=\linewidth]{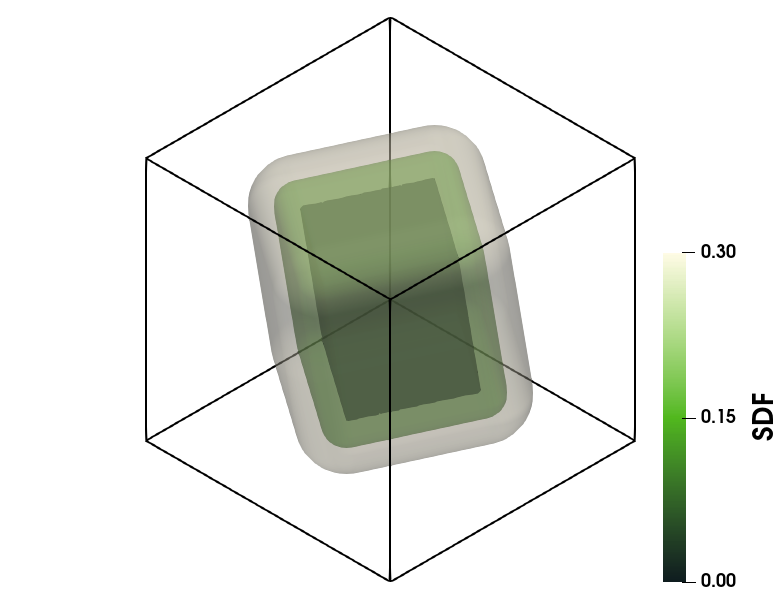}
        \caption{Cube}
    \end{subfigure}
    \hspace{0.02\linewidth}
    \begin{subfigure}[b]{0.45\linewidth}
        \centering
        \includegraphics[width=\linewidth]{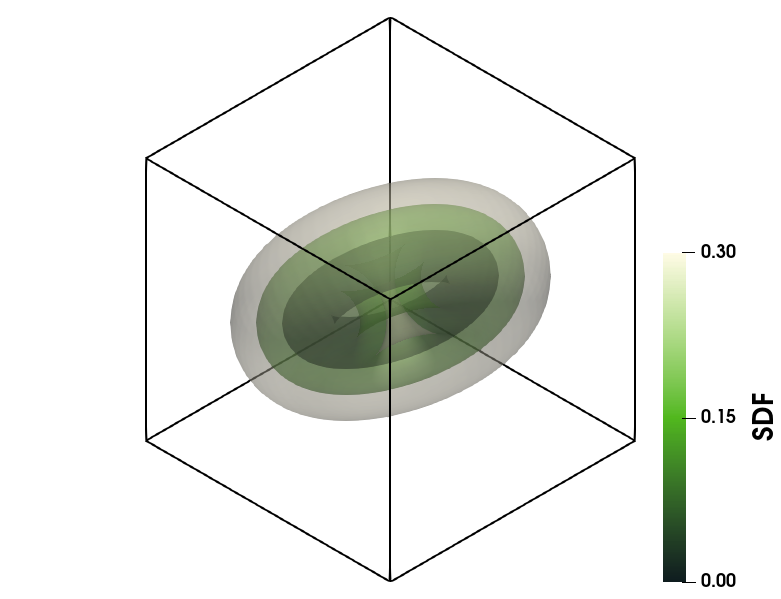}
        \caption{Ring}
    \end{subfigure}
    \hspace{0.02\linewidth}
    % Row 2
    \begin{subfigure}[b]{0.45\linewidth}
        \centering
        \includegraphics[width=\linewidth]{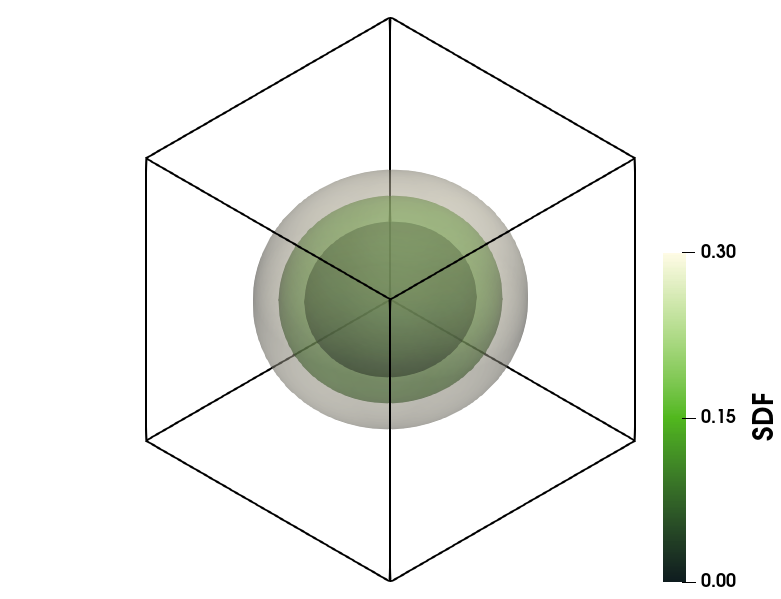}
        \caption{Ellipsoid}
    \end{subfigure}
    \hspace{0.02\linewidth}
    \caption{Level-set visualization of the Signed Distance Field (SDF) for four representative geometries from our dataset, arranged in a 2$\times$2 layout. Panel (a) depicts the SDF of a cylinder, panel (b) shows the SDF of a cube, panel (c) illustrates the SDF of a ring, and panel (d) presents the SDF of an ellipsoid. The level-sets are displayed for three isocontours: SDF = 0 at the geometry's surface, SDF = 0.15, and SDF = 0.3 for two additional distances from the surface respectively.}
    \label{fig:sdf}
\end{figure}

We utilize the Signed Distance Field (SDF) as the geometric representation of our shapes. The SDF is a scalar field that encodes the shortest distance from any point in the prediction domain to the object's boundary. It assigns negative values to points inside the object, positive values to points outside, and zero to points located on the surface. This implicit representation is both continuous and differentiable, offering a more detailed depiction of boundary-layer regions compared to a simple binary mask. Examples of the SDF visualization for four representative geometries from our dataset are shown in \figref{fig:sdf}.

\section{SciML Models} \label{sec:models}

\subsection{DeepONet and Geometric-DeepONet} \label{subsec:depo-geo-models}
\begin{figure}[ht]
    \centering
\begin{tikzpicture}[
    font=\small,
    >=stealth,
    node distance=1.2cm,
    block/.style={
      rectangle,
      draw,
      thick,
      rounded corners,
      align=center,
      fill=blue!10,
      minimum width=2cm,    % reduced width
      minimum height=0.7cm, % reduced height
      inner sep=2pt
    },
    mlp/.style={
      rectangle,
      draw,
      thick,
      rounded corners,
      align=center,
      fill=blue!20,
      minimum width=1.8cm,   % reduced width
      minimum height=0.6cm,  % reduced height
      inner sep=2pt
    },
    circ/.style={
      circle,
      draw,
      thick,
      minimum size=7mm,
      fill=blue!5,
      align=center
    }
]    %------------------ Top Section: DeepONet -----------------%
    \node[font=\large] (titleA) at (0,2.5) {\textbf{DeepONet}};
    % Branch Input
    \node[block] (branchInA) at (-2.0,1.0)
        {\Large {Re, SDF}};
    % Branch Net
    \node[mlp, right=0.8cm of branchInA] (branchMLP) 
        {\Large Branch Net};
    % Trunk Input
    \node[block, below=1.5cm of branchInA] (trunkInA)
        {\Large ${x,y,z}$};
    % Trunk Net
    \node[mlp, right=0.8cm of trunkInA] (trunkMLP)
        {\Large Trunk Net};
    % Dot product node with label on top
    \node[circ, right=1.3cm of branchMLP, yshift=-1.2cm, label=above:{\footnotesize dot product}] (dotA)
    {\Large $\otimes$};

    % Output block
    \node[block, right=1.3cm of dotA] (outA)
        {\Large velocity: {$u,v,w$}};
    
    %----- Draw connections for DeepONet -----
    \draw[->, ultra thick] (branchInA.east) -- (branchMLP.west);
    \draw[->, ultra thick] (trunkInA.east) -- (trunkMLP.west);
    \draw[->, ultra thick] (branchMLP.east) -- (dotA.west);
    \draw[->, ultra thick] (trunkMLP.east) -- (dotA.west);
    \draw[->, ultra thick] (dotA.east) -- (outA.west);
    %------------------ Bottom Section: Geom-DeepONet -----------------%
    \node[font=\large, below=5.0cm of titleA] (titleB)
        {\textbf{Geometric-DeepONet}};
    % Branch Input (Geom)
    \node[block] (branchInB) at (-2.0,-4.5)
        {\Large {Re, SDF}};
    % Branch Net (Geom)
    \node[mlp, right=0.4cm of branchInB] (branchMLPB)
        {\Large Branch Net};
    % Trunk Input (Geom)
    \node[block, below=1.5cm of branchInB] (trunkInB)
        {\Large ${x,y,z}$, SDF};
    % Trunk Net (Geom)
    \node[mlp, right=0.4cm of trunkInB] (trunkMLPB)
        {\Large Trunk Net};
    % Dot product node (geom)
    \node[circ, right=0.6cm of branchMLPB, yshift=-1.2cm, label=above:{\footnotesize dot product}] (dotB)
        {\Large $\otimes$};
        % Branch Net2 (Geom)
    \node[mlp, right=1.9cm of branchMLPB] (branchMLPB2)
        {\Large Branch Net}; 
        % Trunk Net2 (Geom)
    \node[mlp, right=1.9cm of trunkMLPB] (trunkMLPB2)
        {\Large Trunk Net}; 
        % Dot product node (geom)
    \node[circ, right=0.6cm of branchMLPB2, yshift=-1.2cm, label=above:{\footnotesize dot product}] (dotB2)
        {\Large $\otimes$};    
    % Output block (geom)
    \node[block, right=0.8cm of dotB2] (outB)
        {\Large velocity: {$u,v,w$}};
    %----- Draw connections for Geom-DeepONet -----
    \draw[->, ultra thick] (branchInB.east) -- (branchMLPB.west);
    \draw[->, ultra thick] (trunkInB.east) -- (trunkMLPB.west);
    \draw[->, ultra thick] (branchMLPB.east) -- (dotB.west);
    \draw[->, ultra thick] (trunkMLPB.east) -- (dotB.west);
    \draw[->, ultra thick] (dotB.east) -- (branchMLPB2.west);
    \draw[->, ultra thick] (dotB.east) -- (trunkMLPB2.west);
    \draw[->, ultra thick]  (branchMLPB2.east) -- (dotB2.west);
    \draw[->, ultra thick]  (trunkMLPB2.east) -- (dotB2.west); 
    \draw[->, ultra thick] (dotB2.east) -- (outB.west);    
    \end{tikzpicture}
    \caption{Comparison of \emph{DeepONet} (top) and \emph{Geometric-DeepONet} (bottom) on the 3D driven-cavity flow problem. \emph{DeepONet}~\citep{lu2020} uses a dual-network structure: the \emph{branch network} (MLP) encodes input parameters, while the \emph{trunk network} (MLP) processes spatial coordinates \((x,y,z)\). The two latent representations are fused via a dot product to predict velocity. \emph{Geometric-DeepONet}~\citep{he2024} augments the trunk input with geometric information (e.g., SDF) and employs a two-stage process: in \emph{Stage~1}, both branch and trunk networks extract features using conventional ReLU activations; in \emph{Stage~2}, the branch network continues with ReLU while the trunk network refines its features using sinusoidal (SIREN) activations, enabling more accurate capture of complex high frequency geometric details.}
    \label{fig:deepo-geo-archetecture}
\end{figure}
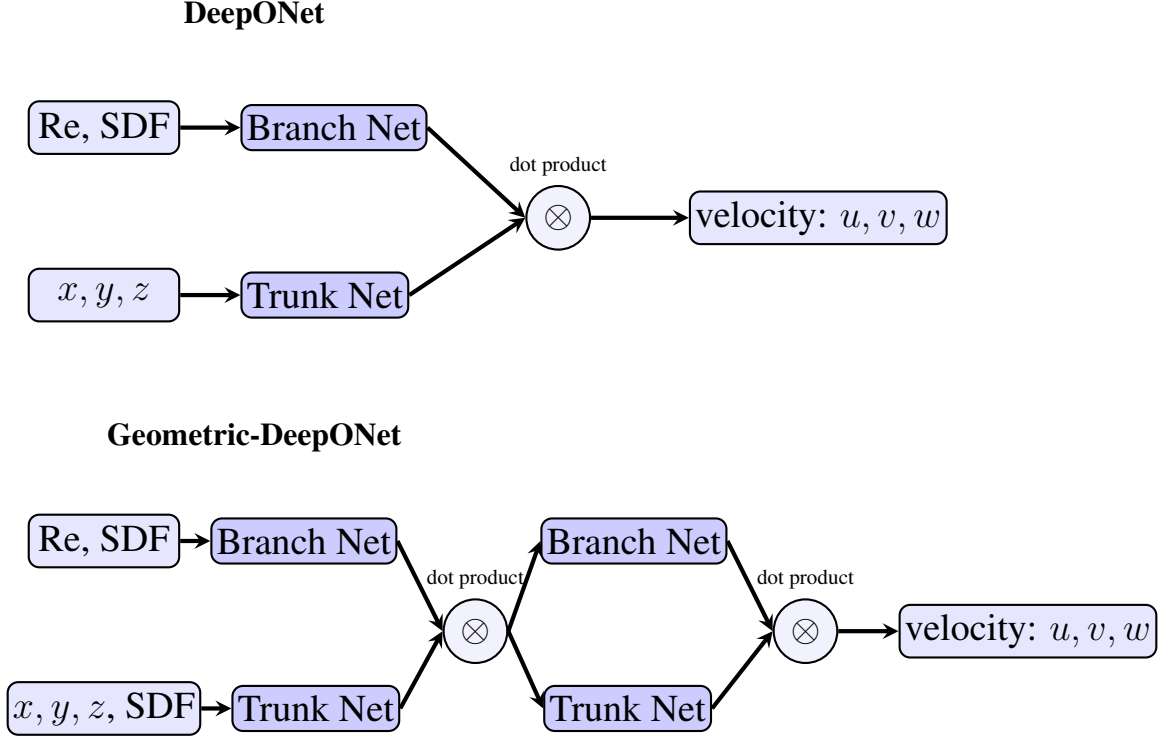

We evaluate the deep operator network \emph{DeepONet} and its geometric extension, \emph{Geometric-DeepONet}, on the 3D driven-cavity flow dataset. \emph{DeepONet}~\citep{lu2020} employs a dual-network architecture consisting of branch and trunk networks, inspired by the universal approximation theorem for arbitrary continuous functions~\citep{Chen1995}. The branch network encodes input functions or parameters (e.g., Reynolds number, boundary masks), while the trunk network processes the spatial coordinates \((x,y,z)\). Their outputs are merged via a dot product to produce predictions. \emph{Geometric-DeepONet}~\citep{he2024} builds on this framework by integrating signed distance functions (SDF) into the trunk network and adopting a two-stage approach: (1)~latent features are extracted from the branch and trunk networks with conventional ReLU activations, and (2)~these features are further refined by sinusoidal (SIREN) activations, which are particularly effective at capturing high-frequency variations and subtle details in complex 3D geometries. This two-stage approach not only improves the overall fidelity of the approximation but also significantly enhances the model’s ability to resolve intricate geometric and flow features. As illustrated in~\figref{fig:deepo-geo-archetecture}, this two-stage design offers enhanced accuracy for problems where intricate geometry is critical to the flow solution.

In this study, we evaluate the performance of \emph{DeepONet} and \emph{Geometric-DeepONet} models in simulating the complex fluid dynamics of the lid-driven cavity flow dataset from FlowBench. These models were trained on a single A100 80GB GPU for 48 hours using the Adam optimizer for 200 epochs, and the list of hyperparameters used is provided in \appendixref{subsec:hyperparameter}. Our study is structured around two core aspects: \emph{loss function design} and \emph{ability to extrapolate} to out-of-sample distributions. The code for the models, along with various loss functions, training procedures, and visualization scripts, is available at \url{https://github.com/baskargroup/DI-DeepONet}.

\subsection{Evaluation Metrics} \label{subsec:metrics}

To comprehensively assess model performance, we introduce four evaluation metrics—\(M1\), \(M2\), \(M3\), and \(M4\)—each focusing on one aspect of the SciML models' predictive accuracy:

\begin{itemize}
    \item \textit{\(M1\): Global Accuracy} --- This metric quantifies the overall fidelity of the predicted velocity fields across the entire computational domain (excluding the embedded geometry). It provides an aggregate measure of how well the model's outputs match the ground truth. The relative \(L_2\) error for \(M1\) is computed as:
    
    \[
    M1 = 100 \times \left( 1 - \frac{\sqrt{\sum_{i=1}^{N}\sum_{k=1}^{3} \left( u_i^{(k)} - \hat{u}_i^{(k)} \right)^2}}{\sqrt{\sum_{i=1}^{N}\sum_{k=1}^{3} \left( u_i^{(k)} \right)^2}} \right),
    \]
    
    where \(u_i^{(1)}\), \(u_i^{(2)}\), and \(u_i^{(3)}\) correspond to the ground truth values of the velocity components \(u\), \(v\), and \(w\) at the \(i\)-th element center, and \(N = 128 \times 128 \times 128\) is the total number of elements.

    \item \textit{\(M2\): Boundary Layer Accuracy} --- Focusing on the near-surface region defined by a Signed Distance Field (SDF) in the range \(0 \leq \text{SDF} \leq 0.2\), this metric evaluates errors within the boundary layer surrounding the object. Since this region is critical for capturing near-surface dynamics, \(M2\) serves as a stringent test of model precision. The formula is computed as:
    
    \[
    M2 = 100 \times \left( 1 - \frac{\sqrt{\sum_{i \in \Omega_{\text{SDF}}} \sum_{k=1}^{3} \left( u_i^{(k)} - \hat{u}_i^{(k)} \right)^2}}{\sqrt{\sum_{i \in \Omega_{\text{SDF}}} \sum_{k=1}^{3} \left( u_i^{(k)} \right)^2}} \right),
    \]

    where \(\Omega_{\text{SDF}}\) represents the set of elements satisfying \(0 \leq \text{SDF} \leq 0.2\).

    \item \textit{\(M3\): Derivative Accuracy} --- This metric measures the accuracy of the spatial gradients of the velocity field across the domain (excluding the geometry). By evaluating the relative \(L_2\) error of the predicted velocity gradients, \(M3\) provides a more rigorous assessment than \(M1\) by ensuring that the model accurately captures derivative information. Since the velocity gradient is a \(3 \times 3\) tensor, the final error is computed by first evaluating the relative error for each component and then averaging over all nine components using a factor of \(1/9\):
    
    \[
    M3 = 100 \times \left( 1 - \frac{1}{9} \sum_{m,n=1}^{3} \frac{\sqrt{\sum_{i=1}^{N} \left( \partial_m u_i^{(n)} - \partial_m \hat{u}_i^{(n)} \right)^2}}{\sqrt{\sum_{i=1}^{N} \left( \partial_m u_i^{(n)} \right)^2}} \right).
    \]
    
    where \(\partial_m u_i^{(n)}\) denotes the \(m\)-th spatial derivative of the \(n\)-th velocity component at element \(i\).
    
    \item \textit{\(M4\): Continuity Consistency} --- This metric assesses the model's adherence to the continuity equation by quantifying the integrated \(L_2\) norm of the continuity residuals. The residuals are computed via Gauss quadrature within each voxel, capturing deviations from mass conservation across the domain:

    \[
    M4 = \sqrt{\sum_{i=1}^{N} (\partial_j \hat{u}_i^{(j)})^2}
    \]

    where \(\sum_{i=1}^{N} (\partial_j \hat{u}_i^{(j)})\) is the divergence of the velocity field at element \(i\).
\end{itemize}

These metrics evaluate model performance, from global flow field accuracy (\(M1\)) to near-boundary precision (\(M2\)), velocity gradient consistency (\(M3\)), and mass conservation enforcement (\(M4\)). The scores for \(M1\), \(M2\), and \(M3\) are computed based on the relative \(L_2\) error of the predicted velocity fields (or their gradients), evaluated at the element centers (see \appendixref{sec:FEM}). A score of 100 indicates perfect prediction, while lower scores indicate larger errors.

\subsection{Train-Test Split} \label{subsec:train-test-split}

We adopt two distinct train-test splitting strategies to evaluate the model’s predictive performance: random and extrapolatory.

In the random split, the training and test sets are drawn randomly, ensuring a diverse distribution of Reynolds numbers across both sets. This serves as a baseline for evaluating the model’s ability to learn from a dataset that spans the full range of Reynolds numbers. In contrast, the extrapolatory split is designed to assess the model’s ability to generalize to unseen flow conditions. Here, the training set consists of flow solutions corresponding to the lower 80\% of the Reynolds number range, while the test set includes cases from the highest 20\%. This split tests the model's ability to predict flow behavior at Reynolds numbers beyond its training range.

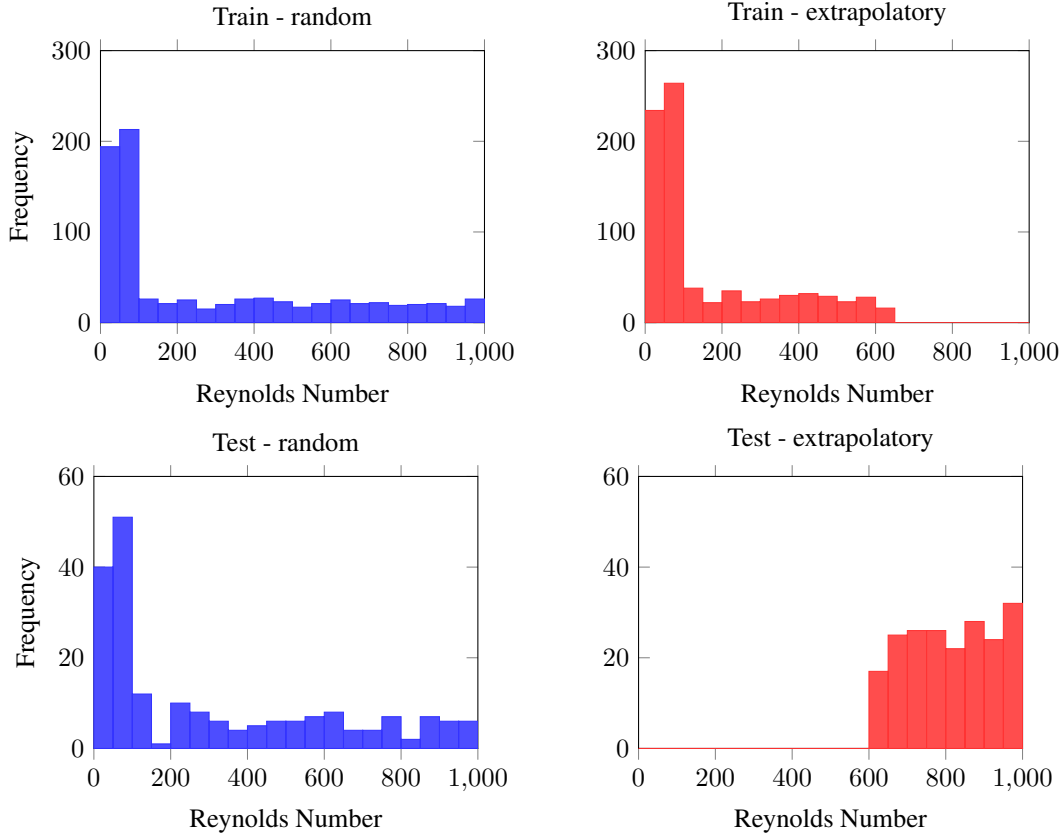
\begin{figure} [!ht]
    \centering
    % Top row: Train Data
    \begin{subfigure}{0.45\linewidth}
        \centering
        \begin{tikzpicture}
            \begin{axis}[
                width=0.9\linewidth,
                height=0.7\linewidth,
                ybar,
                xlabel={Reynolds Number},
                ylabel={Frequency},
                title={Train - random},
                ymin=0, ymax=300,
                xmin=0, xmax=1000,
                bar width=5pt,
                xticklabel style={/pgf/number format/.cd, fixed, precision=0},
                yticklabel style={/pgf/number format/.cd, fixed, precision=0}
            ]
                \addplot+[
                    hist={bins=20},
                    fill=blue!70,
                    draw=blue!80
                ] table [y=train_easy, col sep=comma] {Figures/train-test/train_reynolds.csv};
            \end{axis}
        \end{tikzpicture}
    \end{subfigure}
    \begin{subfigure}{0.45\linewidth}
        \centering
        \begin{tikzpicture}
            \begin{axis}[
                width=0.9\linewidth,
                height=0.7\linewidth,
                ybar,
                xlabel={Reynolds Number},
                title={Train - extrapolatory},
                ymin=0, ymax=300,
                xmin=0, xmax=1000,
                bar width=5pt,
                xticklabel style={/pgf/number format/.cd, fixed, precision=0},
                yticklabel style={/pgf/number format/.cd, fixed, precision=0}
            ]
                \addplot+[
                    hist={bins=20},
                    fill=red!70,
                    draw=red!80
                ] table [y=train_hard, col sep=comma] {Figures/train-test/train_reynolds.csv};
            \end{axis}
        \end{tikzpicture}
    \end{subfigure}
    % Bottom row: Test Data
    \begin{subfigure}{0.45\linewidth}
        \centering
        \begin{tikzpicture}
            \begin{axis}[
                width=0.9\linewidth,
                height=0.7\linewidth,
                ybar,
                xlabel={Reynolds Number},
                ylabel={Frequency},
                title={Test - random},
                ymin=0, ymax=60,
                xmin=0, xmax=1000,
                bar width=5pt,
                xticklabel style={/pgf/number format/.cd, fixed, precision=0},
                yticklabel style={/pgf/number format/.cd, fixed, precision=0}
            ]
                \addplot+[
                    hist={bins=20},
                    fill=blue!70,
                    draw=blue!80
                ] table [y=test_easy, col sep=comma] {Figures/train-test/test_reynolds.csv};
            \end{axis}
        \end{tikzpicture}
    \end{subfigure}
    \begin{subfigure}{0.45\linewidth}
        \centering
        \begin{tikzpicture}
            \begin{axis}[
                width=0.9\linewidth,
                height=0.7\linewidth,
                ybar,
                xlabel={Reynolds Number},
                title={Test - extrapolatory},
                ymin=0,  ymax=60,
                xmin=0, xmax=1000,
                bar width=5pt,
                xticklabel style={/pgf/number format/.cd, fixed, precision=0},
                yticklabel style={/pgf/number format/.cd, fixed, precision=0}
            ]
                \addplot+[
                    hist={bins=20},
                    fill=red!70,
                    draw=red!80
                ] table [y=test_hard, col sep=comma] {Figures/train-test/test_reynolds.csv};
            \end{axis}
        \end{tikzpicture}
    \end{subfigure}
    \caption{Distribution of Reynolds numbers across different train-test splitting strategies. The left column represents the easy case, while the right column corresponds to the hard case. The top row shows the training dataset, and the bottom row shows the test dataset. The baseline random split ensures a uniform distribution of Reynolds numbers in both train and test sets, while the extrapolatory split reserves the highest 20\% of Reynolds numbers for testing, with the training set containing only the lower 80\%.}
    \label{fig:reynolds_numbers_comparison}
\end{figure}

\figref{fig:reynolds_numbers_comparison} provides a comparison of the Reynolds number distributions for both strategies. Models are trained using both splits to examine their robustness in interpolation versus extrapolation scenarios.

\section{Loss Function Design} \label{sec:loss-functions}

\paragraph{Loss Function Design:}  
We explore four distinct loss functions—denoted \(L1\) through \(L4\)—that capture various aspects of predictive performance. All losses are evaluated at the element centers using computational analogous to those used in the Finite Element Method (FEM) as explained in \appendixref{sec:FEM}. The loss is only computed in regions where the signed distance field satisfies \(\text{SDF} > 0\), ensuring that computations are restricted to the physical domain external to the embedded geometry. 

Some of our loss functions utilize the velocity gradient. The velocity gradient tensor is defined as
\[
\partial_j u^i =
\begin{bmatrix}
\frac{\partial u}{\partial x} & \frac{\partial u}{\partial y} & \frac{\partial u}{\partial z} \\
\frac{\partial v}{\partial x} & \frac{\partial v}{\partial y} & \frac{\partial v}{\partial z} \\
\frac{\partial w}{\partial x} & \frac{\partial w}{\partial y} & \frac{\partial w}{\partial z}
\end{bmatrix},
\]
which encapsulates all directional derivatives \(\partial_j u^i\) 
for \(i = 1,2,3\) and \(j = 1,2,3\). Here, we adopt Einstein notation of the velocity components with \(u^1 = u\), \(u^2 = v\), and \(u^3 = w\). The velocity gradient is the mathematical construct that undergirds several physical properties representing the underlying dynamics. Our hypothesis was that incorporating the gradient in the loss function could improve model performance.

The loss functions, per sample, are defined as follows:
\begin{itemize}
    \item \textbf{\(L1\): Relative MSE Loss} --- This loss focuses on the velocity field itself, computing a mean squared error (MSE) that is normalized per velocity component (\(u\), \(v\), \(w\)). It penalizes discrepancies in the velocity components weighted by \(\lambda_u\), \(\lambda_v\), and \(\lambda_w\). In our formulation, the loss is computed as a sum over the Gauss points in the fluid domain:
    \[
    L1 = \frac{1}{N_{\text{out}}} \sum_{i \in \Omega_f}\Bigl[ \lambda_u\Bigl(u_i - u_{\text{true},i}\Bigr)^2 + \lambda_v\Bigl(v_i - v_{\text{true},i}\Bigr)^2 + \lambda_w\Bigl(w_i - w_{\text{true},i}\Bigr)^2 \Bigr],
    \]
    
    \item \textbf{\(L2\): Relative Derivative MSE Loss} --- Extending the approach of \(L1\), \(L2\) incorporates additional terms that account for the spatial derivatives of the velocity field. The derivative components, such as \(\frac{\partial u}{\partial x}\), \(\frac{\partial v}{\partial y}\), etc., are penalized using weights \(\lambda_{u_x}\), \(\lambda_{u_y}\), \(\lambda_{u_z}\) for the \(u\) component, and similarly for \(v\) and \(w\) using \(\lambda_{v_x}\), \(\lambda_{v_y}\), \(\lambda_{v_z}\) and \(\lambda_{w_x}\), \(\lambda_{w_y}\), \(\lambda_{w_z}\) respectively. The loss is defined as:
    \[
    \begin{aligned}
    L2 &= L1 + \frac{1}{N_{\text{out}}} \sum_{i \in \Omega_f} h \times \Biggl[
    \lambda_{u_x}\Bigl(\frac{\partial u}{\partial x}_i - \frac{\partial u_{\text{true}}}{\partial x}_i\Bigr)^2
    + \lambda_{u_y}\Bigl(\frac{\partial u}{\partial y}_i - \frac{\partial u_{\text{true}}}{\partial y}_i\Bigr)^2
    + \lambda_{u_z}\Bigl(\frac{\partial u}{\partial z}_i - \frac{\partial u_{\text{true}}}{\partial z}_i\Bigr)^2 \\
    &\quad + \lambda_{v_x}\Bigl(\frac{\partial v}{\partial x}_i - \frac{\partial v_{\text{true}}}{\partial x}_i\Bigr)^2
    + \lambda_{v_y}\Bigl(\frac{\partial v}{\partial y}_i - \frac{\partial v_{\text{true}}}{\partial y}_i\Bigr)^2
    + \lambda_{v_z}\Bigl(\frac{\partial v}{\partial z}_i - \frac{\partial v_{\text{true}}}{\partial z}_i\Bigr)^2 \\
    &\quad + \lambda_{w_x}\Bigl(\frac{\partial w}{\partial x}_i - \frac{\partial w_{\text{true}}}{\partial x}_i\Bigr)^2
    + \lambda_{w_y}\Bigl(\frac{\partial w}{\partial y}_i - \frac{\partial w_{\text{true}}}{\partial y}_i\Bigr)^2
    + \lambda_{w_z}\Bigl(\frac{\partial w}{\partial z}_i - \frac{\partial w_{\text{true}}}{\partial z}_i\Bigr)^2
    \Biggr],
    \end{aligned}
    \]
    
    \item \textbf{\(L3\): Pure Derivative + Boundary Loss} --- Unlike \(L1\) and \(L2\), \(L3\) focuses exclusively on the derivatives of the velocity field. In addition to the derivative terms weighted as in \(L2\), a boundary loss term weighted by \(\lambda_{\text{boundary}}\) is included to enforce consistency along the domain boundaries, thereby ensuring the uniqueness of the prediction. Here, \(\Gamma_{x_{\min}}\), \(\Gamma_{x_{\max}}\), \(\Gamma_{y_{\min}}\), \(\Gamma_{y_{\max}}\), \(\Gamma_{z_{\min}}\), and \(\Gamma_{z_{\max}}\) represent the sets of nodal points on the boundary planes \(x=0\), \(x=\mathrm{x_{max}}\), \(y=0\), \(y=\mathrm{y_{max}}\), \(z=0\), and \(z=\mathrm{z_{max}}\), respectively. This enforcement is applied at nodal points on the six outer faces enclosing the computational domain:
    \[
    \begin{aligned}
    L3 &= \frac{1}{N_{\text{out}}} \sum_{i \in \Omega_f} h \times \Biggl[
    \lambda_{u_x}\Bigl(\frac{\partial u}{\partial x}_i - \frac{\partial u_{\text{true}}}{\partial x}_i\Bigr)^2
    + \lambda_{u_y}\Bigl(\frac{\partial u}{\partial y}_i - \frac{\partial u_{\text{true}}}{\partial y}_i\Bigr)^2
    + \lambda_{u_z}\Bigl(\frac{\partial u}{\partial z}_i - \frac{\partial u_{\text{true}}}{\partial z}_i\Bigr)^2 \\
    &\quad + \lambda_{v_x}\Bigl(\frac{\partial v}{\partial x}_i - \frac{\partial v_{\text{true}}}{\partial x}_i\Bigr)^2
    + \lambda_{v_y}\Bigl(\frac{\partial v}{\partial y}_i - \frac{\partial v_{\text{true}}}{\partial y}_i\Bigr)^2
    + \lambda_{v_z}\Bigl(\frac{\partial v}{\partial z}_i - \frac{\partial v_{\text{true}}}{\partial z}_i\Bigr)^2 \\
    &\quad + \lambda_{w_x}\Bigl(\frac{\partial w}{\partial x}_i - \frac{\partial w_{\text{true}}}{\partial x}_i\Bigr)^2
    + \lambda_{w_y}\Bigl(\frac{\partial w}{\partial y}_i - \frac{\partial w_{\text{true}}}{\partial y}_i\Bigr)^2
    + \lambda_{w_z}\Bigl(\frac{\partial w}{\partial z}_i - \frac{\partial w_{\text{true}}}{\partial z}_i\Bigr)^2
    \Biggr] \\
    &\quad + \lambda_{\text{boundary}}\Biggl[
    \sum_{j \in \Gamma_{x_{\min}}}\Bigl(u_j - u_{\text{true},j}\Bigr)^2 
    + \sum_{j \in \Gamma_{x_{\max}}}\Bigl(u_j - u_{\text{true},j}\Bigr)^2 \\
    &\quad\quad + \sum_{j \in \Gamma_{y_{\min}}}\Bigl(v_j - v_{\text{true},j}\Bigr)^2 
    + \sum_{j \in \Gamma_{y_{\max}}}\Bigl(v_j - v_{\text{true},j}\Bigr)^2 \\
    &\quad\quad + \sum_{j \in \Gamma_{z_{\min}}}\Bigl(w_j - w_{\text{true},j}\Bigr)^2 
    + \sum_{j \in \Gamma_{z_{\max}}}\Bigl(w_j - w_{\text{true},j}\Bigr)^2
    \Biggr].
    \end{aligned}
    \]
    
    \item \textbf{\(L4\): Pure Derivative + Boundary + Physics-Informed Loss} --- The most comprehensive of the four, \(L4\) combines the derivative and boundary loss components in \(L3\) with a dedicated term that enforces the physical constraint of mass conservation. Specifically, an additional loss term penalizes deviations from the mass continuity equation (i.e., the solenoidality condition \(\partial_j u_j = \frac{\partial u}{\partial x} + \frac{\partial v}{\partial y} + \frac{\partial w}{\partial z} = 0\)), and is weighted by \(\lambda_{\text{solenoidality}}\). The loss is given by:
    \[
    L4 = L3 + \frac{1}{N_{\text{out}}} \lambda_{\text{solenoidality}} \sum_{i \in \Omega_f} h \times \Bigl(\frac{\partial u}{\partial x}_i + \frac{\partial v}{\partial y}_i + \frac{\partial w}{\partial z}_i\Bigr)^2.
    \]
\end{itemize}

The weights \(\lambda_i\) are chosen so that all components contribute equally to the overall loss. We set these values by examining the average magnitudes of velocity and gradient components in the CFD data\footnote{Additionally, the solenoidality weight accommodates the fact that element-wise mass conservation is not perfectly satisfied in the underlying CFD simulations, introducing some numerical error (continuity residuals)}. The values of the weights are as follows:
\begin{itemize}
    \item For the velocity components:
    \[
    \lambda_u = 1,\quad \lambda_v = 3,\quad \lambda_w = 150.
    \]
    \item For the spatial derivatives:
    \[
    \lambda_{u_x} = 15,\quad \lambda_{u_y} = 1,\quad \lambda_{u_z} = 30,
    \]
    \[
    \lambda_{v_x} = 50,\quad \lambda_{v_y} = 30,\quad \lambda_{v_z} = 5,
    \]
    \[
    \lambda_{w_x} = 600,\quad \lambda_{w_y} = 750,\quad \lambda_{w_z} = 600.
    \]
    \item For the boundary consistency term:
    \[
    \lambda_{\text{boundary}} = 5.
    \]
    \item For the solenoidality (mass conservation) term:
    \[
    \lambda_{\text{solenoidality}} = 10.
    \]
\end{itemize}
We note that practitioners can also employ systematic approaches like grid search or Bayesian optimization to adapt the weights for best loss design in other applications. 

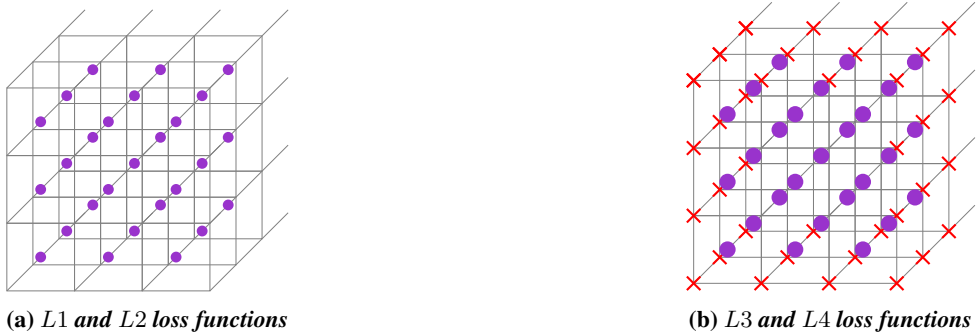
\begin{figure}[!ht]
    \centering
    \captionsetup[subfigure]{justification=centering} % Ensure subcaptions are centered
    \begin{subfigure}[t]{0.45\textwidth}
        \centering
        \begin{tikzpicture}[scale=0.9]
            % FRONT FACE (z = 0)
            \foreach \x in {0,1,2}
                \foreach \y in {0,1,2}
                    \draw[gray, very thin] (\x,\y,0) -- ++(1,0,0) -- ++(0,1,0) -- ++(-1,0,0) -- cycle;

            % MIDDLE FACE (z = -1)
            \foreach \x in {0,1,2}
                \foreach \y in {0,1,2}
                    \draw[gray, very thin] (\x,\y,-1) -- ++(1,0,0) -- ++(0,1,0) -- ++(-1,0,0) -- cycle;

            % BACK FACE (z = -2)
            \foreach \x in {0,1,2}
                \foreach \y in {0,1,2}
                    \draw[gray, very thin] (\x,\y,-2) -- ++(1,0,0) -- ++(0,1,0) -- ++(-1,0,0) -- cycle;

            % CONNECTING LINES (Depth Representation)
            \foreach \x in {0,1,2,3}
                \foreach \y in {0,1,2,3}
                    \foreach \z in {0,-1,-2}
                        \draw[gray, very thin] (\x,\y,\z) -- (\x,\y,\z-1);

            % Velocity Dots (L1) at Element Centers (Black Dots)
            \foreach \x in {0.5,1.5,2.5}
                \foreach \y in {0.5,1.5,2.5}
                    \foreach \z in {0,-1,-2} % Place at all three slices
                        \fill[boundaryColor] (\x,\y,\z) circle (0.08);

        \end{tikzpicture}
        \caption{\textbf{\(L1\) and \(L2\) loss functions}}
        \label{fig:L1-L2}
    \end{subfigure}%
    \hfill
    \begin{subfigure}[t]{0.45\textwidth}
        \centering
        \begin{tikzpicture}[scale=0.9]
            % FRONT FACE (z = 0)
            \foreach \x in {0,1,2}
                \foreach \y in {0,1,2}
                    \draw[gray, very thin] (\x,\y,0) -- ++(1,0,0) -- ++(0,1,0) -- ++(-1,0,0) -- cycle;

            % MIDDLE FACE (z = -1)
            \foreach \x in {0,1,2}
                \foreach \y in {0,1,2}
                    \draw[gray, very thin] (\x,\y,-1) -- ++(1,0,0) -- ++(0,1,0) -- ++(-1,0,0) -- cycle;

            % BACK FACE (z = -2)
            \foreach \x in {0,1,2}
                \foreach \y in {0,1,2}
                    \draw[gray, very thin] (\x,\y,-2) -- ++(1,0,0) -- ++(0,1,0) -- ++(-1,0,0) -- cycle;

            % CONNECTING LINES (Depth Representation)
            \foreach \x in {0,1,2,3}
                \foreach \y in {0,1,2,3}
                    \foreach \z in {0,-1,-2}
                        \draw[gray, very thin] (\x,\y,\z) -- (\x,\y,\z-1);

            % Show gradient (Red X) at boundary points
            \foreach \x/\y in {0/0, 1/0, 2/0, 3/0, 3/1, 3/2, 3/3, 0/3, 1/3, 2/3, 0/1, 0/2, 0/3}
                \foreach \z in {0,-1,-2}
                {
                    \draw[gradColor, thick] (\x-0.1,\y-0.1,\z) -- ++(0.2,0.2,0);
                    \draw[gradColor, thick] (\x+0.1,\y-0.1,\z) -- ++(-0.2,0.2,0);
                }
            
            % Show velocity (Black Dot) at element centers
            \foreach \x in {0.5,1.5,2.5}
                \foreach \y in {0.5,1.5,2.5}
                {
                    \foreach \z in {0,-1,-2}
                        \fill[boundaryColor] (\x,\y,\z) circle (0.12);
                }
        \end{tikzpicture}
        \caption{\textbf{\(L3\) and \(L4\) loss functions}}
        \label{fig:L3-L4}
    \end{subfigure}
    \caption{Visualization of the loss functions in a 3×3×3 grid. \textbf{\(L1\)} and \textbf{\(L2\)} evaluate errors at element centers, while \textbf{\(L3\)} and \textbf{\(L4\)} enforce velocity consistency at boundary points while keeping gradient losses at element centers.}
    \label{fig:loss-voxel}
\end{figure}

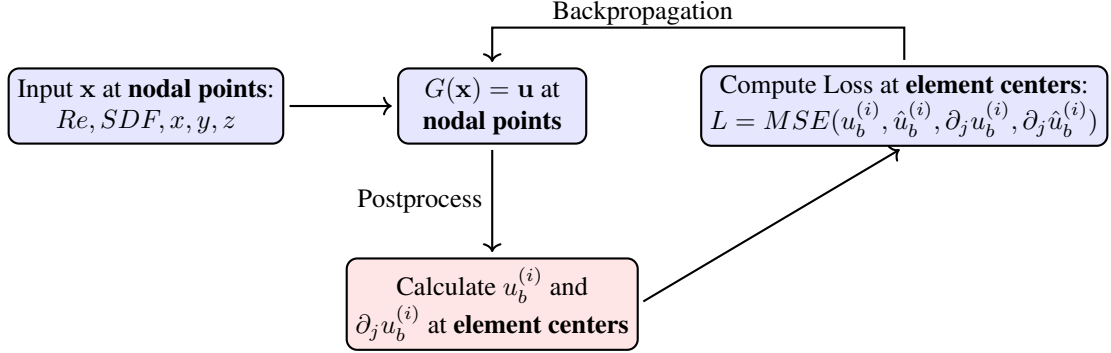
\begin{figure}[!ht]
    \centering
    \begin{tikzpicture}[
        node distance=1.5cm,
        block/.style={rectangle, draw, thick, rounded corners, align=center, fill=blue!10, minimum width=2.5cm, minimum height=1cm},
        arrow/.style={->, thick, shorten >=2pt, shorten <=2pt},
        curvedarrow/.style={->, thick, shorten >=2pt, shorten <=2pt, bend left=20},
        curvedarrow2/.style={->, thick, shorten >=2pt, shorten <=2pt, bend right=20},
        postproc/.style={rectangle, draw, thick, rounded corners, align=center, fill=red!10, minimum width=3cm, minimum height=1cm},
        loss/.style={rectangle, draw, thick, rounded corners, align=center, fill=blue!10, minimum width=4cm, minimum height=1cm}
    ]
    
    % Input
    \node[block] (input) {Input $\mathbf{x}$ at \textbf{nodal points}: \\ $Re, SDF, x, y, z$};
    
    % Neural Network
    \node[block, right=1.5cm of input] (nn) {$G(\mathbf{x}) = \mathbf{u}$ at \\ \textbf{nodal points}};

    % Postprocessing (Moved to second row)
    \node[postproc, below=1.5cm of nn] (postproc) {Calculate $u_b^{(i)}$ and \\ $\partial_j u_b^{(i)}$ at \textbf{element centers}};
    
    % Loss Evaluation
    \node[loss, right=1.5cm of nn] (loss) {Compute Loss at \textbf{element centers}: \\ $L = MSE(u_b^{(i)}, \hat{u}_b^{(i)}, \partial_j u_b^{(i)}, \partial_j \hat{u}_b^{(i)})$};
    
    % Arrows with Better Placement
    \draw[arrow] (input.east) -- (nn.west);
    \draw[arrow] (nn.south) to node[left] {Postprocess} (postproc.north);
    \draw[arrow] (postproc.east) to node[above, xshift=-25pt] {} (loss.south); %{Compute Loss}
    \draw[arrow] (loss.north) -- ++(0,0.5) -| node[midway, anchor=center, xshift=20mm, yshift=2mm] {Backpropagation} (nn.north);
    \end{tikzpicture}
    \caption{Neural network function $G(\mathbf{x})$ predicts velocity components ($u^{(i)} = \{u, v, w\}$) at nodal points, then postprocessing computes velocity ($u_b^{(i)}$) and gradients (\(\partial_j u_b^{(i)}\)) at element centers before evaluating the loss functions. The loss is backpropagated to update the network parameters.}
    \label{fig:nn-postproc-loss}
\end{figure}

Each loss function is computed by first applying a mask that selects only the elements outside the geometry satisfying \(\text{SDF} > 0\) (\(\Omega_f\)), and summed over all training samples. This conditioning guarantees that the loss is evaluated exclusively in the relevant physical domain and is scaled by $\frac{1}{N_{\text{out}}}$ where $N_{\text{out}}$ denotes the total number of points satisfying \(\text{SDF} > 0\). \figref{fig:loss-voxel} provides a visual representation of the locations where velocity and gradient errors are computed in different loss functions, and \appendixref{sec:FEM} explains the element-wise evaluation of velocity values, gradients, and continuity equation residuals. Since our neural network predicts velocity components at nodal points, additional postprocessing is required to obtain the velocity and its gradients at element (or voxel) centers before evaluating the loss. By computing gradients at element (or voxel) centers, we align the discretization with local shape function properties, thereby capturing local velocity variations more accurately, which is particularly useful near complex boundaries. \figref{fig:nn-postproc-loss} illustrates this process, showing how the model first predicts nodal velocities, followed by postprocessing to compute element-centered values, which are then used to evaluate the loss.

By tuning the \(\lambda_i\) coefficients, we ensure that the contribution of each term is balanced, promoting an optimization process that uniformly improves both the field values and their spatial derivatives. Furthermore, the loss terms involving spatial derivatives are scaled by the mesh size, $h$, to account for numerical discretization effects across different grid resolutions. In finite element formulations, derivatives are approximated as the element-wise differences in velocity components divided by the grid spacing, $h$, making them inherently proportional to $\frac{1}{h}$. This adjustment ensures that the loss contributions remain balanced regardless of the discretization level, preserving the accuracy of velocity gradients while facilitating stable and efficient training. The impact of these loss functions on model training \emph{DeepONet} and \emph{Geometric-DeepONet} is reported in \figref{fig:deeponet-geo-loss} in \appendixref{subsec:loss-plots}, where we present the training and validation loss curves for \(L1\) through \(L4\). We observe that while the training loss decreases steadily for all loss functions, the validation loss tends to plateau or even slightly increase indicating potential overfitting in some configurations. Notably, \emph{Geometric-DeepONet} shows a faster and more consistent reduction in both training and validation errors relative to \emph{DeepONet}. Furthermore, the \emph{DeepONet} model trained with \(L3\) and \(L4\), which include derivative and physics-informed loss terms, exhibit more stable convergence on the validation set compared to those trained with loss functions that rely on velocity field values (\(L1\) and \(L2\)). This illustrates that incorporating velocity gradient information, rather than field values, not only improves model accuracy but also enhances generalization and mitigates overfitting.

\section{Results} \label{sec:results}

\tabref{tab:metrics-comparison} summarize the performance of \emph{DeepONet} and \emph{Geometric-DeepONet} under random and extrapolatory train-test splits, evaluated using four distinct loss functions (\(L1\)–\(L4\)). The results highlight that \emph{Geometric-DeepONet} using pure derivative loss (\(L3\)) outperforms other models.

\begin{table}[!ht] 
\centering
\small
\setlength\extrarowheight{2pt}
\caption{Performance scores of \emph{DeepONet} and \emph{Geometric-DeepONet} across evaluation metrics \(M1\text{--}M4\), using random and extrapolatory train-test splits with four loss functions \(L1\text{--}L4\).}
\label{tab:metrics-comparison}
\begin{tabular}{c c | c c c c | c c c c}
\multicolumn{2}{c}{} & \multicolumn{4}{c}{\textbf{Random}} & \multicolumn{4}{c}{\textbf{Extrapolatory}} \\  
\hline
\textbf{Loss Function} & \textbf{Model} & \textbf{\(M1\)} & \textbf{\(M2\)} & \textbf{\(M3\)} & \textbf{\(M4\)} & \textbf{\(M1\)} & \textbf{\(M2\)} & \textbf{\(M3\)} & \textbf{\(M4\)} \\ \hline
\textbf{\(L1\)} & \textbf{Geometric-Deeponet} & $91.31$ & $81.52$ & $56.31$ & $3.22 \times 10^{-3}$ & $67.47$ & $43.56$ & $15.30$ & $3.78 \times 10^{-3}$ \\  
& \textbf{Deeponet} & $83.15$ & $55.03$ & $32.83$ & $3.33 \times 10^{-3}$ & $74.35$ & $33.85$ & $25.12$ & $3.40 \times 10^{-3}$ \\ \hline
\textbf{\(L2\)} & \textbf{Geometric-Deeponet} & $93.71$ & $82.40$ & $78.63$ & $3.74 \times 10^{-3}$ & $81.66$ & $65.81$ & $\mathbf{60.91}$ & $3.56 \times 10^{-3}$ \\  
& \textbf{Deeponet} & $84.51$ & $55.48$ & $63.67$ & $3.62 \times 10^{-3}$ & $77.00$ & $37.03$ & $54.16$ & $3.76 \times 10^{-3}$ \\ \hline
\textbf{\(L3\)} & \textbf{Geometric-Deeponet} & $\mathbf{94.22}$ & $\mathbf{84.86}$ & $\mathbf{81.72}$ & $3.75 \times 10^{-3}$ & $\mathbf{82.19}$ & $\mathbf{67.38}$ & $60.72$ & $3.57 \times 10^{-3}$ \\  
& \textbf{Deeponet} & $81.60$ & $58.27$ & $62.60$ & $3.74 \times 10^{-3}$ & $71.20$ & $24.27$ & $55.51$ & $3.66 \times 10^{-3}$ \\ \hline
\textbf{\(L4\)} & \textbf{Geometric-Deeponet} & $85.77$ & $72.10$ & $75.51$ & $2.54 \times 10^{-3}$ & $76.88$ & $62.25$ & $57.18$ & $2.52 \times 10^{-3}$ \\  
& \textbf{Deeponet} & $71.89$ & $36.34$ & $55.59$ & $\mathbf{2.41 \times 10^{-3}}$ & $62.11$ & $19.22$ & $52.47$ & $\mathbf{2.49 \times 10^{-3}}$ \\ \hline
\end{tabular}
\end{table}
% \figref{fig:depo-vs-geo} presents a sample prediction (cube geometry) of velocity components and gradients \(\left(\sqrt{\sum_{i} \sum_{j} \left(\frac{\partial u_i}{\partial x_j} \right)^2}\right)\) in the \(XZ\) middle plane using the \(L1\) loss function. \emph{Geometric-DeepONet} demonstrates improved accuracy in capturing velocity gradients and finer-scale velocity variations in the z-direction (\(w\)) compared to \emph{DeepONet}. The difference in prediction quality is most evident in the fourth row, where the velocity gradients produced by \emph{Geometric-DeepONet} more closely match the ground truth. Furthermore, \figref{fig:depo-vs-geo-streamlines} illustrates the streamlines of the predicted velocity fields around the geometry, further highlighting the accurate flow structure representation achieved by \emph{Geometric-DeepONet}. Additionally, \figref{fig:depo-vs-geo-xy} and \figref{fig:depo-vs-geo-yz} in~\appendixref{sec:field-predictions} provide a comparative analysis of \emph{DeepONet} and \emph{Geometric-DeepONet} predictions in the \(XY\) and \(YZ\) middle planes, respectively, using the \(L1\) loss function. The improved performance of \emph{Geometric-DeepONet} is consistent across different cross-sections, further validating its robustness.

\begin{figure}[!ht]
    \centering
    % First row: Ground Truth images
    \begin{minipage}[b]{0.45\linewidth}
        \centering
        \includegraphics[width=\linewidth]{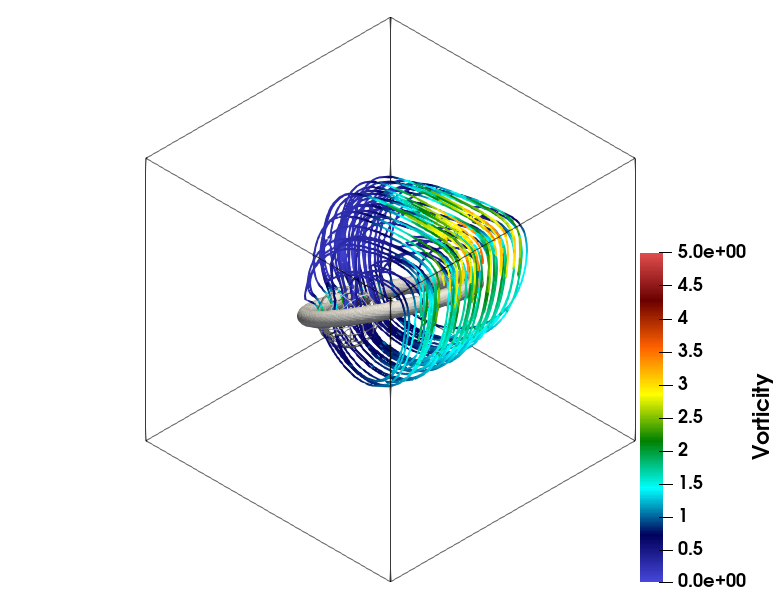}
        \textbf{Ground Truth (Ring)}
    \end{minipage}
    \hfill
    \begin{minipage}[b]{0.45\linewidth}
        \centering
        \includegraphics[width=\linewidth]{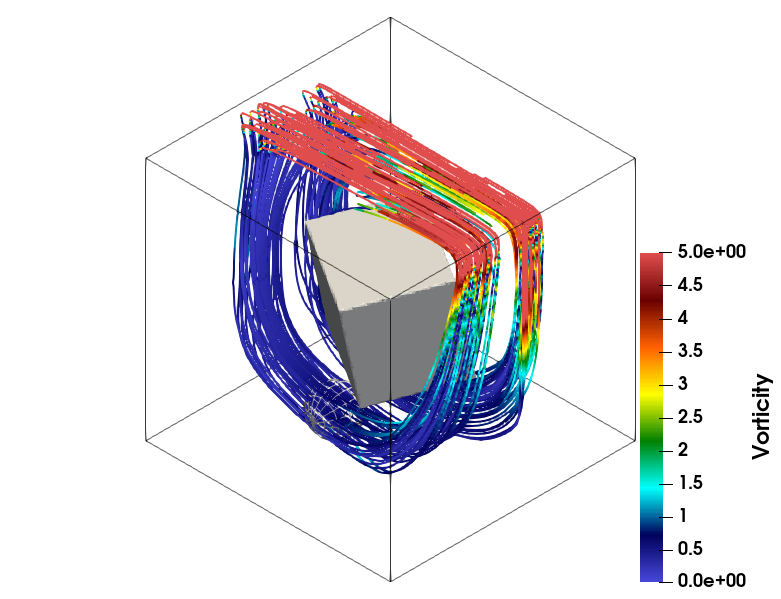}
        \textbf{Ground Truth (Cube)}
    \end{minipage}
    
    \vspace{1em} % vertical space between rows
    
    % Second row: DeepONet images
    \begin{minipage}[b]{0.45\linewidth}
        \centering
        \includegraphics[width=\linewidth]{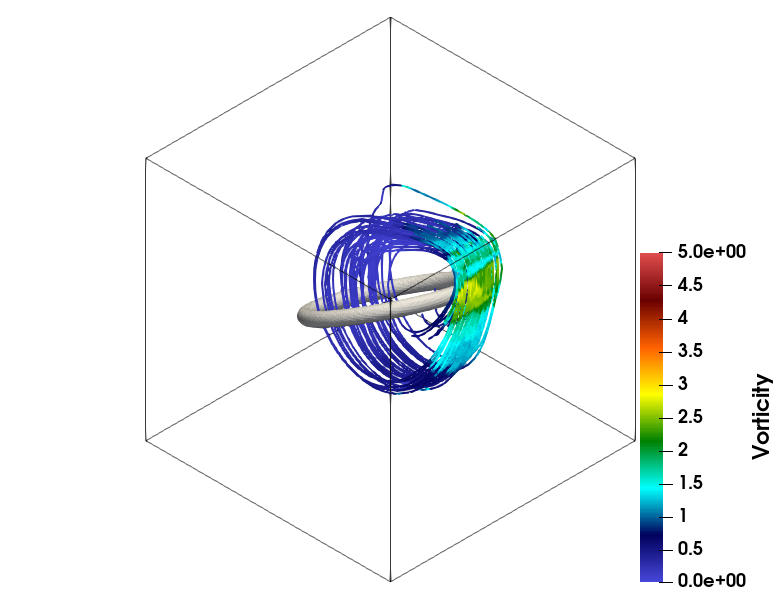}
        \textbf{DeepONet (Ring)}
    \end{minipage}
    \hfill
    \begin{minipage}[b]{0.45\linewidth}
        \centering
        \includegraphics[width=\linewidth]{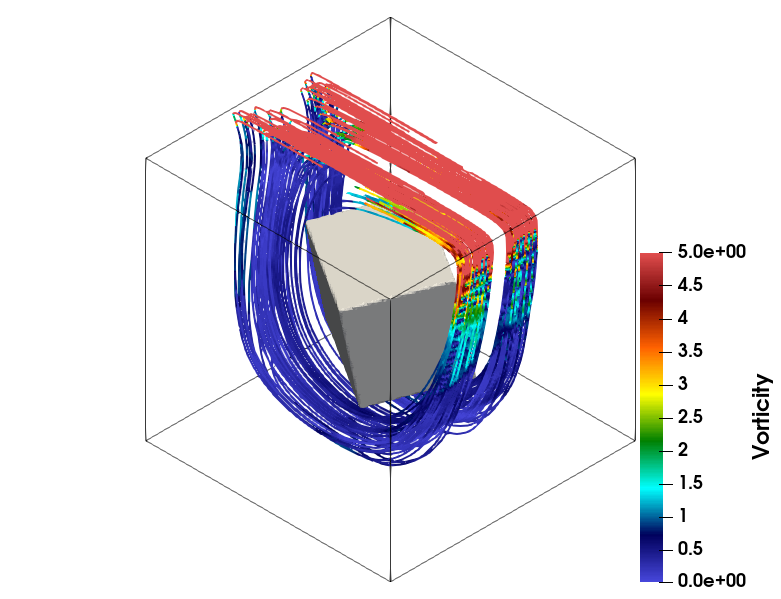}
        \textbf{DeepONet (Cube)}
    \end{minipage}
    
    \vspace{1em} % vertical space between rows
    
    % Third row: Geometric-DeepONet images
    \begin{minipage}[b]{0.45\linewidth}
        \centering
        \includegraphics[width=\linewidth]{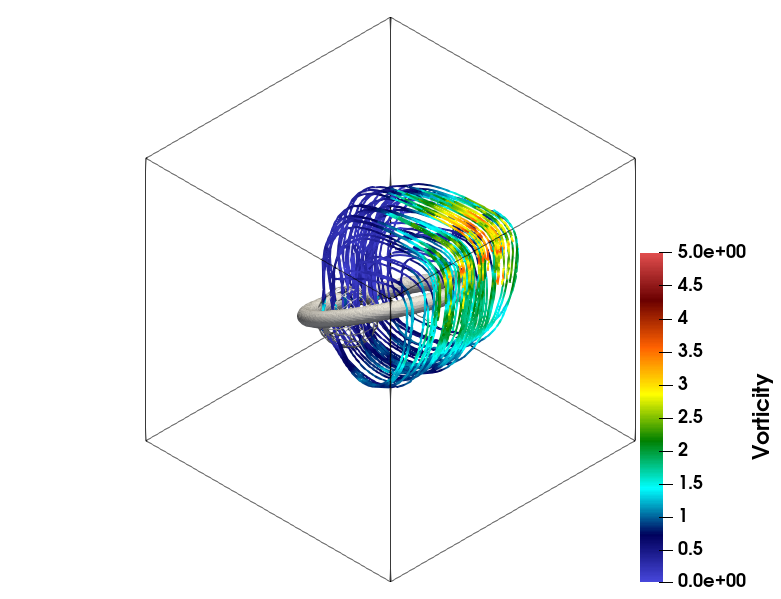}
        \textbf{Geometric-DeepONet (Ring)}
    \end{minipage}
    \hfill
    \begin{minipage}[b]{0.45\linewidth}
        \centering
        \includegraphics[width=\linewidth]{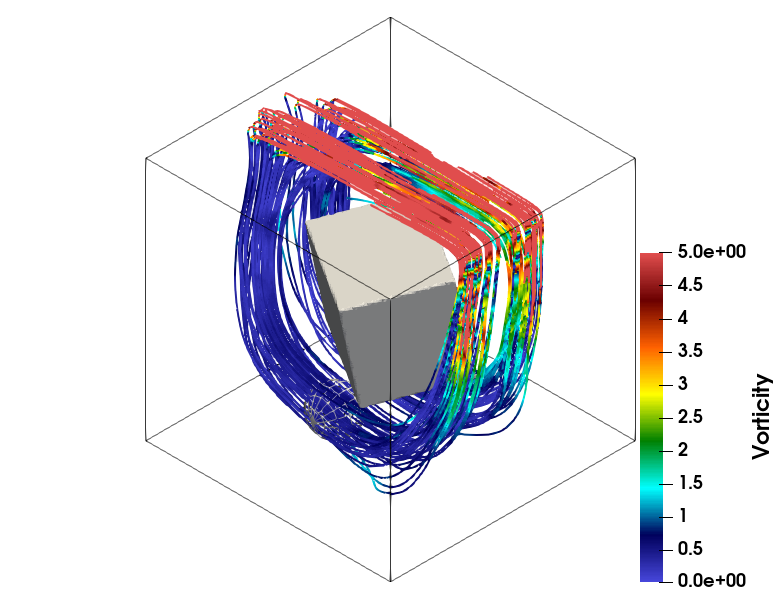}
        \textbf{Geometric-DeepONet (Cube)}
    \end{minipage}
    
    \caption{Streamline comparison of \emph{DeepONet} and \emph{Geometric-DeepONet} for flow around two representative geometries using the \(L1\) loss function. The first row shows the ground truth, the second row shows DeepONet predictions, and the third row shows Geometric-DeepONet predictions. The left column corresponds to the ring geometry, while the right column corresponds to the cube geometry.}
    \label{fig:depo-vs-geo-streamlines} 
\end{figure}

\figref{fig:depo-vs-geo-streamlines} illustrates the streamlines of the predicted velocity fields around two representative geometries (ring and cube) using the standard \emph{DeepONet} and \emph{Geometric-DeepONet}. The streamlines around the geometry of the ring highlight the accurate representation of the flow streamlines achieved by \emph{Geometric-DeepONet}.

\subsection{Impact of Loss Function Design} \label{subsec:loss-function-design}

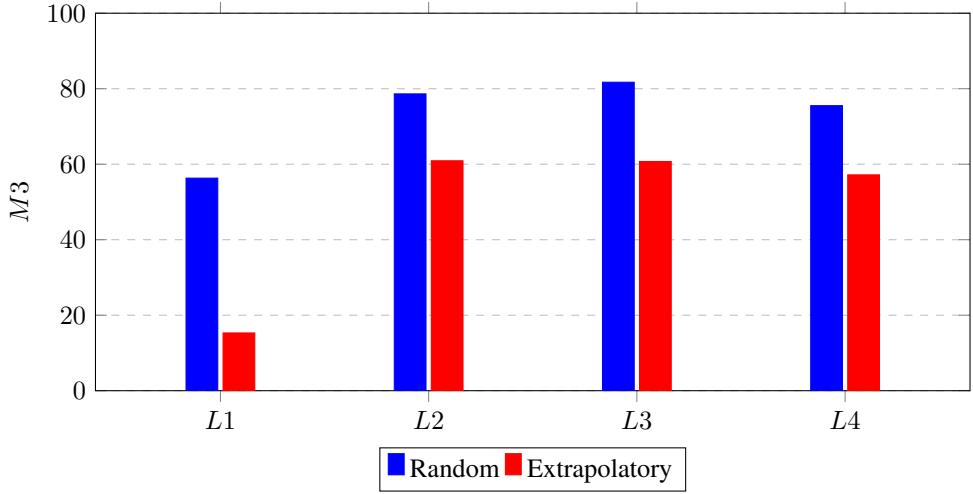
\begin{figure}[!ht]
    \centering
    \begin{tikzpicture}
        \begin{axis}[
            ybar,
            bar width=12pt,
            width=0.8\linewidth,
            height=0.4\linewidth,
            enlarge x limits=0.2,  % Increase spacing between bars
            symbolic x coords={\(L1\), \(L2\), \(L3\), \(L4\)},
            xtick=data,
            xtick style={align=center},
            ylabel={\text{\(M3\)}},
            legend style={at={(0.5,-0.15)},anchor=north,legend columns=-1},
            ymin=0,
            ymax=100,
            ymajorgrids=true,
            grid style=dashed
        ]
            % Plot Random scores for Geometric-DeepONet (M3)
            \addplot+[ybar, fill=blue] coordinates {
                (\(L1\), 56.31) (\(L2\), 78.63) (\(L3\), 81.72) (\(L4\), 75.51)
            };     
            % Plot Extrapolatory scores for Geometric-DeepONet (M3)
            \addplot+[ybar, fill=red] coordinates {
                (\(L1\), 15.30) (\(L2\), 60.91) (\(L3\), 60.72) (\(L4\), 57.18)
            };

            \legend{Random, Extrapolatory}
        \end{axis}
    \end{tikzpicture}
    \caption{Performance comparison of \emph{Geometric-DeepONet} under random and extrapolatory train-test splits for velocity gradients score (\(M3\)) across four loss functions (\(L1\)–\(L4\)) used in training. Higher scores indicate a more accurate predicted velocity gradient.}
    \label{fig:geometric-deeponet-m3}
\end{figure}

The loss function design plays a crucial role in model accuracy. As shown in~\figref{fig:geometric-deeponet-m3}, the performance of \emph{Geometric-DeepONet} in capturing velocity gradients (\(M3\)) across different loss functions (\(L1\)-\(L4\)). The model struggles to predict velocity gradients when trained solely on velocity values (\(L1\)), with particularly poor performance in the extrapolatory train-test split, where \(M3=15\). However, when trained on both velocity gradients and boundary velocity values (\(L3\)), \emph{Geometric-DeepONet} significantly improves its predictions, achieving \(M3=60\) in the extrapolatory case. A similar trend is observed for the random train-test split, where training on velocity gradients (\(L3\)) leads to a substantial increase in all metric (\(M1\)-\(M4\)) compared to training on velocity alone (\(L1\)) (see~\tabref{tab:metrics-comparison}). These findings emphasize the importance of incorporating gradient information in SciML model training to enhance the accuracy and generalization of flow dynamics predictions. Below, we outline how each loss function affects prediction quality and generalization:

\begin{itemize}
    \item \textbf{\(L1\) (Relative MSE Loss)}: This baseline loss focuses solely on velocity discrepancies, leading to moderate \(M1\) and \(M2\) scores. However, it does not explicitly enforce derivative consistency, resulting in poor gradients as shown by the significantly lower \(M3\) accuracy. The gap in performance between random and extrapolatory settings is largest under this loss function, highlighting the model limitation in extrapolating beyond the training distribution flow conditions.
    
    \item \textbf{\(L2\) (Relative Derivative MSE Loss)}: Incorporating velocity gradients into the loss significantly improves \(M3\), demonstrating that enforcing gradient consistency is crucial for capturing complex flow patterns. Additionally, the inclusion of derivatives enhances \(M2\), particularly in the extrapolatory case, suggesting that gradient information plays a critical role in capturing boundary-layer dynamics.

    \item \textbf{\(L3\) (Pure Derivative + Boundary Loss)}: By focusing exclusively on velocity gradients and boundary consistency, \(M3\) scores improve further, surpassing those achieved with \(L2\). Notably, this loss also enhances generalization in the extrapolatory regime, reducing the performance gap with the random splitting case when compared to \(L1\). This indicates that enforcing derivative accuracy leads to more robust generalization for out-of-sample flow conditions. 

    \item \textbf{\(L4\) (Pure Derivative + Boundary + Physics-Informed Loss)}: The addition of the mass continuity constraint improves the continuity residual error (\(M4\)), enforcing better mass conservation across the domain. However, this improvement comes at the cost of reduced \(M1\)-\(M3\) performance compared to \(L2\) and \(L3\), likely due to a trade-off between strictly satisfying physical constraints and achieving local accuracy in velocity and gradient predictions. In essence, the model must balance the ground-truth gradient data with the zero-trace velocity gradient condition (continuity). Because that constraint is not strictly satisfied by the training data, local accuracy is sometimes compromised to preserve global mass continuity.

\end{itemize}

\begin{figure} [!ht]
    \begin{minipage}[b]{0.33\linewidth}
        \centering
        \includegraphics[width=\linewidth]{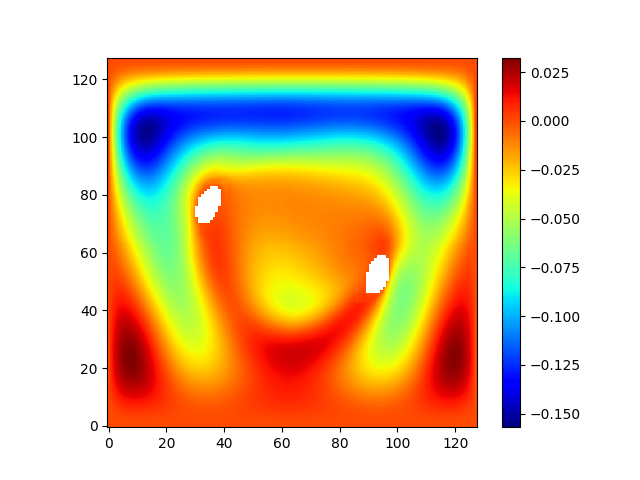}
        \textbf{Ground Truth (\(u\))}
    \end{minipage}
    \begin{minipage}[b]{0.33\linewidth}
        \centering
        \includegraphics[width=\linewidth]{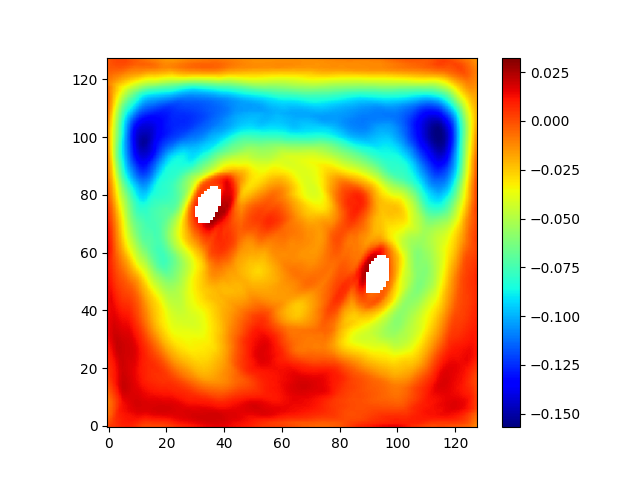}
        \textbf{\(L1\) (\(u\))}
    \end{minipage}
    \begin{minipage}[b]{0.33\linewidth}
        \centering
        \includegraphics[width=\linewidth]{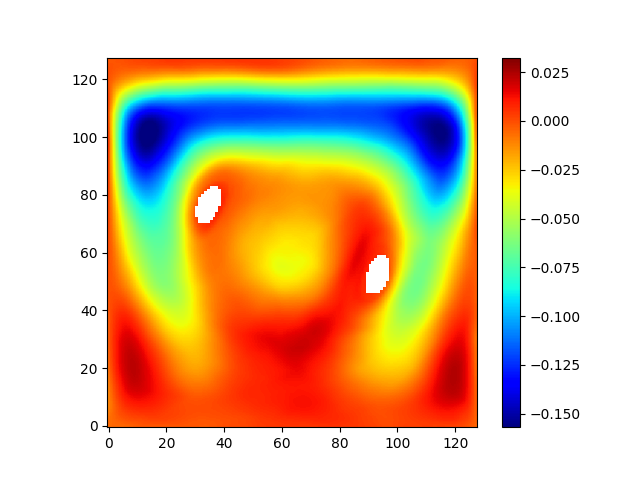}
        \textbf{\(L2\) (\(u\))}
    \end{minipage} \\
    %\vspace{0.1em} % Space between rows
    \begin{minipage}[b]{0.33\linewidth}
        \centering
        \includegraphics[width=\linewidth]{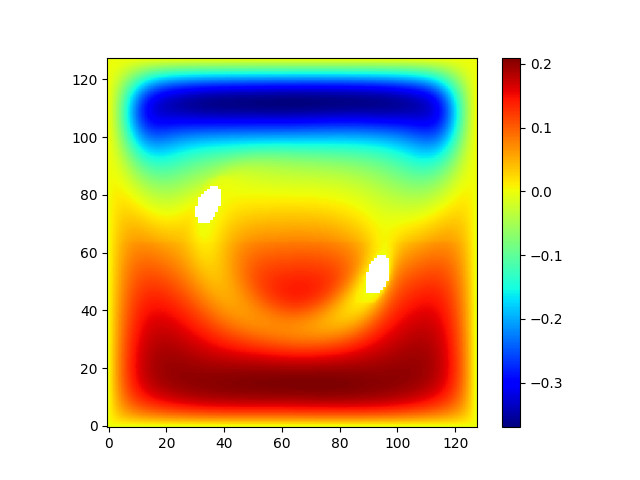}
        \textbf{Ground Truth (\(v\))}
    \end{minipage}
    \begin{minipage}[b]{0.33\linewidth}
        \centering
        \includegraphics[width=\linewidth]{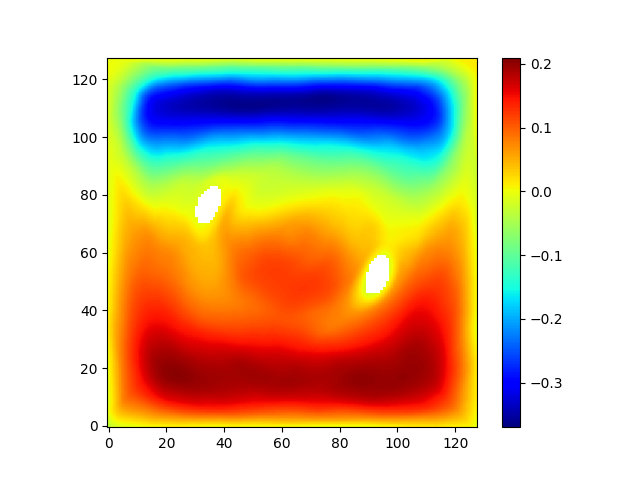}
        \textbf{\(L1\) (\(v\))}
    \end{minipage}
    \begin{minipage}[b]{0.33\linewidth}
        \centering
        \includegraphics[width=\linewidth]{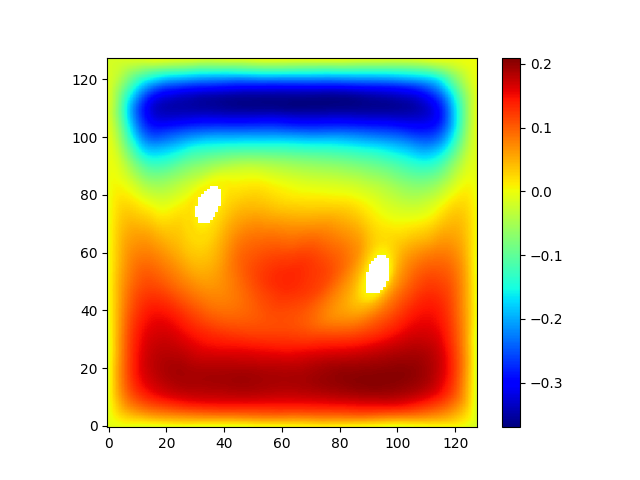}
        \textbf{\(L2\) (\(v\))}
    \end{minipage} \\
    %\vspace{0.1em} % Space between rows
    \begin{minipage}[b]{0.33\linewidth}
        \centering
        \includegraphics[width=\linewidth]{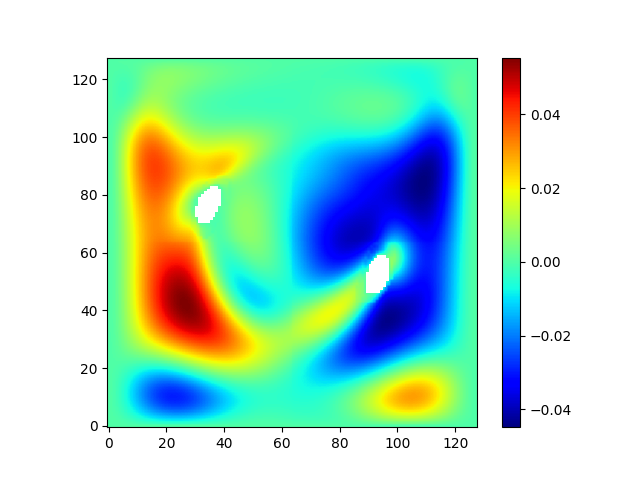}
        \textbf{Ground Truth (\(w\))}
    \end{minipage}
    \begin{minipage}[b]{0.33\linewidth}
        \centering
        \includegraphics[width=\linewidth]{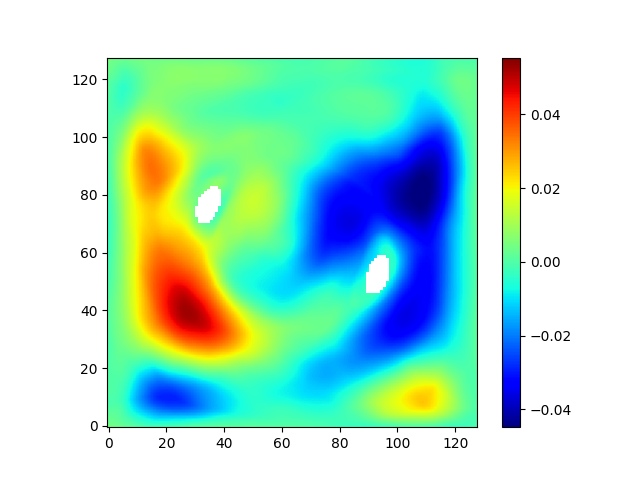}
        \textbf{\(L1\) (\(w\))}
    \end{minipage}
    \begin{minipage}[b]{0.33\linewidth}
        \centering
        \includegraphics[width=\linewidth]{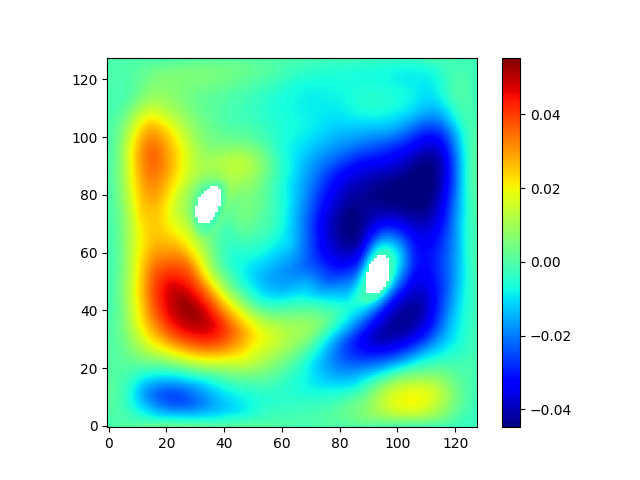}
        \textbf{\(L2\) (\(w\))}
    \end{minipage} \\
    %\vspace{0.1em} % Space between rows
    \begin{minipage}[b]{0.33\linewidth}
        \centering
        \includegraphics[width=\linewidth]{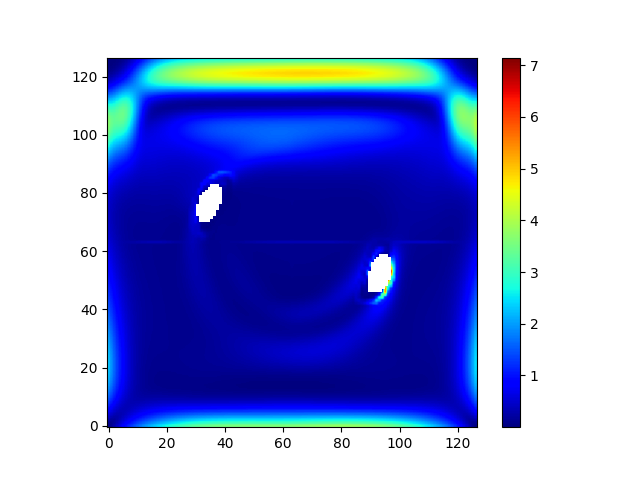}
        \textbf{Ground Truth (\(\frac{\partial u_i}{\partial x_j}\))}
    \end{minipage}
    \begin{minipage}[b]{0.33\linewidth}
        \centering
        \includegraphics[width=\linewidth]{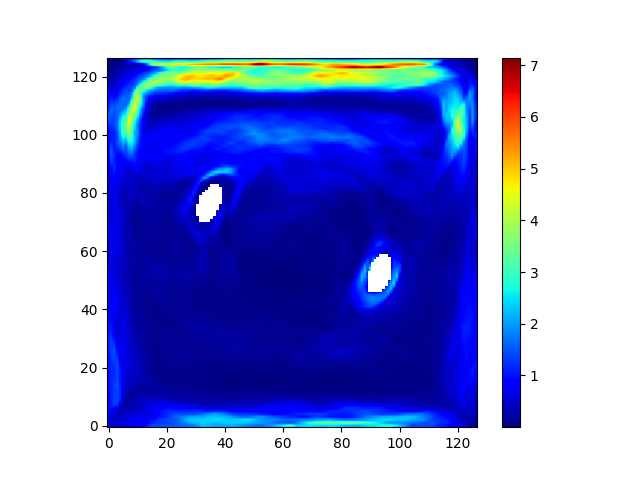}
        \textbf{\(L1\) (\(\frac{\partial u_i}{\partial x_j}\))}
    \end{minipage}
    \begin{minipage}[b]{0.33\linewidth}
        \centering
        \includegraphics[width=\linewidth]{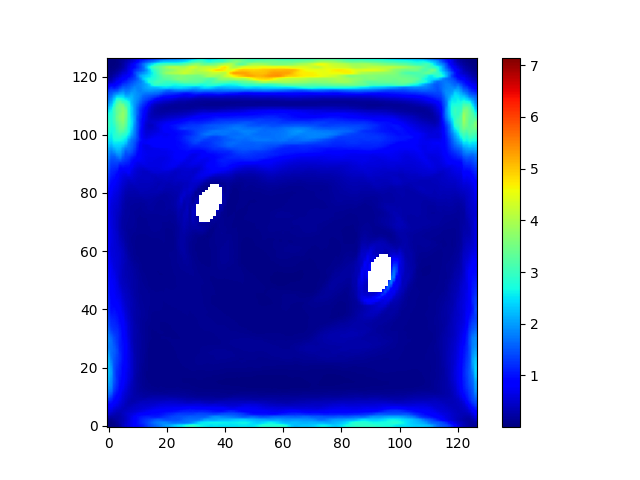}
        \textbf{\(L2\) (\(\frac{\partial u_i}{\partial x_j}\))}
    \end{minipage}
    \caption{Comparison of \emph{Geometric-DeepONet} predictions for velocity components and gradients in the \(XZ\) middle plane using the \(L1\) versus \(L2\) loss functions. Each row begins with the ground truth (first column), followed by predictions using the \(L1\) (second column) and \(L2\) (third column) loss functions. The first row displays the velocity component in the x-direction (\(u\)), the second row shows the velocity in the y-direction (\(v\)), and the third row presents the velocity in the z-direction (\(w\)). The fourth row illustrates the Frobenius norm of the velocity gradient tensor.}
    \label{fig:geo-xz_L1-L2} 
\end{figure}

\figref{fig:geo-xz_L1-L2} illustrates a sample prediction for a ring geometry, using \emph{Geometric-DeepONet}, of velocity components and gradients in the \(XZ\) middle plane, comparing the effects of the \(L1\) and \(L2\) loss functions. Introducing the velocity gradients loss terms in \(L2\) result in noticeably smoother and more accurate predictions, particularly in capturing fine-scale velocity structures and gradient variations. The third column, representing \(L2\), exhibits improved gradient prediction and fewer velocity distortions relative to the second column (\(L1\)), highlighting the importance of explicitly incorporating spatial derivatives in the loss formulation.

\subsection{Generalization in Extrapolatory Splitting} \label{subsec:random-extrapolatory}

Across all loss functions, \emph{Geometric-DeepONet} demonstrates better performance compared to \emph{DeepONet} in both the random and extrapolatory test sets. This performance gap is particularly evident for velocity gradients accuracy (\(M3\)), where \emph{Geometric-DeepONet} consistently achieves higher scores, especially when trained with gradient-aware loss functions (\(L2\) and \(L3\)). However, both models experience a noticeable performance drop in the extrapolatory setting, which is expected due to the increased complexity of predicting out-of-sample flow fields. In particular, higher Reynolds numbers often involve stronger recirculating flow structures, pushing the model beyond the parameter space it was trained on. For instance, while \emph{Geometric-DeepONet} scores \(M3 = 81\) under the random train-test split using \(L3\) loss function, its performance declines to \(M3 = 60\) in the extrapolatory split case using the same \(L3\) loss function. This reduction highlights the challenge of extrapolating flow physics beyond the training distribution.

\begin{figure} [!ht]
    \begin{minipage}[b]{0.33\linewidth}
        \centering
        \includegraphics[width=\linewidth]{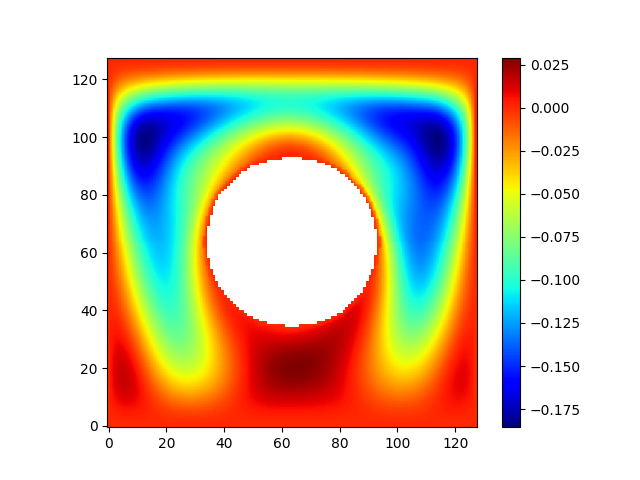}
        \textbf{Ground Truth (\(u\))}
    \end{minipage}
    \begin{minipage}[b]{0.33\linewidth}
        \centering
        \includegraphics[width=\linewidth]{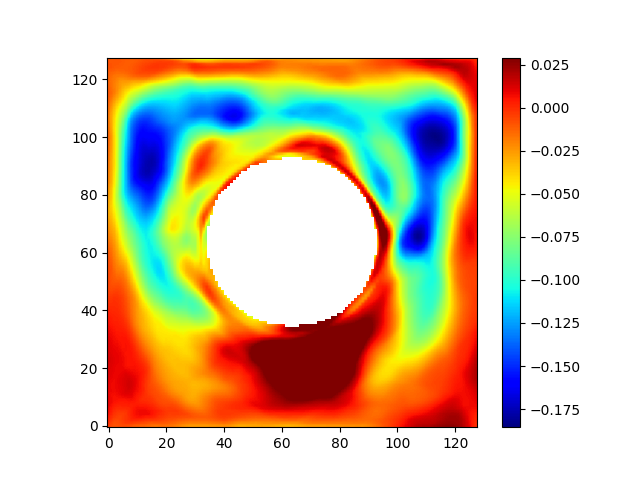}
        \textbf{\(L1\) (\(u\))}
    \end{minipage}
    \begin{minipage}[b]{0.33\linewidth}
        \centering
        \includegraphics[width=\linewidth]{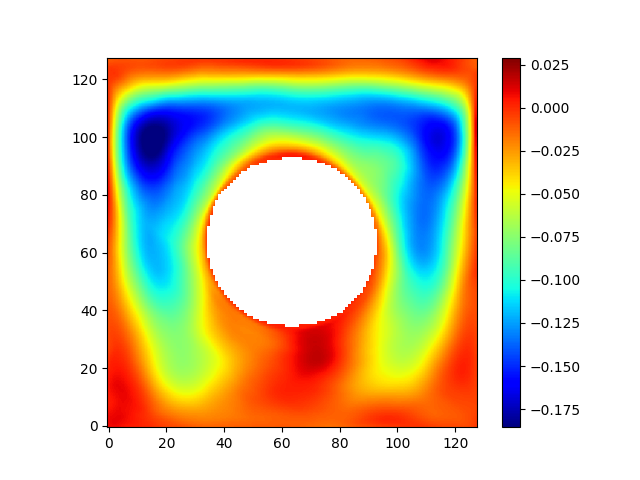}
        \textbf{\(L3\) (\(u\))}
    \end{minipage} \\
    %\vspace{0.1em} % Space between rows
    \begin{minipage}[b]{0.33\linewidth}
        \centering
        \includegraphics[width=\linewidth]{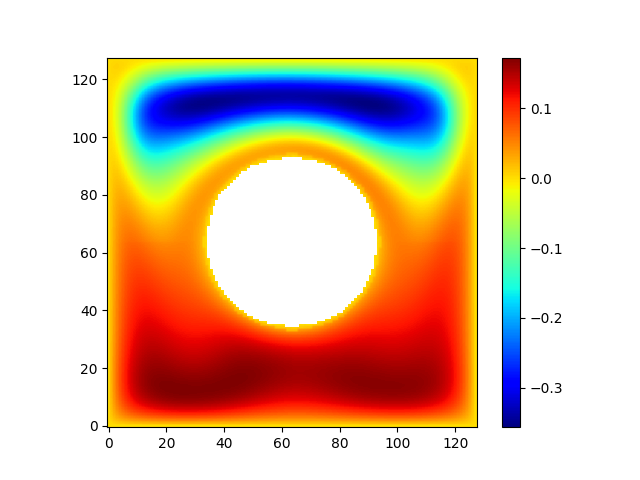}
        \textbf{Ground Truth (\(v\))}
    \end{minipage}
    \begin{minipage}[b]{0.33\linewidth}
        \centering
        \includegraphics[width=\linewidth]{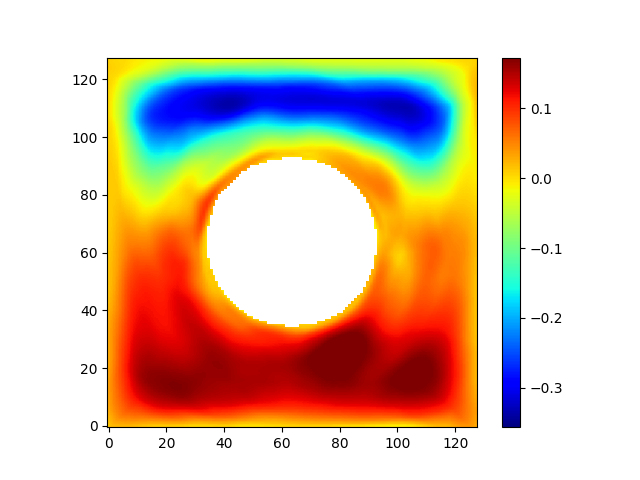}
        \textbf{\(L1\) (\(v\))}
    \end{minipage}
    \begin{minipage}[b]{0.33\linewidth}
        \centering
        \includegraphics[width=\linewidth]{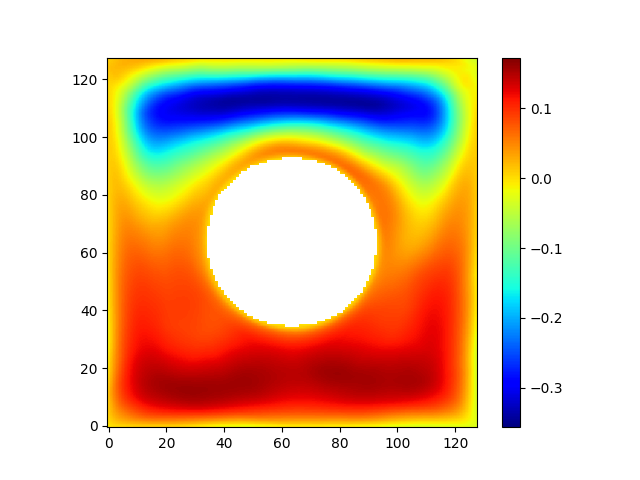}
        \textbf{\(L3\) (\(v\))}
    \end{minipage} \\
    %\vspace{0.1em} % Space between rows
    \begin{minipage}[b]{0.33\linewidth}
        \centering
        \includegraphics[width=\linewidth]{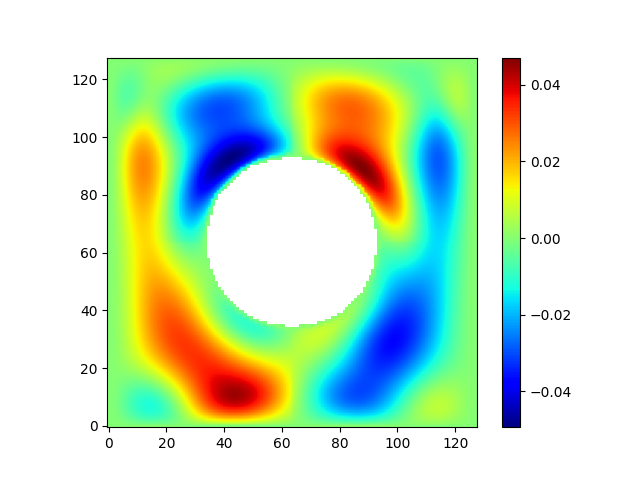}
        \textbf{Ground Truth (\(w\))}
    \end{minipage}
    \begin{minipage}[b]{0.33\linewidth}
        \centering
        \includegraphics[width=\linewidth]{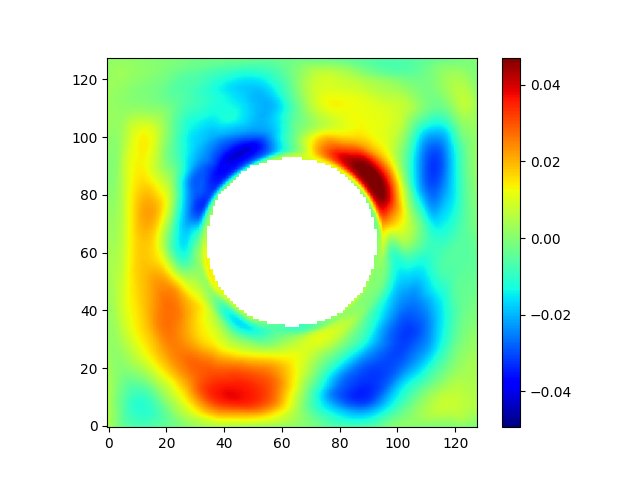}
        \textbf{\(L1\) (\(w\))}
    \end{minipage}
    \begin{minipage}[b]{0.33\linewidth}
        \centering
        \includegraphics[width=\linewidth]{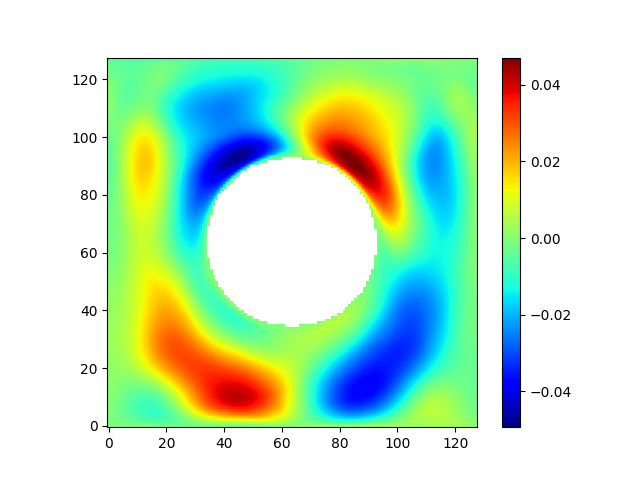}
        \textbf{\(L3\) (\(w\))}
    \end{minipage} \\
    %\vspace{0.1em} % Space between rows
    \begin{minipage}[b]{0.33\linewidth}
        \centering
        \includegraphics[width=\linewidth]{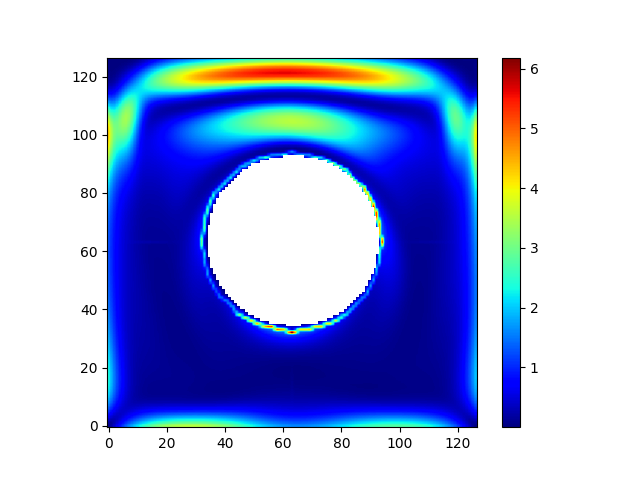}
        \textbf{Ground Truth (\(\frac{\partial u_i}{\partial x_j}\))}
    \end{minipage}
    \begin{minipage}[b]{0.33\linewidth}
        \centering
        \includegraphics[width=\linewidth]{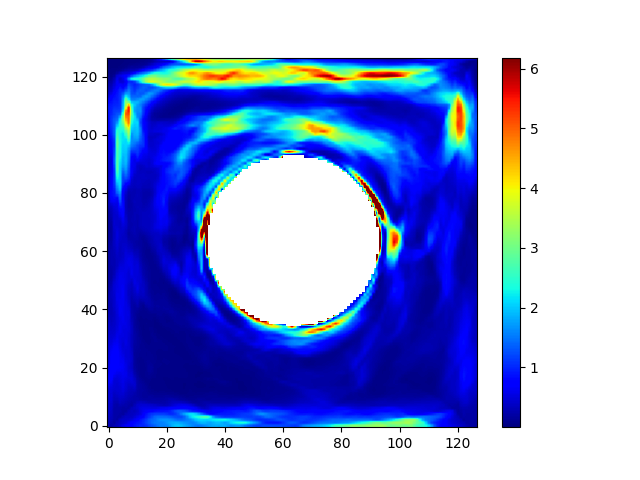}
        \textbf{\(L1\) (\(\frac{\partial u_i}{\partial x_j}\))}
    \end{minipage}
    \begin{minipage}[b]{0.33\linewidth}
        \centering
        \includegraphics[width=\linewidth]{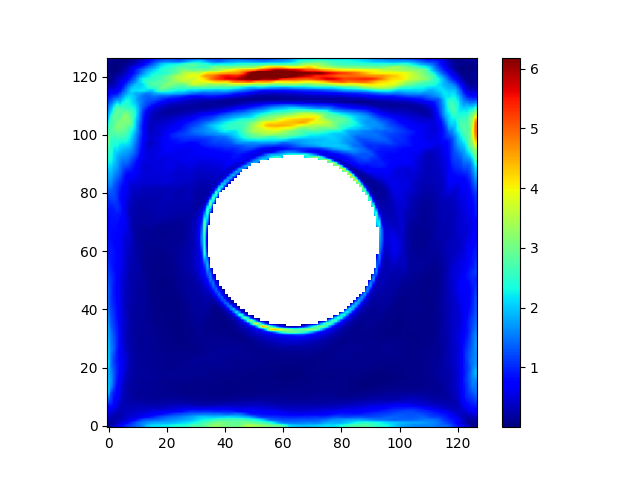}
        \textbf{\(L3\) (\(\frac{\partial u_i}{\partial x_j}\))}
    \end{minipage}
    \caption{Comparison of \emph{Geometric-DeepONet} predictions for velocity components and gradients in the \(XZ\) middle plane using the \(L1\) versus \(L3\) loss functions in the extrapolatory train-test splitting. Each row begins with the ground truth (first column), followed by predictions using the \(L1\) (second column) and \(L3\) (third column) loss functions. The first row displays the velocity component in the x-direction (\(u\)), the second row shows the velocity in the y-direction (\(v\)), and the third row presents the velocity in the z-direction (\(w\)). The fourth row illustrates the Frobenius norm of the velocity gradient tensor.}
    \label{fig:geo-xz_L1-L3-hard} 
\end{figure}

\begin{figure}[!ht]
    \centering
    \begin{minipage}[b]{0.45\linewidth}
        \centering
        \includegraphics[width=\linewidth]{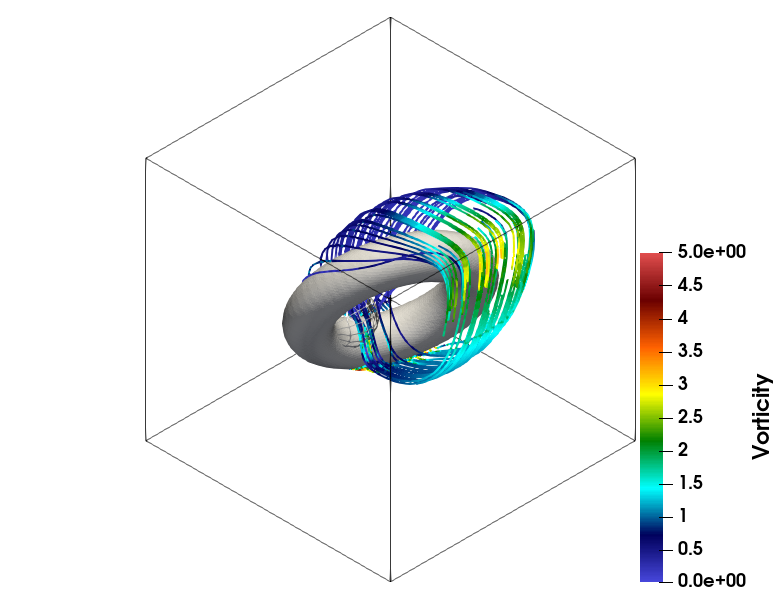}
        \textbf{Ground Truth (Ring)}
    \end{minipage}
    \begin{minipage}[b]{0.45\linewidth}
        \centering
        \includegraphics[width=\linewidth]{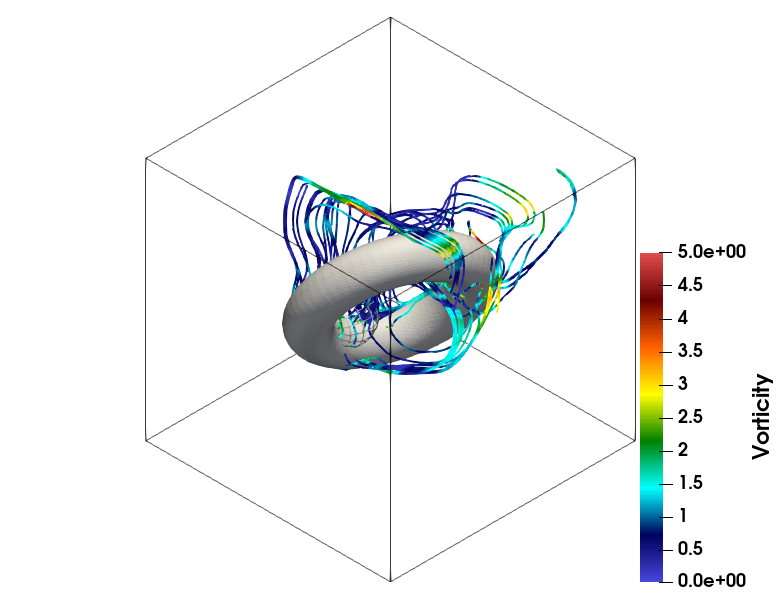}
        \textbf{\(L_1\) Prediction}
    \end{minipage} \\
    \vspace{0.5em} % Adds spacing between rows
    \begin{minipage}[b]{0.45\linewidth}
        \centering
        \includegraphics[width=\linewidth]{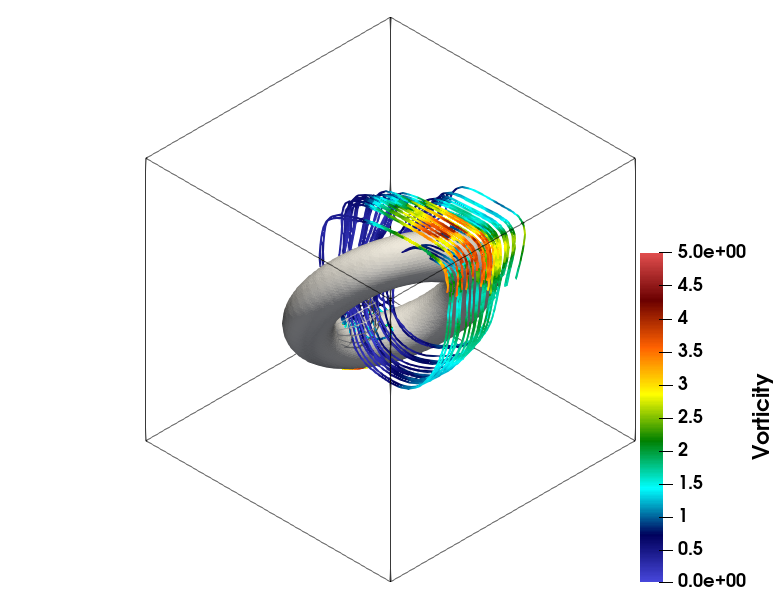}
        \textbf{\(L_2\) Prediction}
    \end{minipage} 
    \begin{minipage}[b]{0.45\linewidth}
        \centering
        \includegraphics[width=\linewidth]{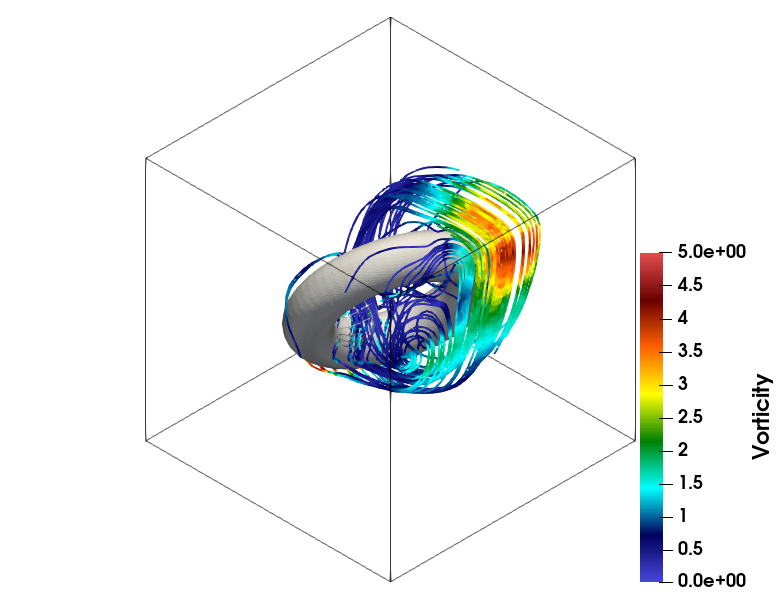}
        \textbf{\(L_3\) Prediction}
    \end{minipage} \\
    \vspace{0.5em} % Adds spacing between rows
    \begin{minipage}[b]{0.45\linewidth}
        \centering
        \includegraphics[width=\linewidth]{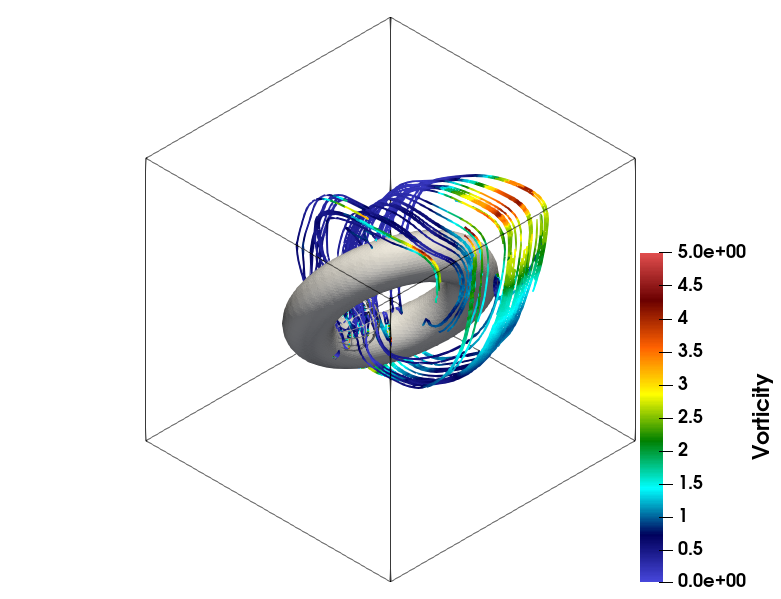}
        \textbf{\(L_4\) Prediction}
    \end{minipage}
    \caption{Comparison of streamline predictions using \emph{Geometric-DeepONet} for flow around a ring geometry using different loss functions (\(L_1\)–\(L_4\)). The first panel shows the ground truth, followed by model predictions trained with different loss functions.}
    \label{fig:geo-hard-streamlines} 
\end{figure}

~\figref{fig:geo-xz_L1-L3-hard} shows velocity components and gradients in the \(XZ\) middle plane for a cylinder geometry using \emph{Geometric-DeepONet}. It compares predictions under the \(L1\) and \(L3\) loss functions in the extrapolatory train-test split. \(L3\) improves accuracy, reducing velocity distortions in out-of-distribution samples. This suggests that training with a pure derivative and boundary loss enhances model robustness for unseen flow conditions. Future work could address extrapolation to higher Reynolds numbers by refining the loss terms with additional physics constraints such as the momentum conservation equation. Additionally,~\figref{fig:geo-hard-streamlines} illustrates the effect of different loss functions on streamline accuracy for the ring geometry. While training with \(L1\) results in visibly incorrect streamlines, the \(L3\) loss function produces the most accurate flow structures, demonstrating its effectiveness in preserving detailed flow features and variations.

\begin{figure} [!ht]
    \begin{minipage}[b]{0.33\linewidth}
        \centering
        \includegraphics[width=\linewidth]{Figures/field_predictions/depo-vs-geo/extrapolatory/u_XZ_truth.png}
        \textbf{Ground Truth (\(u\))}
    \end{minipage}
    \begin{minipage}[b]{0.33\linewidth}
        \centering
        \includegraphics[width=\linewidth]{Figures/field_predictions/depo-vs-geo/extrapolatory/L1_u_XZ.png}
        \textbf{\(L1\) (\(u\))}
    \end{minipage}
    \begin{minipage}[b]{0.33\linewidth}
        \centering
        \includegraphics[width=\linewidth]{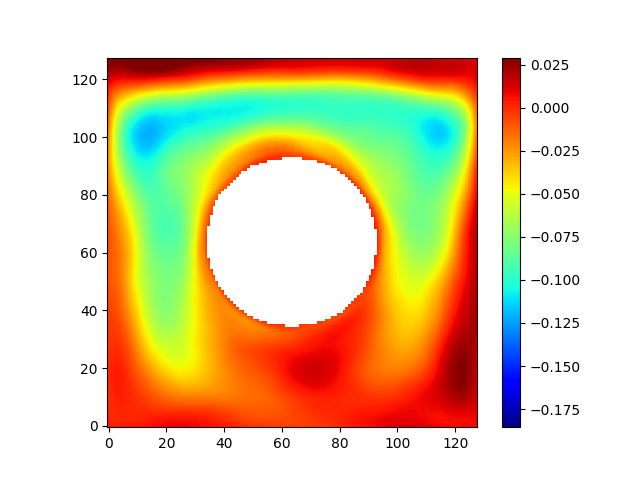}
        \textbf{\(L4\) (\(u\))}
    \end{minipage} \\
    %\vspace{0.1em} % Space between rows
    \begin{minipage}[b]{0.33\linewidth}
        \centering
        \includegraphics[width=\linewidth]{Figures/field_predictions/depo-vs-geo/extrapolatory/v_XZ_truth.png}
        \textbf{Ground Truth (\(v\))}
    \end{minipage}
    \begin{minipage}[b]{0.33\linewidth}
        \centering
        \includegraphics[width=\linewidth]{Figures/field_predictions/depo-vs-geo/extrapolatory/L1_v_XZ.png}
        \textbf{\(L1\) (\(v\))}
    \end{minipage}
    \begin{minipage}[b]{0.33\linewidth}
        \centering
        \includegraphics[width=\linewidth]{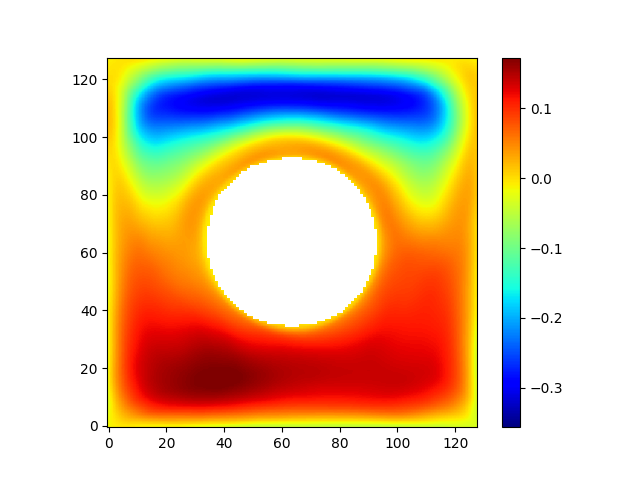}
        \textbf{\(L4\) (\(v\))}
    \end{minipage} \\
    %\vspace{0.1em} % Space between rows
    \begin{minipage}[b]{0.33\linewidth}
        \centering
        \includegraphics[width=\linewidth]{Figures/field_predictions/depo-vs-geo/extrapolatory/w_XZ_truth.png}
        \textbf{Ground Truth (\(w\))}
    \end{minipage}
    \begin{minipage}[b]{0.33\linewidth}
        \centering
        \includegraphics[width=\linewidth]{Figures/field_predictions/depo-vs-geo/extrapolatory/L1_w_XZ.png}
        \textbf{\(L1\) (\(w\))}
    \end{minipage}
    \begin{minipage}[b]{0.33\linewidth}
        \centering
        \includegraphics[width=\linewidth]{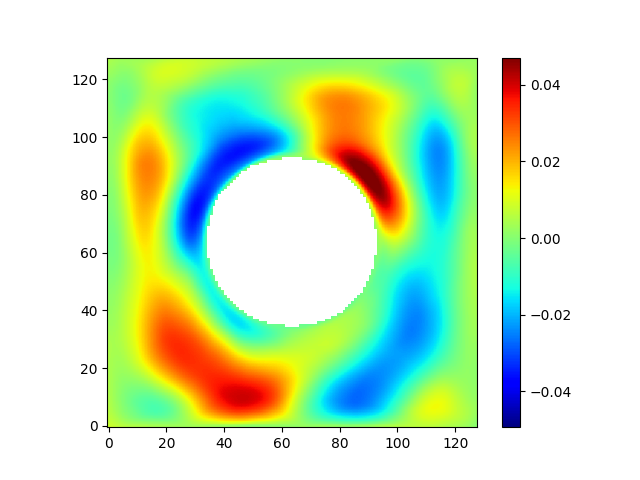}
        \textbf{\(L4\) (\(w\))}
    \end{minipage} \\
    %\vspace{0.1em} % Space between rows
    \begin{minipage}[b]{0.33\linewidth}
        \centering
        \includegraphics[width=\linewidth]{Figures/field_predictions/depo-vs-geo/extrapolatory/true_deriv_XZ.png}
        \textbf{Ground Truth (\(\frac{\partial u_i}{\partial x_j}\))}
    \end{minipage}
    \begin{minipage}[b]{0.33\linewidth}
        \centering
        \includegraphics[width=\linewidth]{Figures/field_predictions/depo-vs-geo/extrapolatory/L1_deriv_XZ.png}
        \textbf{\(L1\) (\(\frac{\partial u_i}{\partial x_j}\))}
    \end{minipage}
    \begin{minipage}[b]{0.33\linewidth}
        \centering
        \includegraphics[width=\linewidth]{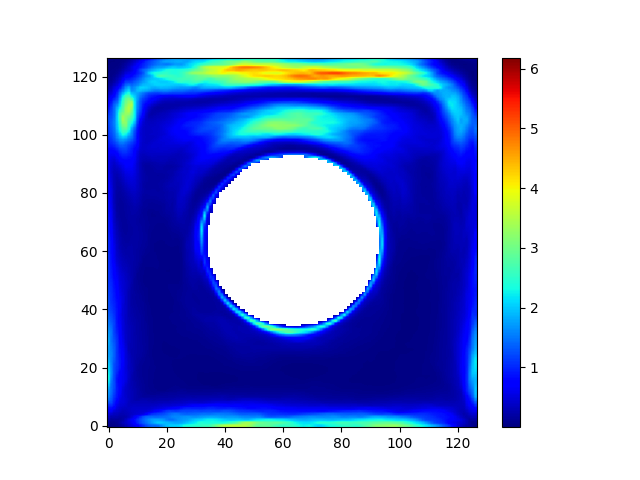}
        \textbf{\(L4\) (\(\frac{\partial u_i}{\partial x_j}\))}
    \end{minipage}
    \caption{Comparison of \emph{Geometric-DeepONet} predictions for velocity components and gradients in the \(XZ\) middle plane using the \(L1\) versus \(L4\) loss functions in the extrapolatory train-test splitting. Each row begins with the ground truth (first column), followed by predictions using the \(L1\) (second column) and \(L4\) (third column) loss functions. The first row displays the velocity component in the x-direction (\(u\)), the second row shows the velocity in the y-direction (\(v\)), and the third row presents the velocity in the z-direction (\(w\)). The fourth row illustrates the Frobenius norm of the velocity gradient tensor.}
    \label{fig:geo-xz_L1-L4-hard} 
\end{figure}

~\figref{fig:geo-xz_L1-L4-hard} presents a comparison between the \(L1\) and \(L4\) loss functions for the same geometry in the extrapolatory train-test split. The inclusion of physics constraints in \(L4\), through continuity equation residuals, leads to an improved prediction of velocity gradients. However, this comes at the cost of increased diffusion in the predicted x-direction velocity component (\(u\)). Because the continuity equation enforces the trace of the velocity gradient tensor to be zero, it can conflict with derivative-based terms when balancing local MSE against global mass conservation, sometimes leading to over-regularization. Mitigating this over-diffusion may require tuning the continuity residual’s weight so that mass conservation does not overshadow local gradient accuracy. This trade-off highlights the impact of incorporating physics-based regularization, balancing physical consistency with local accuracy in SciML predictions.

\section{Conclusion} \label{sec:conclusion}

Our approach leverages an MLP-based \emph{DeepONet} architecture and augments the trunk with SIREN layers for \emph{Geometric-DeepONet}, enabling geometry encoding through the Signed Distance Field (SDF). This study demonstrates that \emph{Geometric-DeepONet} consistently outperforms \emph{DeepONet} across all evaluation metrics, particularly in capturing velocity gradients (\(M3\)) in both random and extrapolatory test settings. A key finding is the significant influence of loss function design on predictive accuracy and generalization to out-of-sample flow conditions. Loss functions incorporating velocity gradient terms (\(L2\) and \(L3\)) significantly improve model accuracy, especially in extrapolatory settings. Since flow-field forces are directly proportional to velocity gradients, incorporating these terms into training enhances accuracy and the model’s ability to generalize to unseen conditions. Overall, this study provides an evaluation of loss function design for SciML applications in fluid dynamics and establishes a framework for optimizing neural operator training for flow modeling around complex geometries. 

We identify several avenues of further work.  Incorporating higher Reynolds flows with time-dependent behavior, potentially using a sequence-to-sequence approach for transient data, would be a natural next step. Similarly, adding the pressure field for the momentum equation as a physics constraint could enhance physical consistency and enable the direct calculation of the lift and drag from the predictions. Future work could also emphasize higher-resolution training, particularly in boundary-layer regions with steep velocity gradients, to improve predictive accuracy. Furthermore, exploring the model's ability to extrapolate to flow around two or more geometries could further validate its generalizability. Collectively, these improvements offer a pathway toward more robust and generalizable machine learning models for scientific computing.

\section*{Acknowledgements}
We gratefully acknowledge support from the NAIRR pilot program for computational access. We also acknowledge computational resources from the ISU HPC cluster Nova and TACC Frontera. This work is supported by the AI Research Institutes program supported by NSF and USDA-NIFA under AI Institute for Resilient Agriculture, Award No. 2021-67021-35329. We also acknowledge partial support through NSF awards CMMI-2053760 and DMREF-2323716.

\section*{Data Availability}
This study utilizes the FlowBench 3D Lid-Driven Cavity (LDC) dataset, which is publicly accessible on HuggingFace at \url{https://huggingface.co/datasets/BGLab/FlowBench/tree/main/LDC_NS_3D}. The dataset is licensed under a CC-BY-NC-4.0 license and serves as a benchmark for developing and evaluating scientific machine learning (SciML) models. The code for the models, along with various loss functions, training procedures, and visualization scripts, is available at \url{https://github.com/baskargroup/DI-DeepONet}.

\clearpage

\bibliography{references}

\begin{thebibliography}{43}
\providecommand{\natexlab}[1]{#1}
\providecommand{\url}[1]{\texttt{#1}}
\expandafter\ifx\csname urlstyle\endcsname\relax
  \providecommand{\doi}[1]{doi: #1}\else
  \providecommand{\doi}{doi: \begingroup \urlstyle{rm}\Url}\fi

\bibitem[Smyth and Moum(2009)]{smyth2009three}
WD~Smyth and JN~Moum.
\newblock \emph{Three-dimensional (3D) turbulence}.
\newblock Academic Press, 2009.

\bibitem[Sreenivasan(2019)]{sreenivasan2019turbulent}
Katepalli~R Sreenivasan.
\newblock Turbulent mixing: A perspective.
\newblock \emph{Proceedings of the National Academy of Sciences}, 116\penalty0 (37):\penalty0 18175--18183, 2019.

\bibitem[Fimbres-Weihs and Wiley(2010)]{Fimbres2010}
GA~Fimbres-Weihs and DE~Wiley.
\newblock Review of 3d cfd modeling of flow and mass transfer in narrow spacer-filled channels in membrane modules.
\newblock \emph{Chemical Engineering and Processing: Process Intensification}, 49\penalty0 (7):\penalty0 759--781, 2010.

\bibitem[Chawner et~al.(2016)Chawner, Dannenhoffer, and Taylor]{Chawner2016}
John~R Chawner, John Dannenhoffer, and Nigel~J Taylor.
\newblock Geometry, mesh generation, and the cfd 2030 vision.
\newblock In \emph{46th AIAA Fluid Dynamics Conference}, page 3485, 2016.

\bibitem[Ong et~al.(2005)Ong, Nair, Keane, and Wong]{Ong2005}
Yew~Soon Ong, PB~Nair, AJ~Keane, and KW~Wong.
\newblock Surrogate-assisted evolutionary optimization frameworks for high-fidelity engineering design problems.
\newblock \emph{Knowledge Incorporation in Evolutionary Computation}, pages 307--331, 2005.

\bibitem[Hou and Behdinan(2022)]{Hou2022}
Chun Kit~Jeffery Hou and Kamran Behdinan.
\newblock Dimensionality reduction in surrogate modeling: A review of combined methods.
\newblock \emph{Data science and engineering}, 7\penalty0 (4):\penalty0 402--427, 2022.

\bibitem[Zhao et~al.(2024)Zhao, Zhang, Lou, Wang, and Yang]{Zhao2024}
Chi Zhao, Feifei Zhang, Wenqiang Lou, Xi~Wang, and Jianyong Yang.
\newblock A comprehensive review of advances in physics-informed neural networks and their applications in complex fluid dynamics.
\newblock \emph{Physics of Fluids}, 36\penalty0 (10), 2024.

\bibitem[Plathottam et~al.(2023)Plathottam, Rzonca, Lakhnori, and Iloeje]{plathottam2023review}
Siby~Jose Plathottam, Arin Rzonca, Rishi Lakhnori, and Chukwunwike~O Iloeje.
\newblock A review of artificial intelligence applications in manufacturing operations.
\newblock \emph{Journal of Advanced Manufacturing and Processing}, 5\penalty0 (3):\penalty0 e10159, 2023.

\bibitem[Keith et~al.(2025)Keith, O'Leary-Roseberry, Sanderse, Scheichl, and van Bloemen~Waanders]{keith2025scientific}
Brendan Keith, Thomas O'Leary-Roseberry, Benjamin Sanderse, Robert Scheichl, and Bart van Bloemen~Waanders.
\newblock Scientific machine learning: A symbiosis.
\newblock \emph{Foundations of Data Science}, 7\penalty0 (1):\penalty0 i--x, 2025.

\bibitem[Choudhary et~al.(2022)Choudhary, DeCost, Chen, Jain, Tavazza, Cohn, Park, Choudhary, Agrawal, Billinge, et~al.]{Choudhary2022}
Kamal Choudhary, Brian DeCost, Chi Chen, Anubhav Jain, Francesca Tavazza, Ryan Cohn, Cheol~Woo Park, Alok Choudhary, Ankit Agrawal, Simon~JL Billinge, et~al.
\newblock Recent advances and applications of deep learning methods in materials science.
\newblock \emph{npj Computational Materials}, 8\penalty0 (1):\penalty0 59, 2022.

\bibitem[Li et~al.(2021)Li, Kovachki, Azizzadenesheli, Liu, Bhattacharya, Stuart, and Anandkumar]{li2021}
Zongyi Li, Nikola Kovachki, Kamyar Azizzadenesheli, Burigede Liu, Kaushik Bhattacharya, Andrew Stuart, and Anima Anandkumar.
\newblock Fourier neural operator for parametric partial differential equations, 2021.

\bibitem[Raonić et~al.(2023)Raonić, Molinaro, Ryck, Rohner, Bartolucci, Alaifari, Mishra, and de~Bézenac]{raonic2023}
Bogdan Raonić, Roberto Molinaro, Tim~De Ryck, Tobias Rohner, Francesca Bartolucci, Rima Alaifari, Siddhartha Mishra, and Emmanuel de~Bézenac.
\newblock Convolutional neural operators for robust and accurate learning of pdes, 2023.

\bibitem[Lu et~al.(2021)Lu, Jin, Pang, Zhang, and Karniadakis]{lu2020}
Lu~Lu, Pengzhan Jin, Guofei Pang, Zhongqiang Zhang, and George~Em Karniadakis.
\newblock Learning nonlinear operators via deeponet based on the universal approximation theorem of operators.
\newblock \emph{Nature Machine Intelligence}, 3:\penalty0 218--–229, 2021.

\bibitem[Herde et~al.(2024)Herde, Raonić, Rohner, Käppeli, Molinaro, de~Bézenac, and Mishra]{herde2024poseidon}
Maximilian Herde, Bogdan Raonić, Tobias Rohner, Roger Käppeli, Roberto Molinaro, Emmanuel de~Bézenac, and Siddhartha Mishra.
\newblock Poseidon: Efficient foundation models for pdes, 2024.

\bibitem[Liu et~al.(2022)Liu, Mao, Wu, Feichtenhofer, Darrell, and Xie]{liu2022convnet2020s}
Zhuang Liu, Hanzi Mao, Chao-Yuan Wu, Christoph Feichtenhofer, Trevor Darrell, and Saining Xie.
\newblock A convnet for the 2020s, 2022.
\newblock URL \url{https://arxiv.org/abs/2201.03545}.

\bibitem[Rabeh et~al.(2024)Rabeh, Herron, Balu, Sarkar, Hegde, Krishnamurthy, and Ganapathysubramanian]{rabeh2024geometry}
Ali Rabeh, Ethan Herron, Aditya Balu, Soumik Sarkar, Chinmay Hegde, Adarsh Krishnamurthy, and Baskar Ganapathysubramanian.
\newblock Geometry matters: Benchmarking scientific ml approaches for flow prediction around complex geometries.
\newblock \emph{arXiv preprint arXiv:2501.01453}, 2024.

\bibitem[Xiao et~al.(2020)Xiao, Lai, Zhang, Li, and Gao]{Xiao2020survey}
Yun-Peng Xiao, Yu-Kun Lai, Fang-Lue Zhang, Chunpeng Li, and Lin Gao.
\newblock A survey on deep geometry learning: From a representation perspective.
\newblock \emph{Computational Visual Media}, 6\penalty0 (2):\penalty0 113--133, 2020.

\bibitem[Khara et~al.(2024{\natexlab{a}})Khara, Balu, Joshi, Sarkar, Hegde, Krishnamurthy, and Ganapathysubramanian]{NeuFENetKhara2024}
Biswajit Khara, Aditya Balu, Ameya Joshi, Soumik Sarkar, Chinmay Hegde, Adarsh Krishnamurthy, and Baskar Ganapathysubramanian.
\newblock Neufenet: Neural finite element solutions with theoretical bounds for parametric pdes.
\newblock \emph{Engineering with Computers}, 40\penalty0 (5):\penalty0 2761--2783, October 2024{\natexlab{a}}.
\newblock \doi{10.1007/s00366-024-01955-7}.
\newblock URL \url{https://doi.org/10.1007/s00366-024-01955-7}.

\bibitem[Khara et~al.(2024{\natexlab{b}})Khara, Herron, Balu, Gamdha, Yang, Saurabh, Jignasu, Jiang, Sarkar, Hegde, et~al.]{Khara2024}
Biswajit Khara, Ethan Herron, Aditya Balu, Dhruv Gamdha, Chih-Hsuan Yang, Kumar Saurabh, Anushrut Jignasu, Zhanhong Jiang, Soumik Sarkar, Chinmay Hegde, et~al.
\newblock Neural pde solvers for irregular domains.
\newblock \emph{Computer-Aided Design}, 172:\penalty0 103709, 2024{\natexlab{b}}.

\bibitem[Perera et~al.(2023)Perera, Premachandra, and Kawanaka]{perera2023enhancing}
Chamika~Janith Perera, Chinthaka Premachandra, and Hiroharu Kawanaka.
\newblock Enhancing feature detection and matching in low-pixel-resolution hyperspectral images using 3d convolution-based siamese networks.
\newblock \emph{Sensors}, 23\penalty0 (18):\penalty0 8004, 2023.

\bibitem[Ullah et~al.(2024)Ullah, Ullah, Khan, Khan, Khan, and Pau]{ullah2024conventional}
Farhan Ullah, Irfan Ullah, Rehan~Ullah Khan, Salabat Khan, Khalil Khan, and Giovanni Pau.
\newblock Conventional to deep ensemble methods for hyperspectral image classification: A comprehensive survey.
\newblock \emph{IEEE Journal of Selected Topics in Applied Earth Observations and Remote Sensing}, 17:\penalty0 3878--3916, 2024.

\bibitem[Choy et~al.(2025)Choy, Kamenev, Kossaifi, Rietmann, Kautz, and Azizzadenesheli]{choy2025factorized}
Chris Choy, Alexey Kamenev, Jean Kossaifi, Max Rietmann, Jan Kautz, and Kamyar Azizzadenesheli.
\newblock Factorized implicit global convolution for automotive computational fluid dynamics prediction.
\newblock \emph{arXiv preprint arXiv:2502.04317}, 2025.

\bibitem[Feng et~al.(2019)Feng, Feng, You, Zhao, and Gao]{feng2019meshnet}
Yutong Feng, Yifan Feng, Haoxuan You, Xibin Zhao, and Yue Gao.
\newblock Meshnet: Mesh neural network for 3d shape representation.
\newblock In \emph{Proceedings of the AAAI conference on artificial intelligence}, volume~33, pages 8279--8286, 2019.

\bibitem[He et~al.(2024)He, Koric, Abueidda, Najafi, and Jasiuk]{he2024}
Junyan He, Seid Koric, Diab Abueidda, Ali Najafi, and Iwona Jasiuk.
\newblock Geom-deeponet: A point-cloud-based deep operator network for field predictions on 3d parameterized geometries.
\newblock \emph{Computer Methods in Applied Mechanics and Engineering}, 429:\penalty0 117130, September 2024.
\newblock ISSN 0045-7825.
\newblock \doi{10.1016/j.cma.2024.117130}.
\newblock URL \url{http://dx.doi.org/10.1016/j.cma.2024.117130}.

\bibitem[Qiu et~al.(2024)Qiu, Bridges, and Chen]{qiu2024derivative}
Yuan Qiu, Nolan Bridges, and Peng Chen.
\newblock Derivative-enhanced deep operator network.
\newblock \emph{Advances in Neural Information Processing Systems}, 37:\penalty0 20945--20981, 2024.

\bibitem[Willard et~al.(2020)Willard, Jia, Xu, Steinbach, and Kumar]{willard2020integrating}
Jared Willard, Xiaowei Jia, Shaoming Xu, Michael Steinbach, and Vipin Kumar.
\newblock Integrating physics-based modeling with machine learning: A survey.
\newblock \emph{arXiv preprint arXiv:2003.04919}, 1\penalty0 (1):\penalty0 1--34, 2020.

\bibitem[Nguyen et~al.(2023)Nguyen, Nguyen, Choi, Seshadri, Udaykumar, and Baek]{nguyen2023parc}
Phong~CH Nguyen, Yen-Thi Nguyen, Joseph~B Choi, Pradeep~K Seshadri, HS~Udaykumar, and Stephen~S Baek.
\newblock Parc: Physics-aware recurrent convolutional neural networks to assimilate meso scale reactive mechanics of energetic materials.
\newblock \emph{Science advances}, 9\penalty0 (17):\penalty0 eadd6868, 2023.

\bibitem[Weikun et~al.(2023)Weikun, Nguyen, Medjaher, Christian, and Morio]{weikun2023physics}
DENG Weikun, Khanh~TP Nguyen, Kamal Medjaher, GOGU Christian, and J{\'e}r{\^o}me Morio.
\newblock Physics-informed machine learning in prognostics and health management: State of the art and challenges.
\newblock \emph{Applied Mathematical Modelling}, 124:\penalty0 325--352, 2023.

\bibitem[Liu et~al.(2024)Liu, Xu, Soroco, Wei, and Chen]{Liu2024}
Yuqiu Liu, Jingxuan Xu, Mauricio Soroco, Yunchao Wei, and Wuyang Chen.
\newblock Data-efficient inference of neural fluid fields via sciml foundation model.
\newblock \emph{arXiv preprint arXiv:2412.13897}, 2024.

\bibitem[Wang et~al.(2020)Wang, Ma, Zhao, and Tian]{Wang2020}
Qi~Wang, Yue Ma, Kun Zhao, and Yingjie Tian.
\newblock A comprehensive survey of loss functions in machine learning.
\newblock \emph{Annals of Data Science}, pages 1--26, 2020.

\bibitem[Ciampiconi et~al.(2023)Ciampiconi, Elwood, Leonardi, Mohamed, and Rozza]{Ciampiconi2023}
Lorenzo Ciampiconi, Adam Elwood, Marco Leonardi, Ashraf Mohamed, and Alessandro Rozza.
\newblock A survey and taxonomy of loss functions in machine learning.
\newblock \emph{arXiv preprint arXiv:2301.05579}, 2023.

\bibitem[Nie et~al.(2018)Nie, Hu, and Li]{Nie2018}
Feiping Nie, Zhanxuan Hu, and Xuelong Li.
\newblock An investigation for loss functions widely used in machine learning.
\newblock \emph{Communications in Information and Systems}, 18\penalty0 (1):\penalty0 37--52, 2018.

\bibitem[Tali et~al.(2024)Tali, Rabeh, Yang, Shadkhah, Karki, Upadhyaya, Dhakshinamoorthy, Saadati, Sarkar, Krishnamurthy, et~al.]{Tali2024}
Ronak Tali, Ali Rabeh, Cheng-Hau Yang, Mehdi Shadkhah, Samundra Karki, Abhisek Upadhyaya, Suriya Dhakshinamoorthy, Marjan Saadati, Soumik Sarkar, Adarsh Krishnamurthy, et~al.
\newblock Flowbench: A large scale benchmark for flow simulation over complex geometries.
\newblock \emph{arXiv preprint arXiv:2409.18032}, 2024.

\bibitem[Elrefaie et~al.(2024)Elrefaie, Morar, Dai, and Ahmed]{elrefaie2024drivaernet++}
Mohamed Elrefaie, Florin Morar, Angela Dai, and Faez Ahmed.
\newblock Drivaernet++: A large-scale multimodal car dataset with computational fluid dynamics simulations and deep learning benchmarks.
\newblock \emph{Advances in Neural Information Processing Systems}, 37:\penalty0 499--536, 2024.

\bibitem[Ashton et~al.(2024)Ashton, Angel, Ghate, Kenway, Wong, Kiris, Walle, Maddix, and Page]{ashton2024windsorml}
Neil Ashton, Jordan Angel, Aditya Ghate, Gaetan Kenway, Man~Long Wong, Cetin Kiris, Astrid Walle, Danielle Maddix, and Gary Page.
\newblock Windsorml: High-fidelity computational fluid dynamics dataset for automotive aerodynamics.
\newblock \emph{Advances in Neural Information Processing Systems}, 37:\penalty0 37823--37835, 2024.

\bibitem[Gao(2023)]{gao2023scientific}
Han Gao.
\newblock \emph{Scientific Deep Learning for Forward and Inverse Modeling of Spatiotemporal Physics}.
\newblock University of Notre Dame, 2023.

\bibitem[Trask et~al.(2019)Trask, Patel, Gross, and Atzberger]{trask2019gmls}
Nathaniel Trask, Ravi~G Patel, Ben~J Gross, and Paul~J Atzberger.
\newblock Gmls-nets: A framework for learning from unstructured data.
\newblock \emph{arXiv preprint arXiv:1909.05371}, 2019.

\bibitem[Main and Scovazzi(2018)]{main2018shifted}
Alex Main and Guglielmo Scovazzi.
\newblock The shifted boundary method for embedded domain computations. part {I}: Poisson and stokes problems.
\newblock \emph{Journal of Computational Physics}, 372:\penalty0 972--995, 2018.

\bibitem[Yang et~al.(2024{\natexlab{a}})Yang, Saurabh, Scovazzi, Canuto, Krishnamurthy, and Ganapathysubramanian]{yang2024optimal}
Cheng-Hau Yang, Kumar Saurabh, Guglielmo Scovazzi, Claudio Canuto, Adarsh Krishnamurthy, and Baskar Ganapathysubramanian.
\newblock Optimal surrogate boundary selection and scalability studies for the shifted boundary method on octree meshes.
\newblock \emph{Computer Methods in Applied Mechanics and Engineering}, 419:\penalty0 116686, 2024{\natexlab{a}}.

\bibitem[Yang et~al.(2024{\natexlab{b}})Yang, Scovazzi, Krishnamurthy, and Ganapathysubramanian]{yang2024simulating}
Cheng-Hau Yang, Guglielmo Scovazzi, Adarsh Krishnamurthy, and Baskar Ganapathysubramanian.
\newblock Simulating incompressible flows over complex geometries using the shifted boundary method with incomplete adaptive octree meshes.
\newblock \emph{arXiv preprint arXiv:2411.00272}, 2024{\natexlab{b}}.

\bibitem[Johnston(2005)]{Johnston2005}
JP~Johnston.
\newblock Internal flows.
\newblock \emph{Turbulence}, pages 109--169, 2005.

\bibitem[Launder et~al.(2010)Launder, Poncet, and Serre]{Launder2010}
Brian Launder, S{\'e}bastien Poncet, and Eric Serre.
\newblock Laminar, transitional, and turbulent flows in rotor-stator cavities.
\newblock \emph{Annual review of fluid mechanics}, 42\penalty0 (1):\penalty0 229--248, 2010.

\bibitem[Chen and Chen(1995)]{Chen1995}
Tianping Chen and Hong Chen.
\newblock Universal approximation to nonlinear operators by neural networks with arbitrary activation functions and its application to dynamical systems.
\newblock \emph{IEEE transactions on neural networks}, 6\penalty0 (4):\penalty0 911--917, 1995.

\end{thebibliography}

\appendix
\newpage

\section{Geometries and Computational Domain} \label{sec:geom-domain}

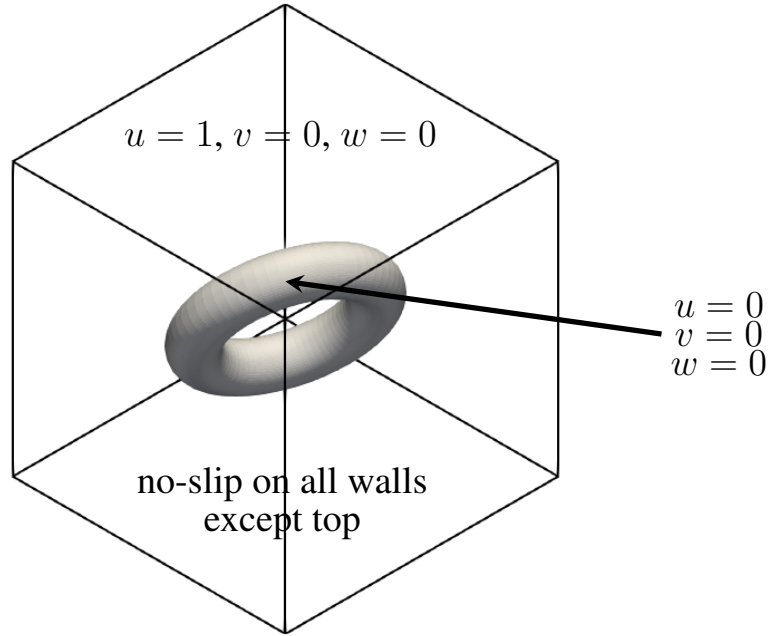
\begin{figure}[!ht]
    \centering
    \begin{tikzpicture}[
        % Adjust as needed for spacing and appearance
        node distance=0cm, 
        >=stealth,
        every node/.style={font=\small}
    ]
        % Place the image in a node
        \node (img) {\includegraphics[width=0.7\textwidth]{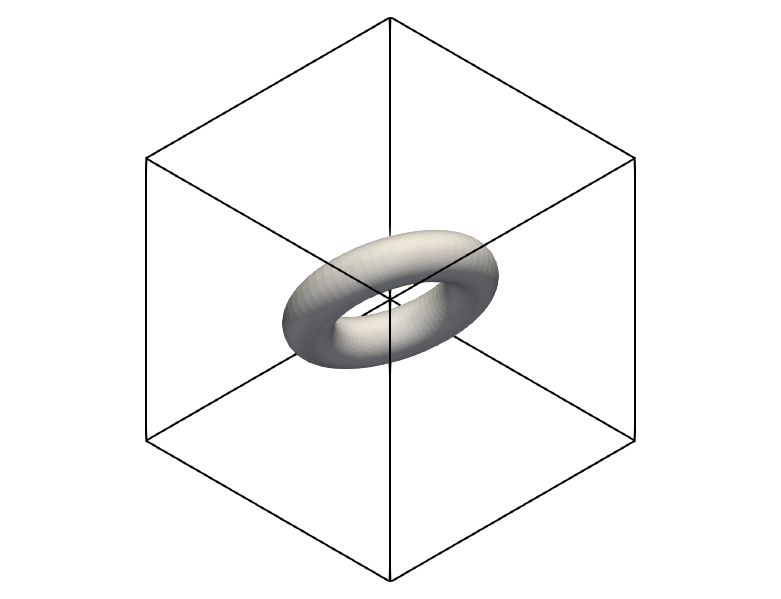}};
        
        % Label at the top boundary
        \node[align=center, below=18mm of img.north] (topBC) { \Large
            $u=1$, $v=0$, $w=0$
        };
        \node[midway, right, xshift=50mm, yshift=-2mm, align=center] (ringBC){\Large $u=0$\\\Large $v=0$\\\Large $w=0$ };
        \node[above=15mm of img.south, align=center] (noslip) {
        \shortstack{\Large no-slip on all walls \\ \Large except top}
        };
        % Draw an arrow from topBC to the center of the image,
        % label the arrow with (u=0, v=0, w=0).
        \draw[->, thick, line width=2pt] (ringBC.west) -- ($(img.center)+(0,5mm)$);

    \end{tikzpicture}
    \caption{Boundary conditions for the 3D lid-driven cavity flow. 
    The top boundary (moving lid) enforces $u = 1,\, v = 0,\, w = 0$; all other walls 
    (including the ring geometry) enforces no-slip condition $u = 0,\, v = 0,\, w = 0$.}
    \label{fig:3D-LDC-BC}
\end{figure}

The 3D LDC cases are simulated in a cubic domain of size \([0,2]\times[0,2]\times[0,2]\) with five stationary walls and one moving lid. An immersed object is positioned at the center of the cavity. In the simulations, a uniform velocity \(u=1\) is applied at the top boundary (with \(v=0\) and \(w=0\)), while a no-slip condition (\(u=v=w=0\)) constrains the other five walls. \figref{fig:3D-LDC-BC} illustrates these boundary conditions.

\begin{figure}[ht]
    \centering   
    % Row 1
    \begin{subfigure}[b]{0.2\linewidth}
        \centering
        \includegraphics[width=\linewidth]{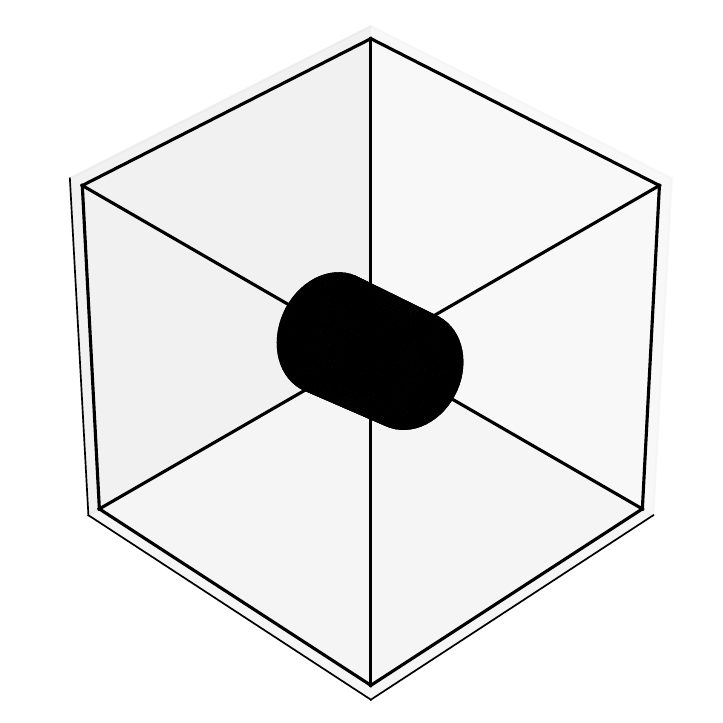}
        \caption{Geometry 1}
    \end{subfigure}
    \hspace{0.02\linewidth}
    \begin{subfigure}[b]{0.2\linewidth}
        \centering
        \includegraphics[width=\linewidth]{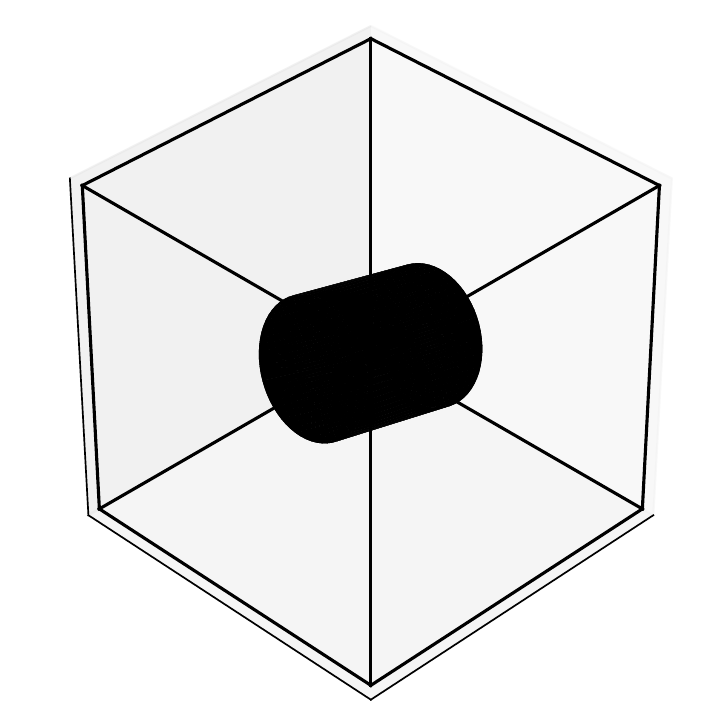}
        \caption{Geometry 2}
    \end{subfigure}
    \hspace{0.02\linewidth}
    \begin{subfigure}[b]{0.2\linewidth}
        \centering
        \includegraphics[width=\linewidth]{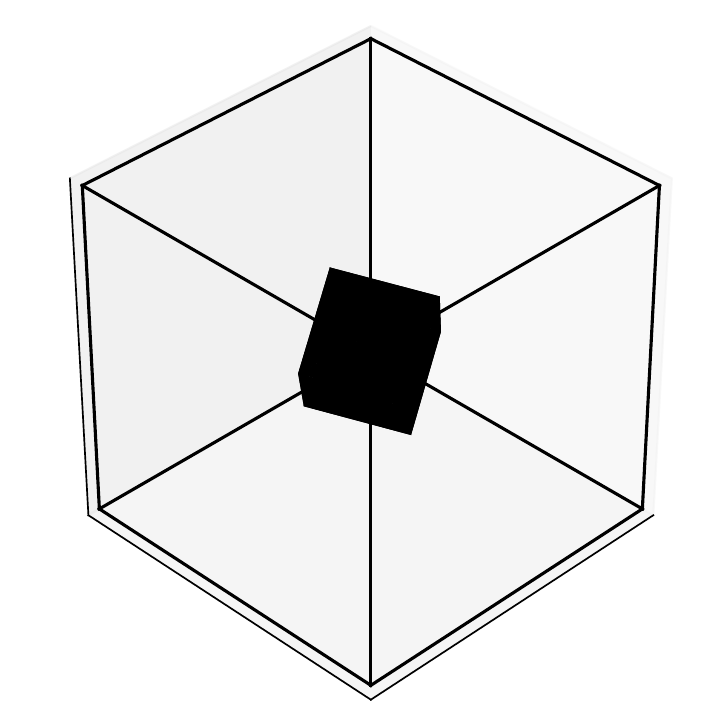}
        \caption{Geometry 3}
    \end{subfigure}
    \hspace{0.02\linewidth}
    % Row 2
    \begin{subfigure}[b]{0.2\linewidth}
        \centering
        \includegraphics[width=\linewidth]{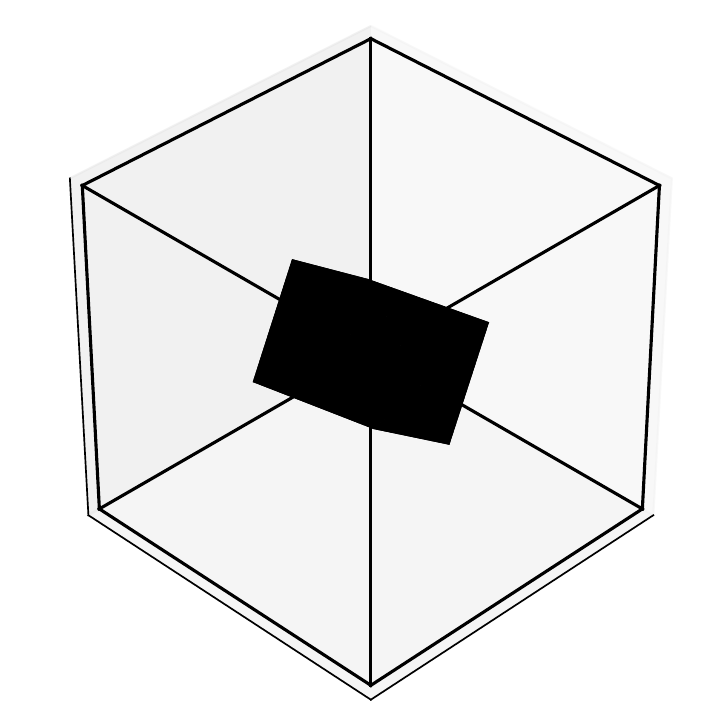}
        \caption{Geometry 4}
    \end{subfigure}
    \hspace{0.02\linewidth}
    \begin{subfigure}[b]{0.2\linewidth}
        \centering
        \includegraphics[width=\linewidth]{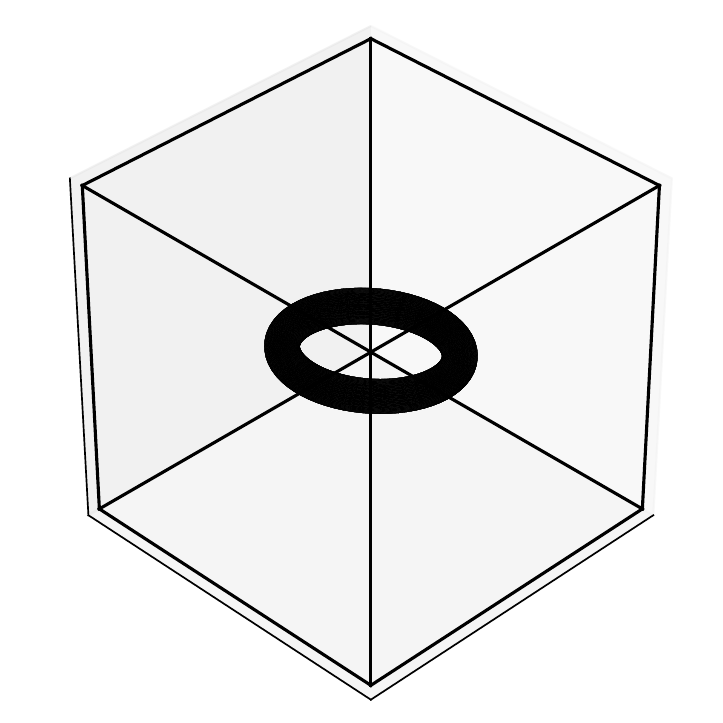}
        \caption{Geometry 5}
    \end{subfigure}
    \hspace{0.02\linewidth}
    \begin{subfigure}[b]{0.2\linewidth}
        \centering
        \includegraphics[width=\linewidth]{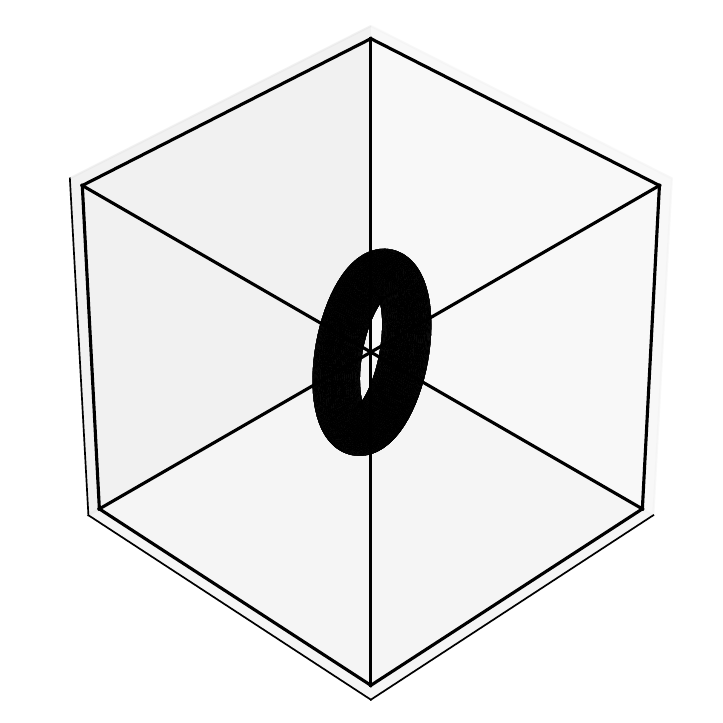}
        \caption{Geometry 6}
    \end{subfigure}
    \hspace{0.02\linewidth}
    % Row 3
    \begin{subfigure}[b]{0.2\linewidth}
        \centering
        \includegraphics[width=\linewidth]{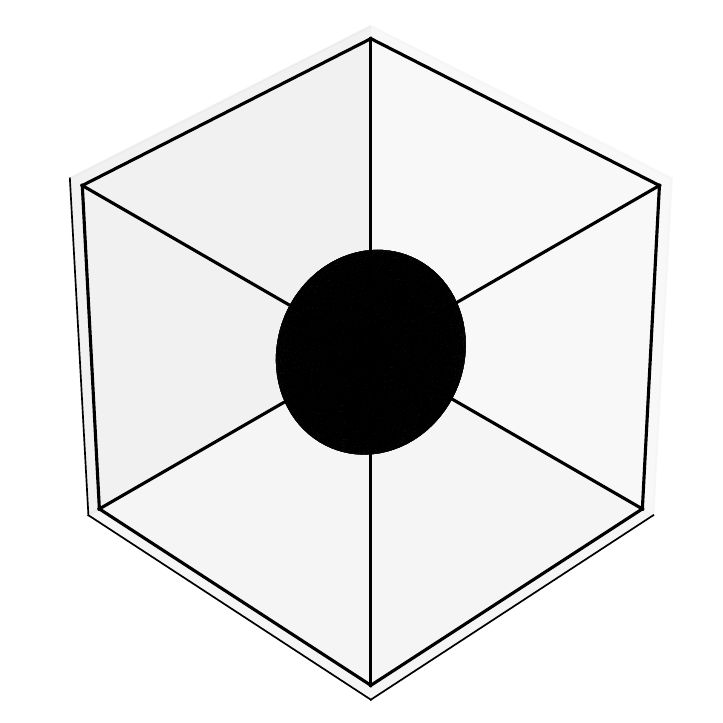}
        \caption{Geometry 7}
    \end{subfigure}
    \hspace{0.02\linewidth}
    \begin{subfigure}[b]{0.2\linewidth}
        \centering
        \includegraphics[width=\linewidth]{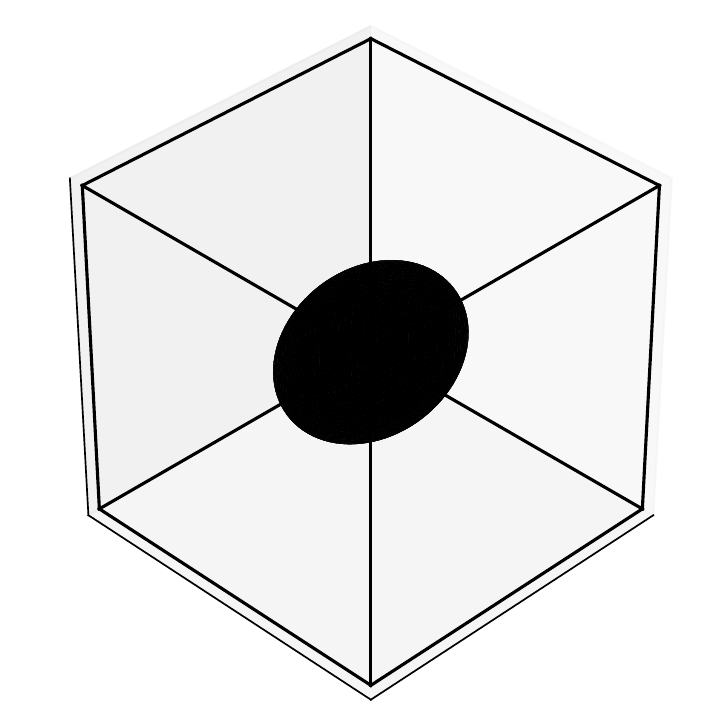}
        \caption{Geometry 8}
    \end{subfigure}
    \hspace{0.02\linewidth}
    \caption{Representative geometries from our dataset, arranged in a 2$\times$4 layout. Geometries 1 and 2 are cylinders with varied orientations and diameters. Geometries 3 and 4 are cubes of different sizes and orientations. Geometries 5 and 6 are rings with different inner and outer radii. Geometries 7 and 8 are ellipsoids that vary in their minor diameters and orientations.}
    \label{fig:geoms}
\end{figure}

\figref{fig:geoms} presents representative geometries from our dataset, which consists of 100 analytical geometries equally distributed among four classes: cylinders, boxes, rings, and ellipsoids. Each geometry is generated so that its major dimension is fixed at 1 (i.e., the cylinder’s height, or the major diameter for rings and ellipsoids), while the other dimensions vary to introduce shape diversity. In detail:

\begin{itemize}
    \item \textbf{Cylinders:} 25 cylinders are produced with the first cylinder with a height of 1 and a base radius of 0.5. To create additional geometries, three other radii (0.35, 0.4, and 0.45) are employed, each subjected to 8 random rotations.
    
    \item \textbf{Boxes:} 25 boxes are produced with the first box with dimensions \((1,1,1)\). Subsequent boxes vary the y and z dimensions (0.7, 0.8, and 0.9) while maintaining the x dimension at 1, with each variant rotated 8 times randomly.
    
    \item \textbf{Rings:} 25 rings are produced, each with a major diameter of 1 but with different inner diameters to modify the thickness (0.65, 0.75, and 0.85). Every ring is randomly rotated 8 times.
    
    \item \textbf{Ellipsoids:} 25 ellipsoids are produced with a major diameter of 1 but with different minor diameters (0.65, 0.75, and 0.85). Similarly, each ellipsoid is randomly rotated 8 times.
\end{itemize}

The generation process is implemented in Python using the \texttt{trimesh} library, and all geometries are exported as STL files. For each geometry, random rotations are applied to introduce variability in orientation. Eight representative examples selected from the dataset are shown in \figref{fig:geoms}.

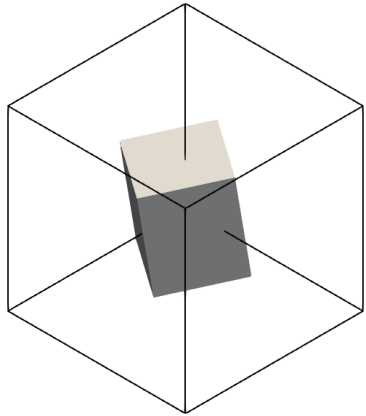
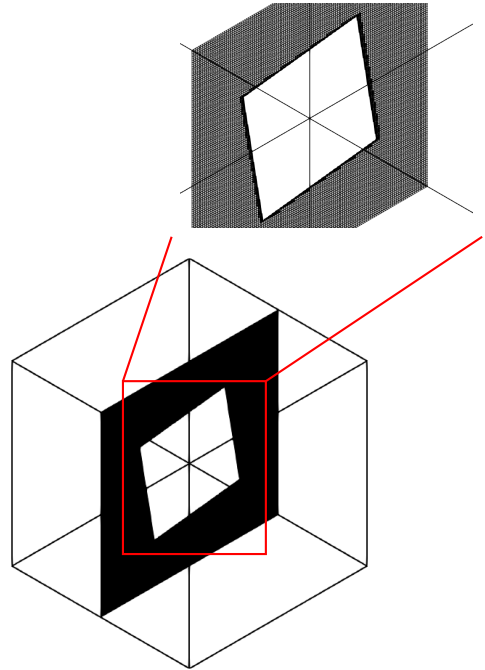
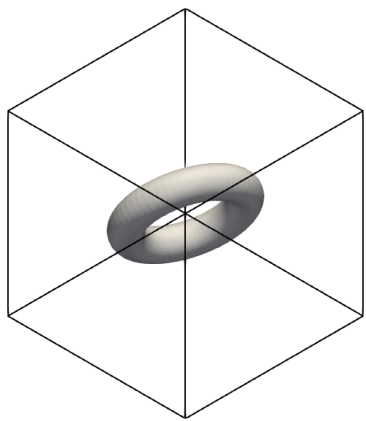
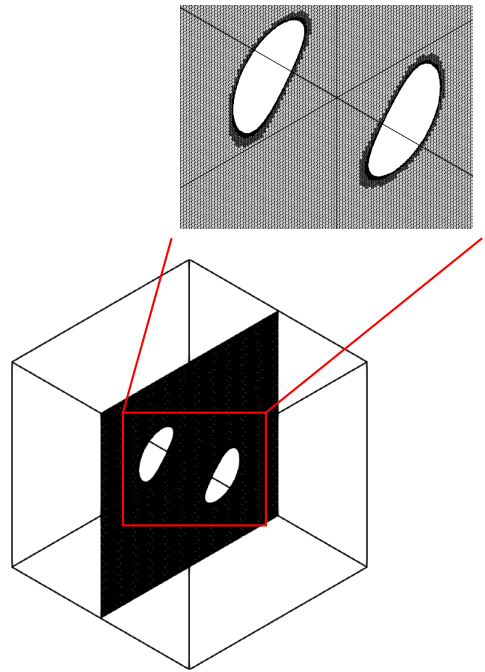
\begin{figure}
    \centering
    \begin{subfigure}[b]{0.47\linewidth}
        \includegraphics[width=0.97\linewidth]{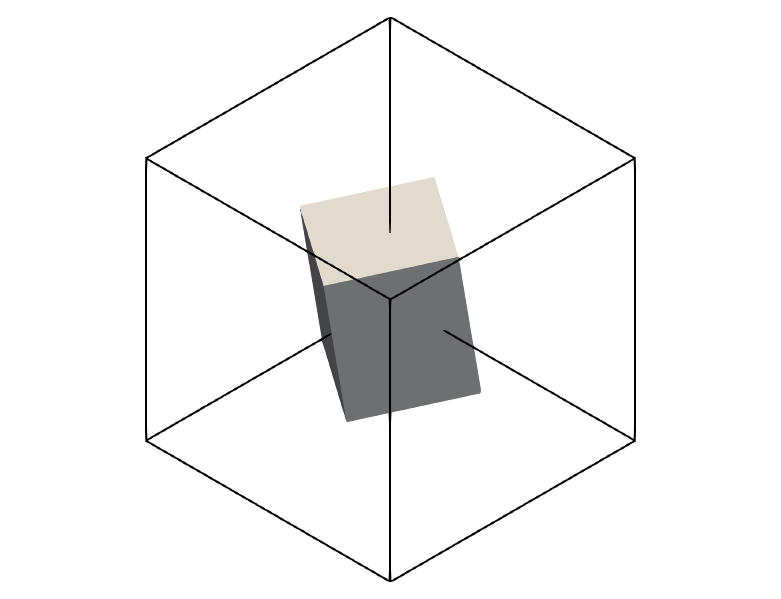}
        \caption{A sample cube inside the lid-driven cavity domain.}
    \end{subfigure}
    \hspace{0.01\linewidth}
    \begin{subfigure}[b]{0.47\linewidth}
        \begin{tikzpicture}
            \node[anchor=south west] (main) at (0,0) {
                \includegraphics[width=0.97\linewidth]{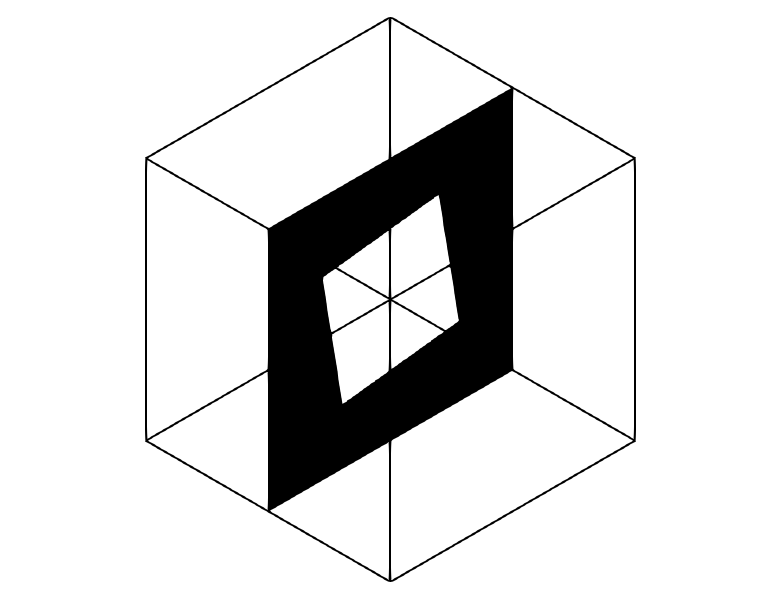}
            };
            % Define inset position and size
            \begin{scope}[xshift=2.5cm, yshift=0.5cm]
                \node[anchor=south east] (zoom) at (main.north east) {
                    \includegraphics[width=0.5\linewidth]{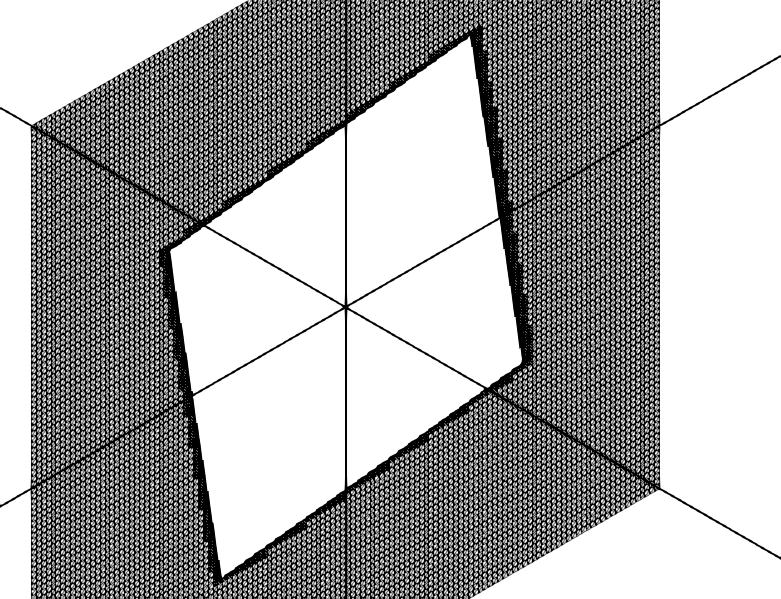}
                };
                % Draw rectangle on the main image to indicate zoom region
                \draw[red, thick] (0.5,1.3) rectangle (2.4,3.6);
                % Draw connecting lines between inset and zoomed region
                \draw[red, thick] (0.5,3.6) -- (zoom.south west);
                \draw[red, thick] (2.4,3.6) -- (zoom.south east);
            \end{scope}
        \end{tikzpicture}
        \caption{Slice of the octree mesh with a zoomed inset showing local refinement near the object boundary.}
    \end{subfigure}
    \begin{subfigure}[b]{0.47\linewidth}
        \includegraphics[width=0.97\linewidth]{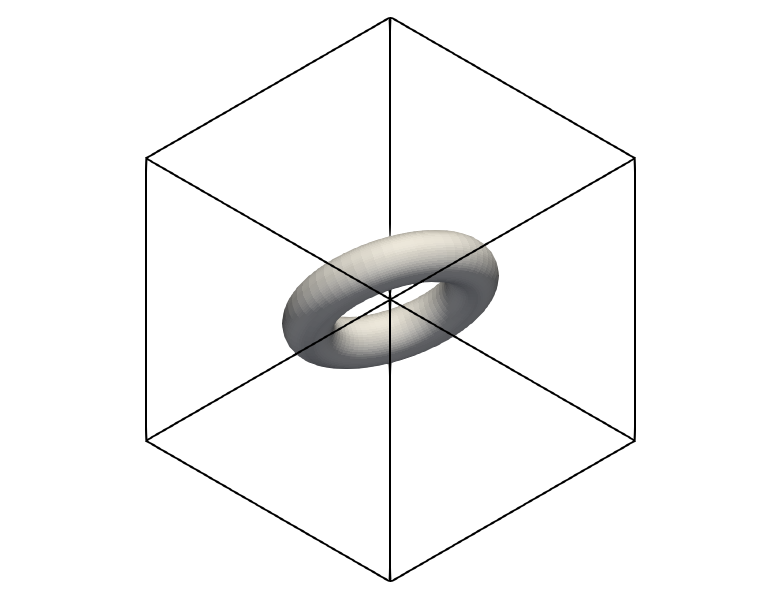}
        \caption{A sample ring inside the lid-driven cavity domain.}
    \end{subfigure}
    \hspace{0.01\linewidth}
    \begin{subfigure}[b]{0.47\linewidth}
        \begin{tikzpicture}
            \node[anchor=south west] (main) at (0,0) {
                \includegraphics[width=0.97\linewidth]{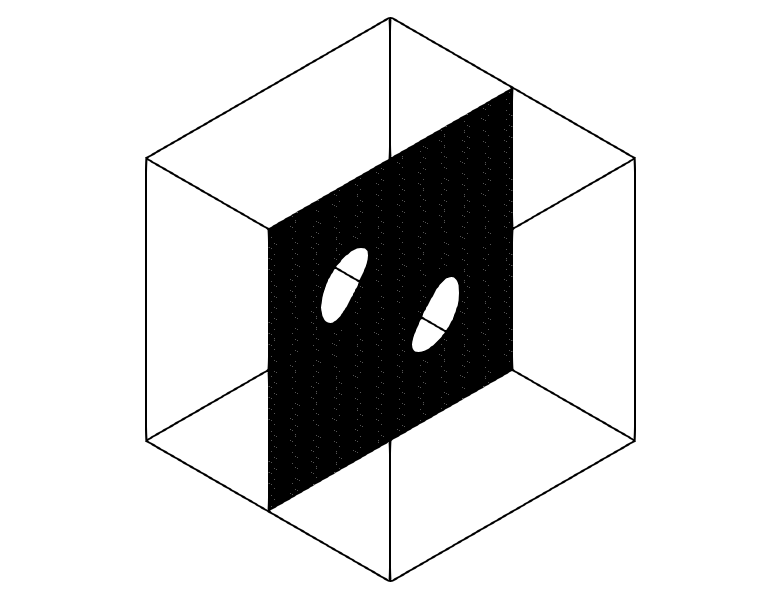}
            };
            % Define inset position and size
            \begin{scope}[xshift=2.5cm, yshift=0.5cm]
                \node[anchor=south east] (zoom) at (main.north east) {
                    \includegraphics[width=0.5\linewidth]{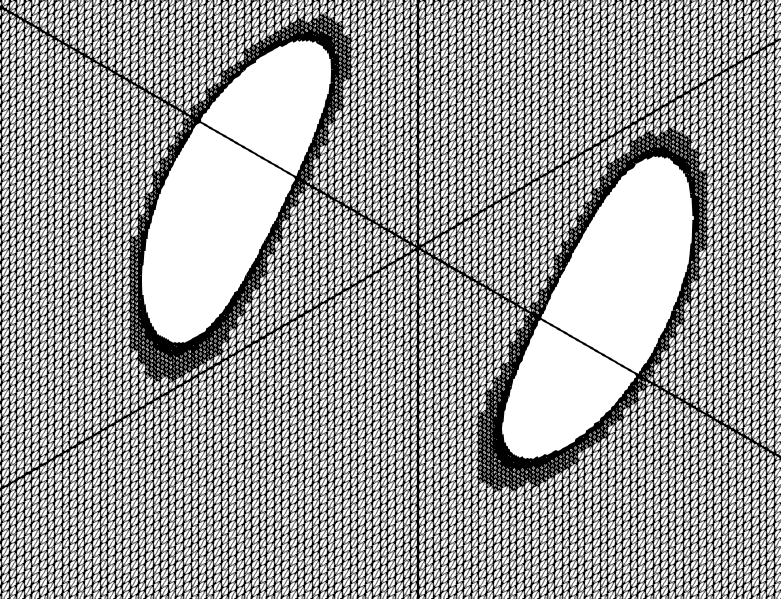}
                };
                % Draw rectangle on the main image to indicate zoom region
                \draw[red, thick] (0.5,1.7) rectangle (2.4,3.2);
                % Draw connecting lines between inset and zoomed region
                \draw[red, thick] (0.5,3.2) -- (zoom.south west);
                \draw[red, thick] (2.4,3.2) -- (zoom.south east);
            \end{scope}
        \end{tikzpicture}
        \caption{Slice of the octree mesh with a zoomed inset showing local refinement near the object boundary.}
    \end{subfigure}    
    \caption{Two examples of the 3D LDC problem showing a zoomed slice of the computational mesh ranging from \(h=\frac{2}{2^9}\) near the geometry to \(h=\frac{2}{2^7}\) in the background.}
    \label{fig:Mesh_LDC}
\end{figure}

To accurately capture fluid--geometry interactions, we employ an octree mesh with local refinement near the immersed object at level~9 (cell size \(2/2^9\)) and a coarser refinement level of 7 (cell size \(2/2^7\)) in the background, resulting in around 2.5 million degrees of freedom. \figref{fig:Mesh_LDC} illustrates two examples of objects inside the cavity and their corresponding mesh. Although the original simulations use this non-uniform refinement scheme, the publicly released FlowBench data are downsampled to a uniform level-7 resolution (\(128 \times 128 \times 128\)) and distributed as compressed NumPy (\texttt{.npz}) files. These files include Signed Distance Fields (SDF), Reynolds numbers, and output velocity components \((u,v,w)\). By offering 1{,}000 high-fidelity samples spanning 100 geometries at multiple Reynolds numbers, FlowBench enables extensive benchmarking of scientific machine-learning models in complex flow scenarios.

\section{Gradient Calculation}  \label{sec:FEM}
\subsection{Evaluation of Velocity and Gradients at Element Center}  \label{subsec:element-center}

Let \(\Omega \subset \mathbb{R}^{3}\) be the flow domain, discretized by a uniform mesh of size \(N_x \times N_y \times N_z\). Denote by \(\mathcal{K}_h\) the resulting partition of finite elements into hexahedral (or cuboidal) elements. Each element \( e \in \mathcal{K}_h\) is endowed with nodal points storing the discrete velocity unknowns 

\[
u_h^i(\mathbf{x}) 
\;=\; 
\bigl( u_h^1(\mathbf{x}),\, u_h^2(\mathbf{x}),\, u_h^3(\mathbf{x}) \bigr),
\]

where \(i \in \{1,2,3\}\) indexes the velocity components \((u, v, w)\).

We assume linear (trilinear) basis functions \(\{\phi_a\}_{a=1}^8\) in each element, so that any point \(\mathbf{x} \in e\) in physical coordinates corresponds to a reference point \(\boldsymbol{\xi} \in [-1,1]^3\). The velocity field at \(\mathbf{x}\) is then

\[    u_h^i(\mathbf{x}) 
    \;=\; 
    \sum_{a=1}^{8} \phi_a(\boldsymbol{\xi})\, u_a^i,
\]

where \(u_a^i\) is the \(i\)-th velocity component at node \(a\) of element \(e\). Because \(\phi_a(\boldsymbol{\xi})\) is linear in each coordinate, its gradient \(\partial_j \phi_a(\boldsymbol{\xi})\) (with \(j \in \{1,2,3\}\) for the spatial directions \(x,y,z\)) is constant throughout the element. Consequently, the velocity gradient becomes

\[
\partial_j u_h^i(\mathbf{x})
    \;=\;
    \sum_{a=1}^{8} 
    u_a^i \,\partial_j \phi_a(\boldsymbol{\xi}).
\]

The repeated spatial indices (\(i,j\)) imply summation (Einstein notation) over \(\{1,2,3\}\), while the node index \(a\) is explicitly summed from 1 to 8. We evaluate the velocity and its gradient at the centroid of each element, i.e.\ at \(\boldsymbol{\xi} = (0,0,0)\) in the reference domain. Hence, the velocity and gradient at the element center become

\[
    u_h^i \Bigr|_{\mathrm{center}} 
    \;=\;
    \sum_{a=1}^{8}
    \phi_a(0,0,0)\,u_a^i,
    \qquad
    \partial_j u_h^i \Bigr|_{\mathrm{center}}
    \;=\;
    \sum_{a=1}^{8}
    u_a^i\,\partial_j \phi_a(0,0,0).
\]

For a standard trilinear element, \(\phi_a(0,0,0) = \tfrac{1}{8}\) at each node, so the centroid evaluation simplifies to the average value of the 8 nodes.

If portions of \(\Omega\) are occupied by solid geometry, we store a signed distance function \(\mathrm{SDF}(\mathbf{x})\) that is strictly positive in fluid regions and negative inside the solid. We ignore any velocity data at element centers for which \(\mathrm{SDF}(\mathbf{x}_c) < 0\), where \(\mathbf{x}_c\) is the centroid of a given element. In effect, we mask out the solution in the solid regions and restrict our velocity and gradient evaluations to those elements whose centers lie in the fluid region of \(\Omega\).

\subsection{Continuity Residuals:} \label{subsec:cont-residuals}

Let \(\mathcal{K}_h\) be the finite element partition of \(\Omega\). To assess the continuity (divergence-free condition) of the velocity field \(u^i(\mathbf{x})\), we note that a perfectly solenoidal flow requires
\[
\partial_i u^i \;=\; 0,
\]
where repeated index \(i\) indicates summation over \(\{1,2,3\}\). In each element \( e \in \mathcal{K}_h\), the derivatives \(\partial_j u^i\) are evaluated at Gauss points, and the divergence at these points is

\[
(\partial_j u^j)_{\text{gp}}
\;=\;
\partial_j u^j \Bigr|_{\text{gp}},
\]
We compute the integral of the squared divergence over each element using a Gauss quadrature scheme with 2 points along each axis (2×2×2):

\[
\| \partial_j u^j \|_{L_2(e)}
\;=\;
\left(
    \int_{e}
    \bigl(\partial_j u^j\bigr)^2 
    \,d\Omega
\right)^{1/2}.
\]
Summing over all elements yields the total continuity residual,
\[
\text{continuity\_sum}
\;=\;
\left(
    \sum_{e \in \mathcal{K}_h}
    \| \partial_j u^j \|_{L_2(e)}^2
\right)^{1/2}.
\]

This metric quantifies how closely the velocity field satisfies the divergence-free condition throughout the domain.

\section{Training Details} 
\label{sec:training-details}
\subsection{Model Hyperparameters} \label{subsec:hyperparameter}

This section provides a detailed overview of the hyperparameters used for training \emph{DeepONet} and \emph{Geometric-DeepONet} for easier reproducibility of our results. These hyperparameters were selected based on extensive tuning to optimize performance for the datasets used in this study.

\begin{itemize}
    \item Branch network layers size (\texttt{branch\_net\_layers}): [512, 512, 512]
    \item Trunk network layers size (\texttt{trunk\_net\_layers}): [256, 256, 256]
    \item Number of latent basis functions (\texttt{modes}): 128
    \item Learning rate (\texttt{lr}): 0.0001
    \item  Batch size (\texttt{batch-size}): 2
\end{itemize}

\subsection{Training and Validation Loss} \label{subsec:loss-plots}

%To further analyze training performance, we present the evolution of training and validation loss for \emph{DeepONet} and \emph{Geometric-DeepONet} across four distinct loss functions: \(L1\)-\(L4\), as shown in~\figref{fig:deeponet-geo-loss}.

\begin{figure} [!ht]
    \centering
    % DeepONet L1
    \begin{subfigure}[b]{0.48\textwidth}
        \centering
        \begin{tikzpicture}
        \begin{semilogyaxis}[
            width=0.9\textwidth,
            height=0.6\textwidth,
            xlabel={Epoch},
            ylabel={Loss},
            title={DeepONet (L1)},
            legend pos=north east,
            grid=major,
            xmin=0, xmax=200,
            ymin=5e-4, ymax=200
        ]
        \addplot[blue, thick] table[x=epoch, y=depo_L1_train, col sep=comma] {Figures/training_loss/train-val_loss.csv};
        \addlegendentry{Training}
        \addplot[orange, thick] table[x=epoch, y=depo_L1_val, col sep=comma] {Figures/training_loss/train-val_loss.csv};
        \addlegendentry{Validation}
        \end{semilogyaxis}
        \end{tikzpicture}
    \end{subfigure}
    \hfill
    % Geometric-DeepONet L1
    \begin{subfigure}[b]{0.48\textwidth}
        \centering
        \begin{tikzpicture}
        \begin{semilogyaxis}[
            width=0.9\textwidth,
            height=0.6\textwidth,
            xlabel={Epoch},
            ylabel={Loss},
            title={Geometric-DeepONet (L1)},
            legend pos=north east,
            grid=major,
            xmin=0, xmax=200,
            ymin=5e-3, ymax=200
        ]
        \addplot[blue, thick] table[x=epoch, y=geo_L1_train, col sep=comma] {Figures/training_loss/train-val_loss.csv};
        \addlegendentry{Training}
        \addplot[orange, thick] table[x=epoch, y=geo_L1_val, col sep=comma] {Figures/training_loss/train-val_loss.csv};
        \addlegendentry{Validation}
        \end{semilogyaxis}
        \end{tikzpicture}
    \end{subfigure}
    % DeepONet L2
    \begin{subfigure}[b]{0.48\textwidth}
        \centering
        \begin{tikzpicture}
        \begin{semilogyaxis}[
            width=0.9\textwidth,
            height=0.6\textwidth,
            xlabel={Epoch},
            ylabel={Loss},
            title={DeepONet (L2)},
            legend pos=north east,
            grid=major,
            xmin=0, xmax=200,
            ymin=5e-4, ymax=200
        ]
        \addplot[blue, thick] table[x=epoch, y=depo_L2_train, col sep=comma] {Figures/training_loss/train-val_loss.csv};
        \addlegendentry{Training}
        \addplot[orange, thick] table[x=epoch, y=depo_L2_val, col sep=comma] {Figures/training_loss/train-val_loss.csv};
        \addlegendentry{Validation}
        \end{semilogyaxis}
        \end{tikzpicture}
    \end{subfigure}
    \hfill
    % Geometric-DeepONet L2
    \begin{subfigure}[b]{0.48\textwidth}
        \centering
        \begin{tikzpicture}
        \begin{semilogyaxis}[
            width=0.9\textwidth,
            height=0.6\textwidth,
            xlabel={Epoch},
            ylabel={Loss},
            title={Geometric-DeepONet (L2)},
            legend pos=north east,
            grid=major,
            xmin=0, xmax=200,
            ymin=5e-3, ymax=200
        ]
        \addplot[blue, thick] table[x=epoch, y=geo_L2_train, col sep=comma] {Figures/training_loss/train-val_loss.csv};
        \addlegendentry{Training}
        \addplot[orange, thick] table[x=epoch, y=geo_L2_val, col sep=comma] {Figures/training_loss/train-val_loss.csv};
        \addlegendentry{Validation}
        \end{semilogyaxis}
        \end{tikzpicture}
    \end{subfigure}
    % DeepONet L3
    \begin{subfigure}[b]{0.48\textwidth}
        \centering
        \begin{tikzpicture}
        \begin{semilogyaxis}[
            width=0.9\textwidth,
            height=0.6\textwidth,
            xlabel={Epoch},
            ylabel={Loss},
            title={DeepONet (L3)},
            legend pos=north east,
            grid=major,
            xmin=0, xmax=200,
            ymin=5e-2, ymax=200
        ]
        \addplot[blue, thick] table[x=epoch, y=depo_L3_train, col sep=comma] {Figures/training_loss/train-val_loss.csv};
        \addlegendentry{Training}
        \addplot[orange, thick] table[x=epoch, y=depo_L3_val, col sep=comma] {Figures/training_loss/train-val_loss.csv};
        \addlegendentry{Validation}
        \end{semilogyaxis}
        \end{tikzpicture}
    \end{subfigure}
    \hfill
    % Geometric-DeepONet L3
    \begin{subfigure}[b]{0.48\textwidth}
        \centering
        \begin{tikzpicture}
        \begin{semilogyaxis}[
            width=0.9\textwidth,
            height=0.6\textwidth,
            xlabel={Epoch},
            ylabel={Loss},
            title={Geometric-DeepONet (L3)},
            legend pos=north east,
            grid=major,
            xmin=0, xmax=200,
            ymin=5e-2, ymax=200
        ]
        \addplot[blue, thick] table[x=epoch, y=geo_L3_train, col sep=comma] {Figures/training_loss/train-val_loss.csv};
        \addlegendentry{Training}
        \addplot[orange, thick] table[x=epoch, y=geo_L3_val, col sep=comma] {Figures/training_loss/train-val_loss.csv};
        \addlegendentry{Validation}
        \end{semilogyaxis}
        \end{tikzpicture}
    \end{subfigure}
    % DeepONet L4
    \begin{subfigure}[b]{0.48\textwidth}
        \centering
        \begin{tikzpicture}
        \begin{semilogyaxis}[
            width=0.9\textwidth,
            height=0.6\textwidth,
            xlabel={Epoch},
            ylabel={Loss},
            title={DeepONet (L4)},
            legend pos=north east,
            grid=major,
            xmin=0, xmax=200,
            ymin=5e-2, ymax=200
        ]
        \addplot[blue, thick] table[x=epoch, y=depo_L4_train, col sep=comma] {Figures/training_loss/train-val_loss.csv};
        \addlegendentry{Training}
        \addplot[orange, thick] table[x=epoch, y=depo_L4_val, col sep=comma] {Figures/training_loss/train-val_loss.csv};
        \addlegendentry{Validation}
        \end{semilogyaxis}
        \end{tikzpicture}
    \end{subfigure}
    \hfill
    % Geometric-DeepONet L4
    \begin{subfigure}[b]{0.48\textwidth}
        \centering
        \begin{tikzpicture}
        \begin{semilogyaxis}[
            width=0.9\textwidth,
            height=0.6\textwidth,
            xlabel={Epoch},
            ylabel={Loss},
            title={Geometric-DeepONet (L4)},
            legend pos=north east,
            grid=major,
            xmin=0, xmax=200,
            ymin=5e-2, ymax=200
        ]
        \addplot[blue, thick] table[x=epoch, y=geo_L4_train, col sep=comma] {Figures/training_loss/train-val_loss.csv};
        \addlegendentry{Training}
        \addplot[orange, thick] table[x=epoch, y=geo_L4_val, col sep=comma] {Figures/training_loss/train-val_loss.csv};
        \addlegendentry{Validation}
        \end{semilogyaxis}
        \end{tikzpicture}
    \end{subfigure}
    \caption{Training and validation loss in semi-log scale for \emph{DeepONet} (left) and \emph{Geometric-DeepONet} (right) across \(L1\)-\(L4\). This figure presents the evolution of both training (blue) and validation (orange) losses over 200 epochs for each model and loss function.}
    \label{fig:deeponet-geo-loss}
\end{figure}
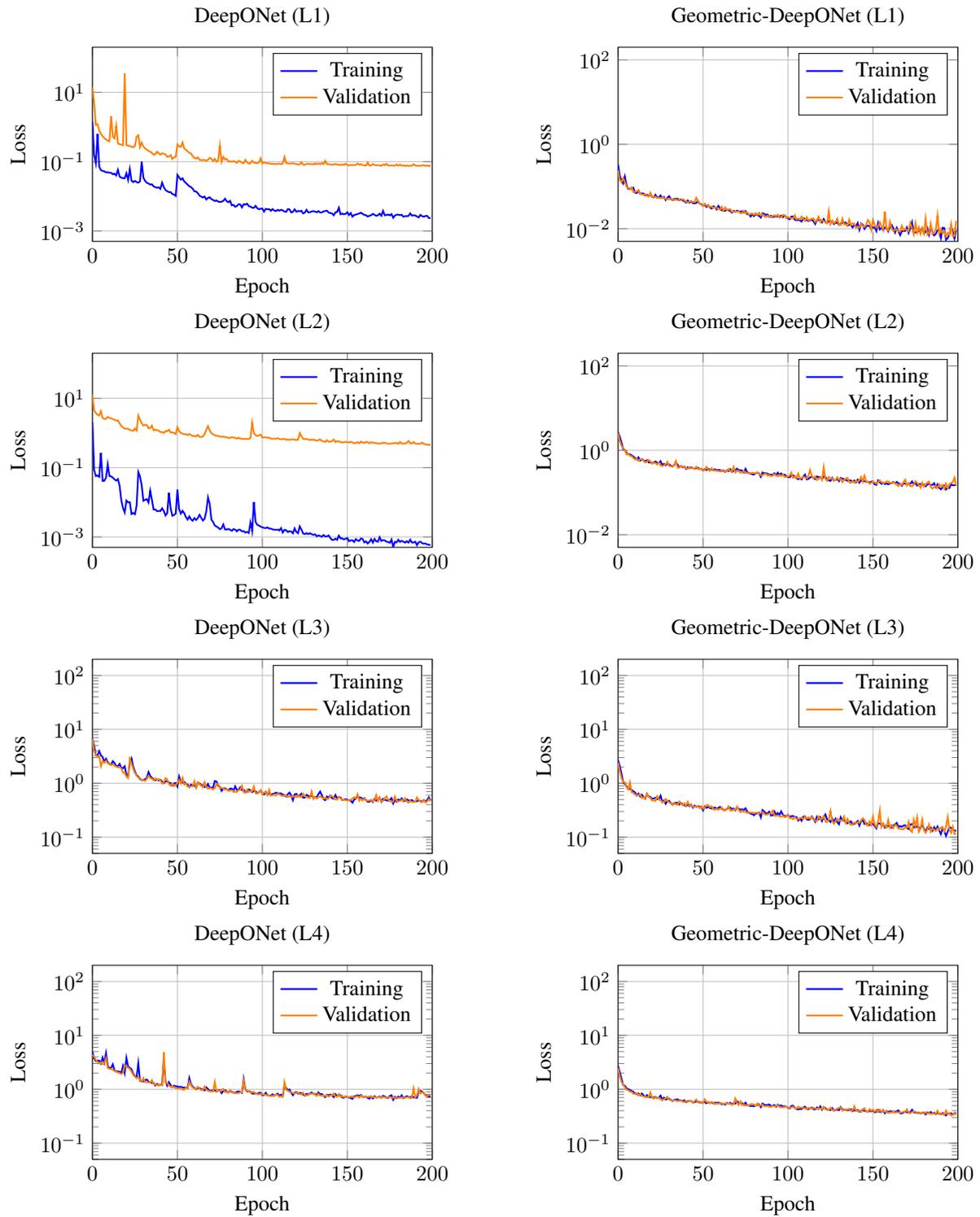

\end{document}